\documentclass[11pt]{article}

\usepackage[preprint]{acl}

\usepackage{times}
\usepackage{latexsym}
\usepackage[T1]{fontenc}
\usepackage[utf8]{inputenc}
\usepackage{microtype}
\usepackage{graphicx}
\graphicspath{{figures/}}
\usepackage{booktabs}
\usepackage{amsmath}
\usepackage{amssymb}
\usepackage{multirow}
\usepackage{xspace}
\usepackage{subcaption}
\usepackage{array}
\usepackage{makecell}
\usepackage{pifont}
\usepackage[dvipsnames]{xcolor}

\newcommand{\method}{SLIM\xspace}
\newcommand{\accat}[1]{\text{Acc}@{#1}}

\title{SLIM: Sparse Latent Steering for Interpretable and Property-Directed LLM-Based Molecular Editing}

\author{
  Mingxu Zhang\textsuperscript{1},
  Yuhan Li\textsuperscript{1},
  Lujundong Li\textsuperscript{1},
  Dazhong Shen\textsuperscript{2}\thanks{Corresponding authors.},
  Hui Xiong\textsuperscript{1}\thanks{Corresponding authors.},
  Ying Sun\textsuperscript{3}\footnotemark[1] \\
  \textsuperscript{1}The Hong Kong University of Science and Technology (Guangzhou) \\
  \textsuperscript{2}Nanjing University of Aeronautics and Astronautics \\
  \textsuperscript{3}The 63rd Research Institute, National University of Defense Technology, Nanjing\\
  \texttt{mzhang630@connect.hkust-gz.edu.cn}, 
  \texttt{shendazhong@nuaa.edu.cn}, \\
  \texttt{sunyinggilly@gmail.com}
}

\begin{document}
\maketitle

\begin{abstract}
Large language models possess strong chemical reasoning capabilities, making them effective molecular editors. However, property-relevant information is implicitly entangled across their dense hidden states, providing no explicit handle for property control: a substantial fraction of edits fail to improve or even degrade target properties. To address these issues, we propose \textbf{SLIM} (Sparse Latent Interpretable Molecular editing), a plug-and-play framework that decomposes the editor's hidden states into sparse, property-aligned features via a Sparse Autoencoder with learnable importance gates. Steering in this sparse feature space precisely activates property-relevant dimensions, improving editing success rate without modifying model parameters. The same sparse basis further supports interpretable analysis of editing behavior. Experiments on the MolEditRL benchmark across four model architectures and eight molecular properties show consistent gains over baselines, with improvements of up to 42.4 points.
\end{abstract}

\section{Introduction}
\label{sec:intro}

Molecular editing, the task of modifying a given molecule to optimize specific properties while preserving its core structure, is a fundamental step in drug discovery and materials science~\citep{hughes2011principles, jimenez2020drug}. In real-world pharmaceutical development, a promising lead compound often requires iterative structural refinements to improve druglikeness, synthetic accessibility, binding affinity, or other pharmacological properties. The ability to perform such targeted modifications efficiently and reliably holds great potential for accelerating the drug design pipeline.

Early computational approaches to molecular optimization have relied on reinforcement learning~\citep{blaschke2020reinvent, edwards2024moleditrl} and diffusion models~\citep{lee2023exploring, igashov2024equivariant}. Recently, large language models (LLMs) have demonstrated strong potential for molecular editing~\citep{ye2023drugassist, liu2024gellmo, fang2024molgen, pei2023biot5}. LLMs acquire rich chemical priors, including structure--activity relationships, functional group chemistry, and pharmacophore patterns, through pretraining on massive corpora of chemical literature~\citep{zhang2025chematp, zhang2025atomdisc}. Post training methods then bridge this chemical knowledge with SMILES-level structural transformations, positioning LLM-based editors to produce chemically meaningful modifications grounded in domain knowledge.

Despite these advances, existing LLM-based editors provide no explicit handle for property control. Property-relevant information is implicitly entangled across the model's dense hidden states, making it impossible to selectively amplify a target property without retraining. As a result, supervised fine-tuning offers only coarse control: a substantial fraction of generated edits fail to improve, or even degrade, the target property. Improving a specific property currently requires collecting new paired data and retraining the model, a costly process that scales poorly across properties and models.

To address this, we propose \method, a plug-and-play framework that decomposes the editor's hidden states into sparse, property-aligned features via a task-oriented Sparse Autoencoder (SAE)~\citep{cunningham2023sparse, bricken2023monosemanticity}. Unlike standard SAEs trained for reconstruction alone, our SAE is jointly trained with contrastive, predictive, and gradient-alignment objectives that shape the sparse basis to be property-aware. Per-property Importance Gates then learn to select, from this shared basis, the subset of features most relevant to each target property. To steer the model, we project a gradient-derived causal direction through the top-$k$ selected features of the SAE, producing a sparse steering vector that precisely activates property-relevant dimensions at inference time without modifying model parameters. The sparse basis further supports interpretable analysis, as individual SAE features encode recognizable chemical semantics.

In summary, our contributions are as follows:
\begin{itemize}
\setlength\itemsep{0pt}
\item We propose \method, a plug-and-play framework that decomposes dense LLM hidden states into sparse, property-aligned features for property-directed steering, requiring no retraining of the base molecular editor.
\item We introduce task-oriented SAE with contrastive, predictive, and gradient-alignment objectives and per-property Importance Gates, ensuring that the learned features are causally aligned with target properties.
\item Experimental results on the MolEditRL benchmark across four model architectures and eight molecular properties show consistent improvements of up to 42.4\%. Case studies further show that individual SAE features encode interpretable chemical semantics.
\end{itemize}
\section{Related Work}
\label{sec:related}
\paragraph{Molecular Editing.}
Molecular editing aims to modify a given molecule to optimize target properties while preserving structural similarity. Early approaches use reinforcement learning with scoring functions~\citep{blaschke2020reinvent} or combine discrete diffusion with RL~\citep{edwards2024moleditrl}. More recently, LLM-based methods have shown strong results. DrugAssist~\citep{ye2023drugassist} fine-tunes LLaMA-2~\citep{touvron2023llama2} on paired molecular data, framing editing as instruction following. GeLLM3O~\citep{liu2024gellmo} extends this to LLaMA-3~\citep{grattafiori2024llama3} and Mistral~\citep{jiang2023mistral}, demonstrating cross-model transferability. MolGen~\citep{fang2024molgen} adopts a BART-based~\citep{lewis2020bart} encoder-decoder trained on SELFIES~\citep{krenn2020selfies} with chemical feedback. These methods optimize molecular properties through training but offer no mechanism to interpret or steer the learned representations at inference time.

\paragraph{Molecular Representation Learning.}
Understanding how models represent molecular structure and properties is central to interpretable editing. ChemATP~\citep{zhang2025chematp} introduces a training-free chemical reasoning framework that leverages LLM internal representations. AtomDisc~\citep{zhang2025atomdisc} proposes atom-level tokenization that reveals interpretable structure--property associations within LLM hidden states. More broadly, the \textit{linear representation hypothesis}~\citep{park2024linear} suggests that high-level concepts are encoded as directions in activation space, motivating methods that extract and manipulate these directions for controllable generation.

\paragraph{Sparse Autoencoders and Activation Steering.}
Neural networks represent more features than they have dimensions through superposition~\citep{elhage2022superposition}. Sparse autoencoders (SAEs) recover these latent features by learning an overcomplete basis~\citep{sharkey2022taking, cunningham2023sparse, bricken2023monosemanticity}. \citet{templeton2024scaling} show that SAE features scale to large models and capture semantically meaningful concepts. Gated SAEs~\citep{rajamanoharan2024improving} improve feature quality by separating gating from magnitude estimation, and a recent survey~\citep{gao2025saesurvey} covers SAE architectures comprehensively. On the steering side, activation addition~\citep{turner2023activation} modifies model behavior by adding direction vectors to hidden states during inference. Contrastive activation addition~\citep{rimsky2024steering} computes directions as mean differences between positive and negative activations, and representation engineering~\citep{zou2023representation} extends this with a top-down approach. However, existing SAEs are trained with reconstruction objectives alone and used only for post-hoc analysis, while steering methods rely on mean-difference directions that we show fail for molecular property control (\S\ref{sec:ablation_direction}). \method bridges these two lines by integrating task-oriented SAE training with gradient-based steering, producing sparse property-aligned features that enable precise, training-free property control.

\section{Method}
\label{sec:method}

\begin{figure*}[t]
\centering
\includegraphics[width=\textwidth]{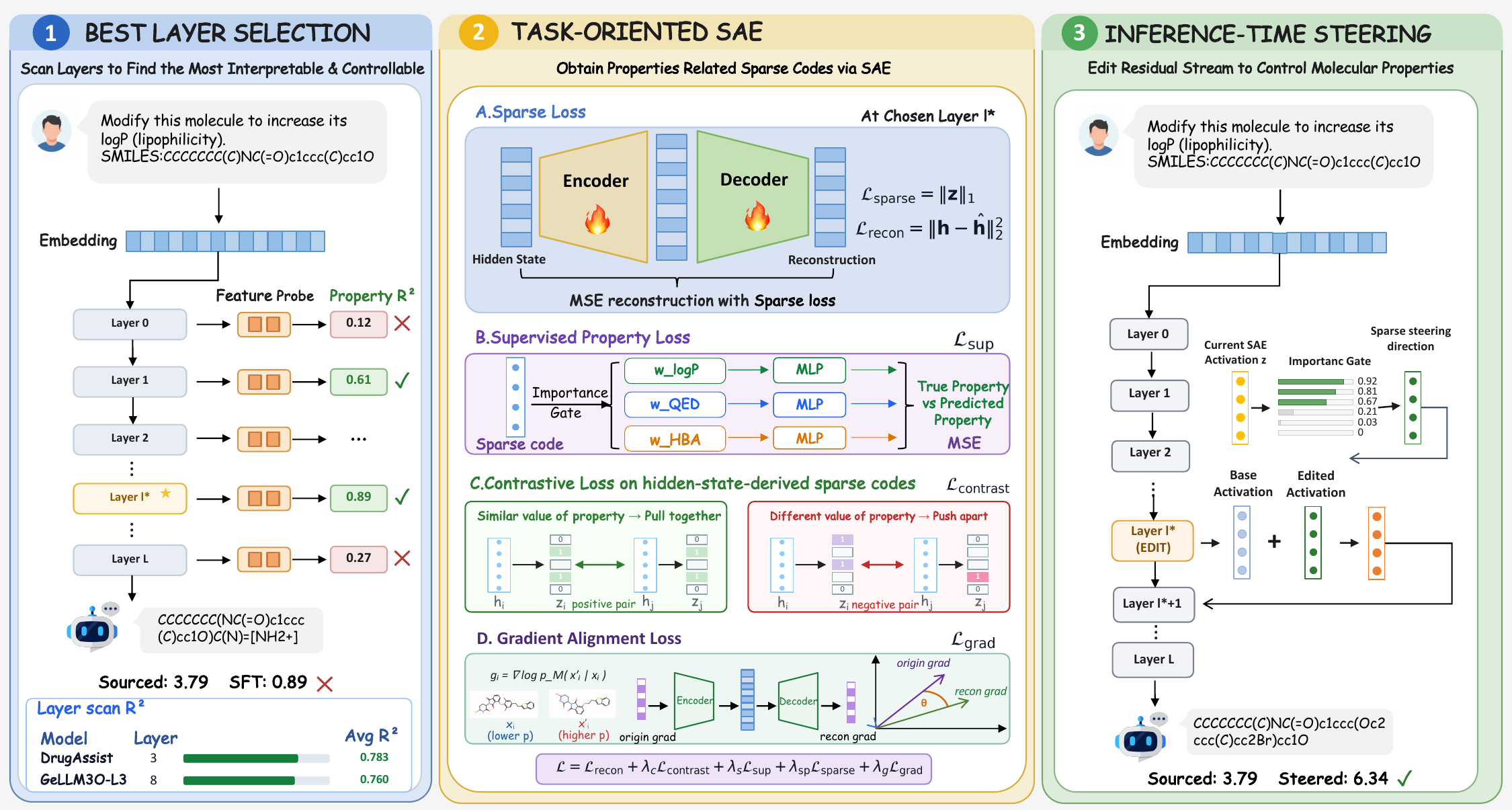}
\caption{Overview of the \method framework. \textbf{Stage~1}: Ridge probes scan all layers to identify the optimal intervention point $l^*$. \textbf{Stage~2}: A task-oriented SAE is trained at layer $l^*$ with four objectives: (A)~sparse reconstruction, (B)~supervised property prediction via per-property Importance Gates, (C)~contrastive alignment of importance-gated sparse codes, and (D)~gradient alignment to ensure the SAE basis faithfully represents causal steering directions. \textbf{Stage~3}: At inference time, a sparse steering vector is added to the residual stream at layer $l^*$, directing the model toward improved molecular properties without modifying model parameters.}
\label{fig:pipeline}
\end{figure*}

\method operates as a plug-and-play module on top of any molecular editing LLM. The pipeline consists of three stages: (1) a layer scan to identify the optimal intervention point, (2) task-oriented SAE training with gradient alignment, and (3) inference-time activation steering via interpretable sparse directions. Figure~\ref{fig:pipeline} illustrates the overall framework.

\subsection{Layer Scan}
\label{sec:layer_select}

As shown in Figure~\ref{fig:pipeline} (Stage~1), different layers encode different levels of chemical abstraction. We select the layer $l^*$ whose hidden states are most informative about molecular properties by fitting ridge regression probes~\citep{park2024linear} on activations extracted from 5,000 ZINC molecules:
\begin{equation}
l^* = \arg\max_l \frac{1}{|P|} \sum_{p \in P} R^2(l, p),
\end{equation}
where $P$ is the set of target properties and $R^2(l, p)$ is the coefficient of determination for property $p$ at layer $l$. The layer with the highest average $R^2$ concentrates the most property-relevant information in its hidden states, making it the most effective point for steering: interventions at this layer can maximally influence generation because the representation is richest in the signals we aim to amplify.

\subsection{Task-Oriented SAE}
\label{sec:sae}

As illustrated in Figure~\ref{fig:pipeline} (Stage~2A), we train a Gated SAE~\citep{rajamanoharan2024improving} on hidden states $\mathbf{h} \in \mathbb{R}^d$ extracted at layer $l^*$ from 50,000 molecules. The SAE uses $D = 8d$ latent dimensions. Encoding produces a sparse code $\mathbf{z} \in \mathbb{R}^D$:
\begin{equation}
\mathbf{z} = \sigma(\mathbf{W}_g(\mathbf{h} - \mathbf{b}_d)) \odot \text{ReLU}(\mathbf{W}_m(\mathbf{h} - \mathbf{b}_d)),
\end{equation}
where $\mathbf{W}_g, \mathbf{W}_m \in \mathbb{R}^{D \times d}$ are gate and magnitude matrices, $\sigma$ is the sigmoid function, and $\odot$ denotes element-wise multiplication. Reconstruction is $\hat{\mathbf{h}} = \mathbf{W}_d \mathbf{z} + \mathbf{b}_d$.
\paragraph{Importance Gates.}
As shown in Figure~\ref{fig:pipeline} (Stage~2B), each property $p$ has a learnable Importance Gate $\mathbf{w}_p \in \mathbb{R}^D$ that identifies which SAE features are relevant to $p$. The gated sparse code for property $p$ is $\mathbf{z}_p = \mathbf{w}_p \odot \mathbf{z}$, which feeds into per-property contrastive and predictive heads. This enables the SAE to learn property-aligned feature decompositions while sharing the same underlying sparse basis across all properties.

\paragraph{Training Objective.}
Our task-oriented SAE is trained with a multi-objective loss:
\begin{equation}
\tiny
\mathcal{L} = \mathcal{L}_\text{recon} + \lambda_c \mathcal{L}_\text{contrast} + \lambda_s \mathcal{L}_\text{sup} + \lambda_\text{sp} \mathcal{L}_\text{sparse} + \lambda_g \mathcal{L}_\text{grad},
\label{eq:sae_loss}
\end{equation}
where $\mathcal{L}_\text{recon} = \|\mathbf{h} - \hat{\mathbf{h}}\|^2$ ensures the SAE faithfully reconstructs activations, preserving property-related information needed for downstream generation. $\mathcal{L}_\text{sparse} = \|\mathbf{z}\|_1$ encourages activation sparsity, ensuring each molecule is represented by a small number of active features.

\paragraph{Supervised Loss.}
As shown in Figure~\ref{fig:pipeline} (Stage~2B), $\mathcal{L}_\text{sup}$ trains a per-property MLP head $f_p$ to predict the property value from the gated code:
$\mathcal{L}_\text{sup} = \frac{1}{|P|} \sum_{p \in P} \frac{1}{n} \sum_{i=1}^{n} (f_p(\mathbf{w}_p \odot \mathbf{z}_i) - y_i^{(p)})^2$,
where $y_i^{(p)}$ is the oracle-computed property value. While $\mathcal{L}_\text{contrast}$ only requires the gate to distinguish high from low property values, $\mathcal{L}_\text{sup}$ demands that the selected features contain sufficient information to regress the precise numerical value, providing a stronger training signal for the Importance Gate.

\paragraph{Contrastive Loss.}
As shown in Figure~\ref{fig:pipeline} (Stage~2C), $\mathcal{L}_\text{contrast}$ is a group contrastive loss that operates on importance-gated sparse codes. For each property $p$, molecules in a batch are ranked by oracle-computed property values; the top 25\% form the positive group $\mathcal{P}$ and the bottom 25\% form the negative group $\mathcal{N}$. The gated codes $\tilde{\mathbf{z}}_i = \text{norm}(\mathbf{w}_p \odot \mathbf{z}_i)$ are L2-normalized, and the loss is:
$\mathcal{L}_\text{contrast}^{(p)} = -\frac{1}{|\mathcal{P}|}\sum_{i \in \mathcal{P}} \log \frac{\sum_{j \in \mathcal{P}, j \neq i} \exp(\tilde{\mathbf{z}}_i^\top \tilde{\mathbf{z}}_j / \tau_c)}{\sum_{k \neq i} \exp(\tilde{\mathbf{z}}_i^\top \tilde{\mathbf{z}}_k / \tau_c)}$. All same-group pairs are treated as positives and cross-group pairs as negatives, encouraging the Importance Gate to select features that separate molecules by property value.

\paragraph{Gradient-Alignment Loss.}
As shown in Figure~\ref{fig:pipeline} (Stage~2D), the key innovation is $\mathcal{L}_\text{grad}$, which ensures the SAE basis faithfully represents property-relevant steering directions. For each property $p$, we pre-compute a gradient-based direction from the SFT training pairs $\{(\mathbf{x}_i, \mathbf{x}'_i)\}_{i=1}^N$:
\begin{equation}
\mathbf{d}_\text{grad}^{(p)} = \frac{1}{N} \sum_{i=1}^{N} \frac{\nabla_{\mathbf{h}_{l^*}} \log p_\mathcal{M}(\mathbf{x}'_i \mid \mathbf{x}_i)}{\|\nabla_{\mathbf{h}_{l^*}} \log p_\mathcal{M}(\mathbf{x}'_i \mid \mathbf{x}_i)\|}
\label{eq:grad_dir}
\end{equation}
where the gradient is averaged across all token positions and $\mathbf{x}'_i$ has improved property value. This direction captures the perturbation to $\mathbf{h}_{l^*}$ that maximally increases the model's probability of generating the property-improved target. The gradient-alignment loss then encourages the SAE's encode-then-decode pipeline to preserve this direction:
\begin{equation}
\tiny
\mathcal{L}_\text{grad} = \sum_{p \in P} \left(1 - \cos\left(\mathbf{W}_d \cdot \text{top}_k\!\left(\text{enc}(\mathbf{d}_\text{grad}^{(p)})\right),\; \mathbf{d}_\text{grad}^{(p)}\right)\right)
\end{equation}

\paragraph{Joint Objective.}
These five losses serve complementary roles. $\mathcal{L}_\text{recon}$ and $\mathcal{L}_\text{sparse}$ learn a faithful sparse decomposition of model hidden states. $\mathcal{L}_\text{contrast}$ and $\mathcal{L}_\text{sup}$ shape the Importance Gates to select property-relevant features. $\mathcal{L}_\text{grad}$ ensures the resulting sparse basis can faithfully reconstruct causal steering directions for downstream tasks.

\subsection{Inference-Time Activation Steering}
\label{sec:steering}

As shown in Figure~\ref{fig:pipeline} (Stage~3), after training we extract a steering direction for each property $p$ by projecting $\mathbf{d}_\text{grad}^{(p)}$ through the trained SAE:
\begin{equation}
\mathbf{d}_\text{steer}^{(p)} = \frac{\mathbf{W}_d \cdot \text{top}_k(\text{enc}(\mathbf{d}_\text{grad}^{(p)}), k)}{\|\mathbf{W}_d \cdot \text{top}_k(\text{enc}(\mathbf{d}_\text{grad}^{(p)}), k)\|}
\label{eq:csae}
\end{equation}
where $\text{top}_k(\cdot, k)$ retains the $k$ largest-magnitude features. This decomposes the dense direction into a sparse, interpretable set of SAE features, where each feature has a known chemical semantics.

At inference, the steering vector is added to the residual stream at all token positions in layer $l^*$:
\begin{equation}
\mathbf{h}'_{l^*,t} = \mathbf{h}_{l^*,t} + \alpha \cdot \mathbf{d}_\text{steer}^{(p)} \quad \forall\, t
\label{eq:steer}
\end{equation}
where $\alpha > 0$ controls steering strength. This requires \textit{no retraining}: the base model weights remain frozen, and only the added direction changes. For multi-property steering, directions combine via vector addition: $\mathbf{d}_\text{multi} = \sum_p \alpha_p \cdot \mathbf{d}_\text{steer}^{(p)}$.

\section{Experiments}
\label{sec:experiments}
\begin{table*}[t]
\centering
\resizebox{\linewidth}{!}{
\begin{tabular}{l cccc cccc c cccc cccc}
\toprule
& \multicolumn{8}{c}{\textbf{Acc@0.15 (\%)}} && \multicolumn{8}{c}{\textbf{Acc@0.65 (\%)}} \\
\cmidrule(lr){2-9} \cmidrule(lr){11-18}
& \multicolumn{2}{c}{\textbf{DA}} & \multicolumn{2}{c}{\textbf{L3}} & \multicolumn{2}{c}{\textbf{Mi}} & \multicolumn{2}{c}{\textbf{MG}} && \multicolumn{2}{c}{\textbf{DA}} & \multicolumn{2}{c}{\textbf{L3}} & \multicolumn{2}{c}{\textbf{Mi}} & \multicolumn{2}{c}{\textbf{MG}} \\
\cmidrule(lr){2-3} \cmidrule(lr){4-5} \cmidrule(lr){6-7} \cmidrule(lr){8-9} \cmidrule(lr){11-12} \cmidrule(lr){13-14} \cmidrule(lr){15-16} \cmidrule(lr){17-18}
\textbf{Property} & SFT & +S & SFT & +S & SFT & +S & SFT & +S && SFT & +S & SFT & +S & SFT & +S & SFT & +S \\
\midrule
QED $\uparrow$     & 77.0 & \textbf{77.6} & 95.8 & \textbf{97.2} & 71.0 & \textbf{98.2} & 42.4 & \textbf{76.2} && 60.4 & \textbf{61.8} & 32.4 & \textbf{34.4} & 52.2 & 48.6          & 19.4 & \textbf{36.6} \\
DRD2 $\uparrow$    & 47.6 & \textbf{50.6} & 88.2 & \textbf{88.8} & 52.4 & \textbf{90.8} & 37.2 & \textbf{83.6} && 35.8 & \textbf{42.6} & 29.6 & \textbf{30.4} & 40.2 & \textbf{47.2} & 23.0 & \textbf{33.4} \\
logP $\uparrow$    & 55.8 & \textbf{61.2} & 88.2 & \textbf{89.6} & 58.0 & \textbf{94.4} & 46.2 & \textbf{86.4} && 43.4 & \textbf{51.4} & 43.2 & \textbf{48.6} & 44.0 & \textbf{60.2} & 27.2 & \textbf{46.2} \\
MW $\uparrow$      & 41.8 & \textbf{46.8} & 86.8 & \textbf{89.8} & 51.4 & \textbf{93.0} & 48.2 & \textbf{87.0} && 32.2 & \textbf{38.6} & 44.6 & \textbf{48.4} & 37.0 & \textbf{63.8} & 22.6 & \textbf{46.4} \\
RotBond $\uparrow$ & 36.4 & \textbf{41.2} & 55.8 & \textbf{64.2} & 38.6 & \textbf{75.0} & 43.2 & \textbf{85.6} && 27.0 & \textbf{33.8} & 26.4 & \textbf{29.8} & 29.6 & \textbf{43.6} & 18.0 & \textbf{35.4} \\
SA $\downarrow$    & 67.6 & 67.6          & 94.0 & \textbf{94.8} & 65.4 & \textbf{95.8} & 44.4 & \textbf{49.4} && 54.8 & \textbf{55.6} & 32.6 & \textbf{35.4} & 49.8 & \textbf{55.4} & 25.2 & \textbf{31.2} \\
HBA $\uparrow$     & 50.6 & \textbf{77.4} & 45.0 & \textbf{47.6} & 49.4 & \textbf{54.2} & 42.6 & \textbf{55.0} && 38.4 & \textbf{56.0} & 20.8 & \textbf{27.2} & 37.0 & 30.6          & 23.0 & \textbf{32.4} \\
HBD $\uparrow$     & 49.6 & \textbf{74.2} & 28.8 & \textbf{42.2} & 37.4 & \textbf{47.4} & 33.8 & \textbf{40.8} && 37.2 & \textbf{55.2} & 13.8 & \textbf{23.8} & 25.4 & \textbf{30.0} & 16.8 & \textbf{18.2} \\
\midrule
\textbf{Avg}       & 53.3 & \textbf{62.1} & 72.8 & \textbf{76.8} & 52.9 & \textbf{81.1} & 42.3 & \textbf{70.5} && 41.4 & \textbf{49.4} & 30.4 & \textbf{34.8} & 39.4 & \textbf{47.4} & 21.9 & \textbf{35.0} \\
$\Delta$           & \multicolumn{2}{c}{\textbf{+8.8}} & \multicolumn{2}{c}{\textbf{+3.9}} & \multicolumn{2}{c}{\textbf{+28.1}} & \multicolumn{2}{c}{\textbf{+28.2}} && \multicolumn{2}{c}{\textbf{+8.0}} & \multicolumn{2}{c}{\textbf{+4.3}} & \multicolumn{2}{c}{\textbf{+8.0}} & \multicolumn{2}{c}{\textbf{+13.1}} \\
\bottomrule
\end{tabular}}
\caption{Forward-direction results on the MolEditRL benchmark (500 test molecules, $n\!=\!5$ candidates). +S denotes SFT with \method steering. $\Delta$ = average gain over SFT. Bold = better of SFT vs.\ +S per cell.}
\label{tab:main_results}
\end{table*}

In this section, we empirically validate \method's effectiveness and design choices. We aim to answer four research questions:
\textbf{RQ1}: Does \method improve molecular editing across diverse model architectures and properties? (\S\ref{sec:main_results})
\textbf{RQ2}: Are gradient-based steering directions essential, or do simpler alternatives like mean-difference (CAA) suffice? (\S\ref{sec:ablation_direction})
\textbf{RQ3}: Does the task-oriented SAE training provide gains beyond a vanilla reconstruction-only SAE? (\S\ref{sec:ablation_sae})
\textbf{RQ4}: Can \method's steering directions compose for multi-property optimization? (\S\ref{sec:reverse_multi})

\subsection{Experimental Setup}
\label{sec:setup}

\paragraph{Models.}
We evaluate \method on four molecular editing models spanning diverse architectures (Table~\ref{tab:models}): \textbf{DrugAssist}~\citep{ye2023drugassist} (LLaMA-2-7B, $l^*\!=\!3$), \textbf{GeLLM3O-LLaMA3}~\citep{liu2024gellmo} (LLaMA-3.1-8B, $l^*\!=\!8$), \textbf{GeLLM3O-Mistral} (Mistral-7B, $l^*\!=\!14$), and \textbf{MolGen}~\citep{fang2024molgen} (BART-Large, $l^*\!=\!5$). This covers three decoder-only causal LMs and one encoder-decoder.

\paragraph{Properties and Metrics.}
We evaluate on eight molecular properties from the MolEditRL benchmark~\citep{edwards2024moleditrl}: QED~\citep{bickerton2012quantifying}, DRD2, logP, MW, RotBond, SA~\citep{ertl2009estimation}, HBA, and HBD (full descriptions in Appendix~\ref{sec:app_settings}). We report $\accat{\tau}$: the percentage of test molecules for which at least one generated candidate improves the target property and has Tanimoto similarity $\geq \tau$ to the input, using Morgan fingerprints~\citep{rogers2010extended}. We use $\tau = 0.15$ (permissive) and $\tau = 0.65$ (strict). The test set contains 500 molecules, with $n=5$ candidates per molecule.

\subsection{Main Results}
\label{sec:main_results}

To answer \textbf{RQ1}, we evaluate \method on all four models across eight properties at both similarity thresholds (Table~\ref{tab:main_results}).

As shown in Table~\ref{tab:main_results}, \method improves 31/32 model-property pairs at $\tau=0.15$. The gains are largest on models with weak SFT baselines: \textbf{Mistral} QED jumps from 71.0 to 98.2 (+27.2), and \textbf{MolGen} RotBond from 43.2 to 85.6 (+42.4), the single largest improvement. \textbf{DrugAssist} shows consistent moderate gains, with counting properties HBA (+26.8) and HBD (+24.6) as standouts. Even \textbf{LLaMA-3}, the strongest SFT model (72.8 avg), improves by +3.9 on average. At the strict threshold $\tau = 0.65$, all four models maintain positive average gains (DA +8.0, L3 +4.3, Mi +8.0, MG +13.1), indicating that \method improves target properties \textit{without sacrificing molecular similarity}.

\paragraph{Comparison with external baselines.}
To position \method against methods from different paradigms, we compare with REINVENT4~\citep{blaschke2020reinvent} (RL-based) and MolEditRL~\citep{edwards2024moleditrl} (discrete diffusion + RL) using their published results on the same 500-molecule test set. Table~\ref{tab:external} reports the best \method result across our four base models for each property, compared against these external baselines on the six properties reported in both works (QED and DRD2 are omitted as MolEditRL does not report single-property results for these).

\begin{table}[t]
\centering
\small
\resizebox{\columnwidth}{!}{
\begin{tabular}{l ccc ccc}
\toprule
& \multicolumn{3}{c}{Acc@0.15 (\%)} & \multicolumn{3}{c}{Acc@0.65 (\%)} \\
\cmidrule(lr){2-4} \cmidrule(lr){5-7}
\textbf{Property} & REINV. & MolEd. & +\method & REINV. & MolEd. & +\method \\
\midrule
logP $\uparrow$    & 36.0 & 91.0          & \textbf{94.4} & 11.4 & 57.8          & \textbf{60.2} \\
MW $\uparrow$      & 34.0 & 85.6          & \textbf{93.0} &  6.8 & 40.4          & \textbf{63.8} \\
RotBond $\uparrow$ & 38.4 & 76.4          & \textbf{85.6} & 11.2 & 39.2          & \textbf{43.6} \\
SA $\downarrow$    &  2.4 & 82.8          & \textbf{95.8} &  1.0 & \textbf{62.8} & 55.6          \\
HBA $\uparrow$     & 40.0 & \textbf{82.6} & 77.4          & 19.0 & 48.4          & \textbf{56.0} \\
HBD $\uparrow$     & 45.8 & \textbf{84.2} & 74.2          & 26.8 & \textbf{58.2} & 55.2          \\
\midrule
\textbf{Avg}       & 32.8 & 83.8          & \textbf{86.7} & 12.7 & 51.1          & \textbf{55.7} \\
\bottomrule
\end{tabular}}
\caption{Comparison with external baselines on six overlapping properties. REINVENT4 and MolEditRL results are from published papers on the same 500-molecule test set. +\method reports the best result across our four base models for each property. Bold = best per property.}
\label{tab:external}
\end{table}

As shown in Table~\ref{tab:external}, \method outperforms MolEditRL on 4/6 properties at both thresholds and achieves higher averages (86.7 vs.\ 83.8 at $\tau\!=\!0.15$; 55.7 vs.\ 51.1 at $\tau\!=\!0.65$), despite being a training-free plug-and-play module. The gains are largest on MW at $\tau\!=\!0.65$ (+23.4 points). MolEditRL retains an edge on counting properties HBA and HBD at $\tau\!=\!0.15$, where its end-to-end RL optimization may better capture discrete structural changes.

\subsection{Ablation: Steering Direction}
\label{sec:ablation_direction}

To answer \textbf{RQ2}, we compare \method's gradient-based directions against simpler alternatives: \textbf{SFT} (no steering), a unit-norm \textbf{Random} vector, and \textbf{CAA}~\citep{rimsky2024steering} (mean-difference between high- and low-property activations). All steered methods use the same per-property strength $\alpha$. Table~\ref{tab:ablation_direction} shows per-property results on DrugAssist and Mistral.

\begin{table*}[t]
\centering
\resizebox{\linewidth}{!}{
\begin{tabular}{l cccc cccc c cccc cccc}
\toprule
& \multicolumn{8}{c}{\textbf{DrugAssist (Layer 3)}} && \multicolumn{8}{c}{\textbf{GeLLM3O-Mistral (Layer 14)}} \\
\cmidrule(lr){2-9} \cmidrule(lr){11-18}
& \multicolumn{4}{c}{Acc@0.15} & \multicolumn{4}{c}{Acc@0.65} && \multicolumn{4}{c}{Acc@0.15} & \multicolumn{4}{c}{Acc@0.65} \\
\cmidrule(lr){2-5} \cmidrule(lr){6-9} \cmidrule(lr){11-14} \cmidrule(lr){15-18}
& SFT & Rand. & CAA & \method & SFT & Rand. & CAA & \method && SFT & Rand. & CAA & \method & SFT & Rand. & CAA & \method \\
\midrule
QED     & 77.0 & 76.8 & 76.8 & \textbf{77.6} & 60.4 & 61.0 & 61.6 & \textbf{61.8} && 71.0 & 96.4 & 96.6 & \textbf{98.2} & 52.2 & 47.6 & 47.2 & \textbf{48.6} \\
DRD2    & 47.6 & 48.0 & 47.0 & \textbf{50.6} & 35.8 & 36.4 & 37.4 & \textbf{42.6} && 52.4 & 87.2 & 88.0 & \textbf{90.8} & 40.2 & 40.6 & 39.2 & \textbf{47.2} \\
logP    & 55.8 & 55.4 & 56.4 & \textbf{61.2} & 43.4 & 43.8 & 43.0 & \textbf{51.4} && 58.0 & 88.8 & 86.4 & \textbf{94.4} & 44.0 & 56.6 & 49.6 & \textbf{60.2} \\
MW      & 41.8 & 43.6 & 43.8 & \textbf{46.8} & 32.2 & 31.6 & 30.2 & \textbf{38.6} && 51.4 & 86.8 & 85.8 & \textbf{93.0} & 37.0 & 57.8 & 55.0 & \textbf{63.8} \\
RotBond & 36.4 & 36.8 & 37.2 & \textbf{41.2} & 27.0 & 27.8 & 26.0 & \textbf{33.8} && 38.6 & 50.2 & 50.4 & \textbf{75.0} & 29.6 & 28.2 & 24.8 & \textbf{43.6} \\
SA      & 67.6 & 67.8 & \textbf{69.2} & 67.6 & 54.8 & 55.6 & \textbf{57.2} & 55.6 && 65.4 & \textbf{97.4} & 97.0 & 95.8 & 49.8 & 53.0 & 51.0 & \textbf{55.4} \\
HBA     & 50.6 & 61.0 & 59.6 & \textbf{77.4} & 38.4 & 43.8 & 37.6 & \textbf{56.0} && 49.4 & 33.8 & 39.4 & \textbf{54.2} & 37.0 & 18.8 & 21.4 & \textbf{30.6} \\
HBD     & 49.6 & 63.8 & 63.0 & \textbf{74.2} & 37.2 & 47.2 & 48.4 & \textbf{55.2} && 37.4 & 28.4 & 19.8 & \textbf{47.4} & 25.4 & 16.6 & 13.2 & \textbf{30.0} \\
\midrule
\textbf{Avg} & 53.3 & 54.2 & 54.5 & \textbf{62.1} & 41.4 & 43.4 & 42.7 & \textbf{49.4} && 52.9 & 71.1 & 70.4 & \textbf{81.1} & 39.4 & 39.9 & 37.7 & \textbf{47.4} \\
\bottomrule
\end{tabular}}
\caption{Steering direction ablation on DrugAssist and Mistral. Four steering directions compared: no steering (SFT), random unit vector (Rand.), contrastive activation addition (CAA), and \method. All steered methods use the same per-property strength $\alpha$. Bold = best per row.}
\label{tab:ablation_direction}
\end{table*}

As shown in Table~\ref{tab:ablation_direction}, on DrugAssist, Random and CAA are indistinguishable from SFT (avg $\Delta < 2$ at both thresholds), confirming that mean-difference directions fail to capture property-relevant steering information. On Mistral, Random and CAA inflate $\accat{0.15}$ for continuous properties (logP +30.8, MW +35.4) but \textit{decrease} it for counting properties (HBA $-$15.6, HBD $-$17.6 with CAA), revealing that mean-difference directions trade property gain for structural disruption. \method is the only method that improves $\accat{0.65}$ on both models (DA +8.0, Mi +8.0 vs.\ CAA's +1.3 and $-$1.7), demonstrating that gradient-based directions preserve molecular similarity while increasing target properties.

\subsection{Ablation: SAE Training Objective}
\label{sec:ablation_sae}

To answer \textbf{RQ3}, we compare \textbf{Vanilla SAE} ($\mathcal{L}_\text{recon}$ only) and our full \textbf{Task-Oriented SAE} (Eq.~\ref{eq:sae_loss}), both using the same pre-computed gradient direction. The key question is whether the SAE basis must be aligned with property-relevant axes, or whether reconstruction alone suffices. Table~\ref{tab:ablation_sae} shows per-property results.

\begin{table*}[t]
\centering
\resizebox{\linewidth}{!}{
\begin{tabular}{l ccc ccc c ccc ccc}
\toprule
& \multicolumn{6}{c}{\textbf{DrugAssist (Layer 3)}} && \multicolumn{6}{c}{\textbf{GeLLM3O-Mistral (Layer 14)}} \\
\cmidrule(lr){2-7} \cmidrule(lr){9-14}
& \multicolumn{3}{c}{Acc@0.15} & \multicolumn{3}{c}{Acc@0.65} && \multicolumn{3}{c}{Acc@0.15} & \multicolumn{3}{c}{Acc@0.65} \\
\cmidrule(lr){2-4} \cmidrule(lr){5-7} \cmidrule(lr){9-11} \cmidrule(lr){12-14}
& SFT & Van. & \method & SFT & Van. & \method && SFT & Van. & \method & SFT & Van. & \method \\
\midrule
QED     & 77.0 & 77.4 & \textbf{77.6} & 60.4 & \textbf{62.0} & 61.8 && 71.0 & 97.2 & \textbf{98.2} & 52.2 & 50.6 & \textbf{48.6} \\
DRD2    & 47.6 & 47.8 & \textbf{50.6} & 35.8 & 38.4 & \textbf{42.6} && 52.4 & 90.6 & \textbf{90.8} & 40.2 & 45.0 & \textbf{47.2} \\
logP    & 55.8 & 55.4 & \textbf{61.2} & 43.4 & 45.8 & \textbf{51.4} && 58.0 & 91.4 & \textbf{94.4} & 44.0 & 61.2 & \textbf{60.2} \\
MW      & 41.8 & 44.0 & \textbf{46.8} & 32.2 & 35.8 & \textbf{38.6} && 51.4 & 90.8 & \textbf{93.0} & 37.0 & 60.2 & \textbf{63.8} \\
RotBond & 36.4 & 38.8 & \textbf{41.2} & 27.0 & 30.4 & \textbf{33.8} && 38.6 & 63.8 & \textbf{75.0} & 29.6 & 36.6 & \textbf{43.6} \\
SA      & 67.6 & 67.2 & 67.6          & 54.8 & \textbf{56.6} & 55.6 && 65.4 & \textbf{96.2} & 95.8 & 49.8 & 54.4 & \textbf{55.4} \\
HBA     & 50.6 & 68.2 & \textbf{77.4} & 38.4 & 48.2 & \textbf{56.0} && 49.4 & \textbf{55.4} & 54.2 & 37.0 & \textbf{31.0} & 30.6 \\
HBD     & 49.6 & 66.8 & \textbf{74.2} & 37.2 & 50.4 & \textbf{55.2} && 37.4 & 39.4 & \textbf{47.4} & 25.4 & 26.4 & \textbf{30.0} \\
\midrule
\textbf{Avg} & 53.3 & 55.7 & \textbf{62.1} & 41.4 & 45.9 & \textbf{49.4} && 52.9 & 78.1 & \textbf{81.1} & 39.4 & 45.7 & \textbf{47.4} \\
\bottomrule
\end{tabular}}
\caption{SAE training objective ablation on DrugAssist and Mistral. Three settings compared: no steering (SFT), steering with vanilla SAE ($\mathcal{L}_\text{recon}$ only), and steering with task-oriented SAE (Eq.~\ref{eq:sae_loss}). All use the same gradient direction. Bold = best per row.}
\label{tab:ablation_sae}
\end{table*}

As shown in Table~\ref{tab:ablation_sae}, a consistent ordering emerges: \textbf{SFT $<$ Vanilla SAE $<$ \method}. Vanilla SAE already provides partial gains (DA: +2.4, Mi: +25.2 at $\accat{0.15}$), suggesting that even a reconstruction-only sparse bottleneck filters noise from the gradient direction. The task-oriented training adds further improvement (DA: +6.4, Mi: +3.0 over Vanilla at $\accat{0.15}$), with the largest gains on counting properties: HBA improves by +9.2 over Vanilla on DA, and RotBond by +11.2 on Mi. The cosine similarity between Vanilla and task-oriented directions is near zero ($< 0.05$ for DA, $< 0.12$ for Mi), confirming that our losses fundamentally reorganize the latent basis rather than merely refining it.

\paragraph{Gradient alignment loss.}
We further isolate the contribution of the gradient alignment loss $\mathcal{L}_\text{grad}$ by training a variant with $\lambda_\text{grad}=0$ (denoted \textbf{w/o $\mathcal{L}_\text{grad}$}), retaining all other task-oriented losses ($\mathcal{L}_\text{recon}$, $\mathcal{L}_\text{contrastive}$, $\mathcal{L}_\text{supervised}$). Table~\ref{tab:ablation_grad} shows results on Mistral and MolGen.

\begin{table*}[t]
\centering
\resizebox{\linewidth}{!}{
\begin{tabular}{l ccc ccc c ccc ccc}
\toprule
& \multicolumn{6}{c}{\textbf{GeLLM3O-Mistral (Layer 14)}} && \multicolumn{6}{c}{\textbf{MolGen-Large (Layer 5)}} \\
\cmidrule(lr){2-7} \cmidrule(lr){9-14}
& \multicolumn{3}{c}{Acc@0.15} & \multicolumn{3}{c}{Acc@0.65} && \multicolumn{3}{c}{Acc@0.15} & \multicolumn{3}{c}{Acc@0.65} \\
\cmidrule(lr){2-4} \cmidrule(lr){5-7} \cmidrule(lr){9-11} \cmidrule(lr){12-14}
& SFT & w/o $\mathcal{L}_\text{grad}$ & \method & SFT & w/o $\mathcal{L}_\text{grad}$ & \method && SFT & w/o $\mathcal{L}_\text{grad}$ & \method & SFT & w/o $\mathcal{L}_\text{grad}$ & \method \\
\midrule
QED     & 71.0 & 96.4 & \textbf{98.2} & 52.2 & 47.6 & \textbf{48.6} && 42.4 & 74.8 & \textbf{76.2} & 19.4 & 34.0 & \textbf{36.6} \\
DRD2    & 52.4 & 88.8 & \textbf{90.8} & 40.2 & 44.6 & \textbf{47.2} && 37.2 & 79.0 & \textbf{83.6} & 23.0 & 31.2 & \textbf{33.4} \\
logP    & 58.0 & 89.6 & \textbf{94.4} & 44.0 & 55.8 & \textbf{60.2} && 46.2 & 78.0 & \textbf{86.4} & 27.2 & 37.8 & \textbf{46.2} \\
MW      & 51.4 & 87.4 & \textbf{93.0} & 37.0 & 58.2 & \textbf{63.8} && 48.2 & 83.6 & \textbf{87.0} & 22.6 & 43.8 & \textbf{46.4} \\
RotBond & 38.6 & 58.0 & \textbf{75.0} & 29.6 & 30.4 & \textbf{43.6} && 43.2 & 75.2 & \textbf{85.6} & 18.0 & 25.4 & \textbf{35.4} \\
SA      & 65.4 & 95.2 & \textbf{95.8} & 49.8 & 53.8 & \textbf{55.4} && 44.4 & 46.8 & \textbf{49.4} & 25.2 & 30.4 & \textbf{31.2} \\
HBA     & 49.4 & 38.4 & \textbf{54.2} & 37.0 & 20.4 & \textbf{30.6} && 42.6 & 52.0 & \textbf{55.0} & 23.0 & 30.2 & \textbf{32.4} \\
HBD     & 37.4 & 28.2 & \textbf{47.4} & 25.4 & 16.6 & \textbf{30.0} && 33.8 & 39.6 & \textbf{40.8} & 16.8 & 18.8 & \textbf{18.2} \\
\midrule
\textbf{Avg} & 52.9 & 72.8 & \textbf{81.1} & 39.4 & 40.9 & \textbf{47.4} && 42.3 & 66.1 & \textbf{70.5} & 21.9 & 31.5 & \textbf{35.0} \\
\bottomrule
\end{tabular}}
\caption{Gradient alignment loss ablation on Mistral and MolGen. w/o $\mathcal{L}_\text{grad}$ removes gradient alignment while retaining $\mathcal{L}_\text{recon}$, $\mathcal{L}_\text{contrast}$, $\mathcal{L}_\text{sup}$. Bold = best per row.}
\label{tab:ablation_grad}
\end{table*}

The gradient alignment loss is essential for robust steering. On MolGen, a clean ordering \textbf{SFT $<$ w/o $\mathcal{L}_\text{grad}$ $<$ \method} holds across all 8 properties at $\accat{0.15}$, with $\mathcal{L}_\text{grad}$ contributing +4.4 average improvement. On Mistral, the pattern reveals a striking failure mode: without $\mathcal{L}_\text{grad}$, counting properties (HBA, HBD) \textit{degrade below SFT} ($-$11.0 and $-$9.2 at $\accat{0.15}$), even though continuous properties improve substantially. This indicates that $\mathcal{L}_\text{grad}$ specifically aligns the SAE basis with the gradient structure needed for discrete property control. At $\accat{0.65}$, the gap widens further: \method gains +6.5 (Mi) and +3.5 (MG) over the ablated variant, confirming that gradient alignment preserves molecular similarity during steering.

\subsection{Multi-Property Steering}
\label{sec:reverse_multi}

To answer \textbf{RQ4}, we test whether \method's sparse steering directions can be composed for multi-property optimization. Directions combine via vector addition: $\mathbf{d}_\text{multi} = \sum_p \alpha_p \cdot \mathbf{d}_\text{steer}^{(p)}$. Table~\ref{tab:multi_main} evaluates six property pairs on DrugAssist and Mistral, where a candidate is ``successful'' only if it improves \textit{all} target properties simultaneously with Tanimoto $\geq \tau$.

As shown in Table~\ref{tab:multi_main}, \method improves joint success on 5/6 pairs at both thresholds. The largest gain is on Mi HBA+HBD (+13.6 at $\accat{0.15}$, +6.0 at $\accat{0.65}$), counting properties that are individually difficult but benefit from complementary feature directions. DA logP+MW also shows strong improvement (+9.4/+4.0), as both properties share lipophilicity-related features. The composability of steering directions is a direct consequence of the sparse decomposition: because each property's direction activates a distinct subset of SAE features (Table~\ref{tab:feature_interp}), adding directions produces minimal interference.

\section{Case Study: Feature Interpretability}
\label{sec:analysis}

A key advantage of \method over black-box steering is that each SAE feature encodes a specific molecular property, not just a structural motif. Table~\ref{tab:feature_interp} quantifies this: for each property, the top Importance Gate feature strongly discriminates molecules by property value.
\begin{table}[t]
\centering
\small
\setlength{\tabcolsep}{3pt}
\begin{tabular}{ll cccc}
\toprule
& & \multicolumn{2}{c}{Acc@0.15} & \multicolumn{2}{c}{Acc@0.65} \\
\cmidrule(lr){3-4} \cmidrule(lr){5-6}
\textbf{Model} & \textbf{Properties} & SFT & +\method & SFT & +\method \\
\midrule
DA & logP+MW     & 42.2 & \textbf{51.6} & 38.8 & \textbf{42.8} \\
DA & QED+SA      & 63.8 & \textbf{64.4} & 54.2 & 54.2 \\
DA & DRD2+SA     & 16.6 & \textbf{17.4} & 13.4 & \textbf{15.0} \\
Mi & HBA+HBD     & 16.4 & \textbf{30.0} & 11.4 & \textbf{17.4} \\
Mi & QED+SA      & 91.4 & \textbf{93.4} & 75.8 & \textbf{77.4} \\
Mi & DRD2+SA     & 73.4 & \textbf{73.6} & 51.0 & \textbf{51.4} \\
\bottomrule
\end{tabular}
\caption{Multi-property steering via direction addition on DrugAssist and Mistral. Joint success = candidate improves \textit{all} target properties simultaneously with Tanimoto $\geq \tau$.}
\label{tab:multi_main}
\end{table}
\begin{table}[t]
\centering
\small
\setlength{\tabcolsep}{3pt}
\begin{tabular}{llrrrc}
\toprule
\textbf{Feat.} & \textbf{Prop.} & \textbf{Top-25\%} & \textbf{Bot-25\%} & $\boldsymbol{\Delta}$ & $\boldsymbol{\rho}$ \\
\midrule
F31553 & MW      & 574.0 & 323.1 & +250.9 & +0.93 \\
F5418  & QED     & 0.27  & 0.77  & $-$0.50 & $-$0.85 \\
F17369 & RotBond & 10.60 & 4.93  & +5.67  & +0.68 \\
F31602 & logP    & 6.25  & 2.63  & +3.62  & +0.65 \\
F18632 & HBA     & 6.66  & 3.45  & +3.21  & +0.63 \\
F17369 & SA      & 3.97  & 3.01  & +0.96  & +0.52 \\
\midrule
F4291  & HBD     & 2.07  & 0.98  & +1.10  & +0.32 \\
F7906  & DRD2    & 0.13  & 0.08  & +0.06  & +0.32 \\
\bottomrule
\end{tabular}
\caption{SAE feature interpretability (DrugAssist, Layer 3, 50K ZINC molecules). For each property, we report the best feature among the top-50 Importance Gate features ranked by $|\rho|$. Top/Bot-25\% = mean property value of molecules with highest/lowest 25\% feature activation. $\rho$ = Spearman rank correlation between feature activation and property value.}
\label{tab:feature_interp}
\end{table}

For six of eight properties, the best Importance Gate feature achieves $|\rho| \geq 0.52$ (Table~\ref{tab:feature_interp}). Feature F31553 ($\rho = +0.93$) encodes molecular weight with a 251 Da gap between high- and low-activation quartiles. Feature F5418 ($\rho = -0.85$) captures molecular complexity, cleanly separating drug-like (QED=0.77) from non-drug-like (QED=0.27) molecules. For counting properties (HBD) and binary properties (DRD2), single-feature correlations are weaker ($\rho \approx 0.32$), indicating multi-feature encoding, consistent with the composite steering directions that successfully improve these properties (Table~\ref{tab:main_results}).

To confirm that features are not merely correlational, we steer using individual SAE decoder columns ($\alpha = 2.0$, 500 test molecules). Single-feature steering produces targeted molecular changes: F3714 (logP-relevant) adds aromatic groups (+3.3 avg logP), F28089 (HBD-relevant) introduces NH groups (+1--2 donors), with 13--96 contrast pairs per feature where SFT fails but single-feature steering succeeds. Full molecule visualizations are in Appendix~\ref{sec:app_case}.

Figure~\ref{fig:case_examples} shows representative cases where SFT fails but \method succeeds, each reflecting a distinct failure mode of the SFT editor.

\textit{Property-reversing edit (logP).} The source molecule (logP=4.51) contains an isopentyl chain contributing to hydrophobicity. SFT removes this chain and replaces it with a hydroxyl group, \textit{decreasing} logP to 1.82, a structurally valid but property-reversing edit. \method instead replaces the methoxypyridine ring with a dibromophenoxy moiety, adding halogenated aromatic bulk that raises logP to 7.82 (+3.31) with higher structural similarity (Tanimoto 0.79 vs.\ 0.65).

\textit{Ineffective edit (HBA).} SFT removes a side chain but leaves acceptor count unchanged (5$\to$5). \method introduces a guanidinium-like group with multiple nitrogen and oxygen atoms, adding 7 new acceptor sites (5$\to$12) by activating SAE features associated with nitrogen-rich heterocycles.

\textit{Overly conservative edit (RotBond).} SFT modifies a substituent on the triazole ring but does not introduce any additional rotatable bonds (4$\to$4). \method instead extends the molecule with a flexible alkenyl-alkyl chain, adding 9 rotatable bonds (4$\to$13, Tanimoto 0.72), overcoming SFT's tendency toward rigid-scaffold-preserving edits.

These cases share a common pattern: SFT produces valid edits but lacks a mechanism to ensure they align with the target property. \method provides this directional signal through property-relevant SAE features.

\begin{figure}[t]
\centering
\setlength{\tabcolsep}{1pt}
\begin{tabular}{cccc}
& \textbf{Source} & \textbf{SFT} & \textbf{+\method} \\[2pt]
\rotatebox{90}{\small\textbf{logP$\uparrow$}} &
\includegraphics[height=1.6cm]{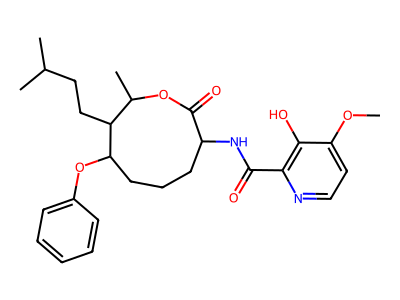} &
\includegraphics[height=1.6cm]{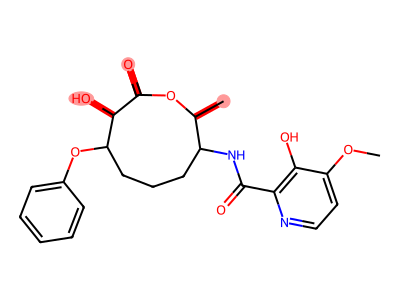} &
\includegraphics[height=1.6cm]{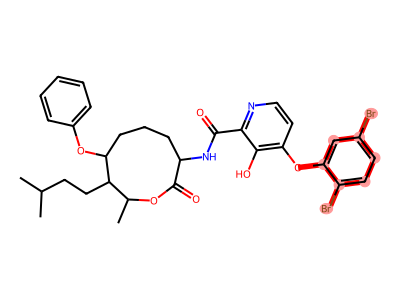} \\[-2pt]
& {\scriptsize 4.51} & {\scriptsize 1.82 (\textcolor{red}{$-$2.69})} & {\scriptsize \textbf{7.82} (\textcolor{green!50!black}{+3.31})} \\[4pt]
\rotatebox{90}{\small\textbf{HBA$\uparrow$}} &
\includegraphics[height=1.6cm]{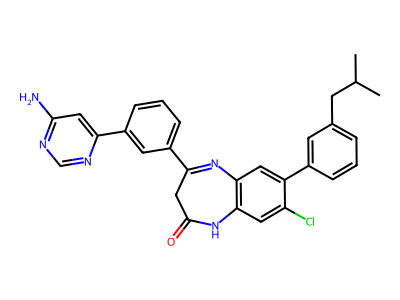} &
\includegraphics[height=1.6cm]{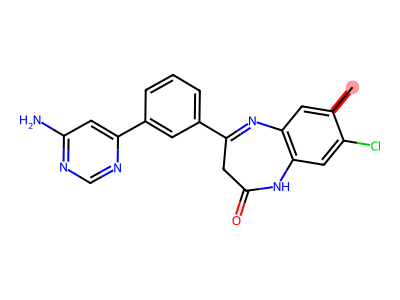} &
\includegraphics[height=1.6cm]{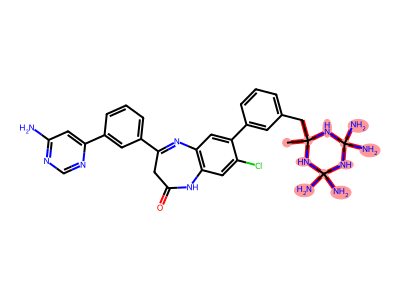} \\[-2pt]
& {\scriptsize 5} & {\scriptsize 5 (\textcolor{red}{$\pm$0})} & {\scriptsize \textbf{12} (\textcolor{green!50!black}{+7})} \\[4pt]
\rotatebox{90}{\small\textbf{Rot.$\uparrow$}} &
\includegraphics[height=1.6cm]{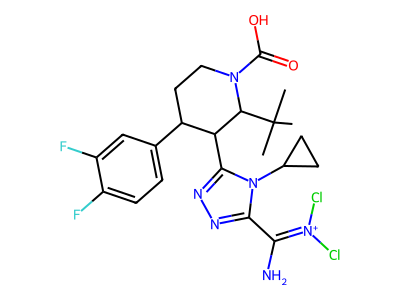} &
\includegraphics[height=1.6cm]{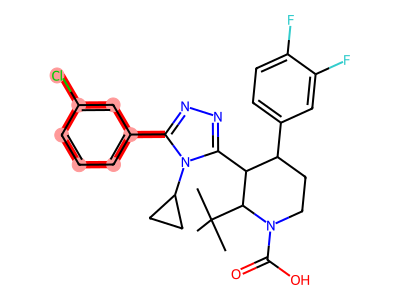} &
\includegraphics[height=1.6cm]{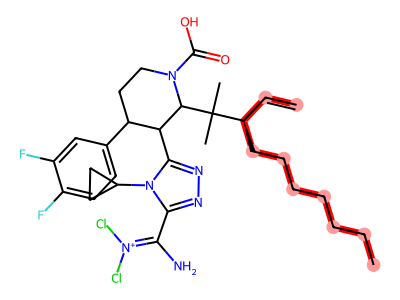} \\[-2pt]
& {\scriptsize 4} & {\scriptsize 4 (\textcolor{red}{$\pm$0})} & {\scriptsize \textbf{13} (\textcolor{green!50!black}{+9})} \\
\end{tabular}
\caption{Molecular editing examples on DrugAssist. Each row shows source, SFT output, and \method output for one property. Numbers below molecules indicate property values and changes ($\Delta$). Tanimoto similarities (SFT / \method): logP 0.65 / 0.79, HBA 0.69 / 0.75, RotBond 0.67 / 0.72.}
\label{fig:case_examples}
\end{figure}

\section{Conclusion}
\label{sec:conclusion}

We presented \method, a plug-and-play framework that decomposes the dense hidden states of molecular editing LLMs into sparse, property-aligned features, enabling training-free property control at inference time without modifying model parameters. Our task-oriented SAE, trained with contrastive, predictive, and gradient-alignment objectives alongside per-property Importance Gates, produces a sparse basis that is causally aligned with target properties rather than merely correlated. Experiments on the MolEditRL benchmark across four architectures and eight properties show that \method consistently improves property-directed editing accuracy, with gains of up to 42.4 percentage points. Case studies further reveal that the learned SAE features encode interpretable chemical semantics and that steering corrects specific failure modes of SFT editors, including property-reversing edits, ineffective modifications, and overly conservative structural changes.

\clearpage
\section*{Limitations}
\label{sec:limitations}

\method has several limitations. The steering strength and intervention layer require per-property and per-model tuning, adding setup cost before deployment. The SAE training pipeline introduces additional computational overhead beyond the base SFT model, though inference-time cost is negligible. Finally, our evaluation relies on computable property oracles, and the framework has not yet been validated on experimental assay endpoints.


\nocite{sterling2015zinc}
\bibliography{references}

\appendix

\section{Experimental Settings}
\label{sec:app_settings}

\paragraph{Evaluation Metrics.}
We follow the MolEditRL benchmark protocol~\citep{edwards2024moleditrl}. For each test molecule $x$ and property $p$, we generate $n=5$ candidate molecules $\{x'_1, \ldots, x'_n\}$ via nucleus sampling. A candidate $x'_i$ is considered \textit{successful} if it satisfies three conditions: (1) it is a chemically valid SMILES string (verified by RDKit~\citep{rdkit} parsing), (2) it improves the target property relative to the input ($\text{oracle}(x'_i) > \text{oracle}(x)$ for increase tasks, or $<$ for decrease tasks), and (3) its Tanimoto similarity to the input molecule meets or exceeds the threshold: $\text{Tanimoto}(x, x'_i) \geq \tau$. Tanimoto similarity is computed on Morgan fingerprints~\citep{rogers2010extended} with radius 2 and 2048 bits. We report $\accat{\tau}$: the percentage of test molecules for which at least one candidate is successful. We evaluate at two thresholds: $\tau = 0.15$ (permissive, allowing larger structural changes) and $\tau = 0.65$ (strict, requiring high structural preservation).

\paragraph{Test Set.}
We use the 500-molecule test set from the DrugAssist evaluation suite, which is a standardized subset also adopted by MolEditRL. These molecules span diverse drug-like chemical space from the DrugAssist training distribution.

\paragraph{Properties and Oracles.}
Table~\ref{tab:properties} describes all eight molecular properties and their computational oracles.

\begin{table}[t]
\centering
\small
\begin{tabular}{lp{3.3cm}c}
\toprule
\textbf{Property} & \textbf{Oracle} & \textbf{Dir.} \\
\midrule
QED & RDKit QED~\citep{bickerton2012quantifying} — quantitative estimate of drug-likeness & $\uparrow$ \\
DRD2 & Random Forest classifier on Morgan FP (radius 2, 2048 bits) & $\uparrow$ \\
logP & RDKit Crippen MolLogP — octanol-water partition coefficient & $\uparrow$ \\
MW & RDKit MolWt — molecular weight in Daltons & $\uparrow$ \\
RotBond & RDKit NumRotatableBonds & $\uparrow$ \\
SA & SA scorer~\citep{ertl2009estimation} — synthetic accessibility (lower = easier) & $\downarrow$ \\
HBA & RDKit NumHAcceptors — hydrogen bond acceptor count & $\uparrow$ \\
HBD & RDKit NumHDonors — hydrogen bond donor count & $\uparrow$ \\
\bottomrule
\end{tabular}
\caption{Molecular properties and oracles. Dir.\ indicates optimization direction. All oracles are deterministic and computed from molecular structure.}
\label{tab:properties}
\end{table}

\paragraph{Models.}
Table~\ref{tab:models} summarizes the four base models evaluated.

\begin{table}[t]
\centering
\small
\begin{tabular}{lccc}
\toprule
\textbf{Model} & \textbf{Architecture} & \textbf{Layer $l^*$} & \textbf{$d$} \\
\midrule
DrugAssist (DA) & LLaMA-2-7B & 3 & 4096 \\
GeLLM3O-L3 (L3) & LLaMA-3.1-8B & 8 & 4096 \\
GeLLM3O-Mi (Mi) & Mistral-7B & 14 & 4096 \\
MolGen (MG) & BART-Large & 5 & 1024 \\
\bottomrule
\end{tabular}
\caption{Models evaluated. Layer $l^*$ is selected by ridge probing (\S\ref{sec:layer_select}). $d$ = hidden dimension. SAE expansion = 8$\times$.}
\label{tab:models}
\end{table}

\paragraph{Hyperparameters.}
Table~\ref{tab:hyperparams} lists all key hyperparameters for each component.

\begin{table*}[t]
\centering
\small
\setlength{\tabcolsep}{4pt}
\resizebox{\textwidth}{!}{
\begin{tabular}{ll@{\hskip 12pt}ll@{\hskip 12pt}ll}
\toprule
\multicolumn{2}{l}{\textit{SFT (LoRA}~\citep{hu2022lora}\textit{)}} & \multicolumn{2}{l}{\textit{Task-Oriented SAE}} & \multicolumn{2}{l}{\textit{Gradient Direction}} \\
\cmidrule{1-2} \cmidrule{3-4} \cmidrule{5-6}
LoRA rank $r$ & 16 & Architecture & Gated SAE & Precision & fp32 \\
LoRA alpha $\alpha_\text{lora}$ & 32 & Expansion factor & 8$\times$ ($d \!\rightarrow\! 8d$) & Pairs per property & 5{,}000 \\
Target modules & q\_proj, v\_proj & Learning rate & 1e-3 (Adam) & Token aggregation & Mean all positions \\
Learning rate & 2e-4 (cosine) & Batch size & 256 & Per-example norm & L2 before avg \\
Batch size & 128 & Epochs & 50 & & \\
Epochs & 3 & Molecules & 50K (ZINC) & \multicolumn{2}{l}{\textit{Inference}} \\
\cmidrule{5-6}
Optimizer & AdamW & Decoder norm & Unit-norm cols & Candidates/molecule & 5 \\
Precision & bf16 & $\lambda_c$ (contrastive) & 0.1 & Temperature & 0.8 \\
& & $\lambda_s$ (supervised) & 0.1 & Top-$p$ (nucleus) & 0.95 \\
& & $\lambda_\text{sp}$ (sparsity) & 1e-3 & Max new tokens & 256 \\
& & $\lambda_g$ (gradient) & 0.5 & Thresholds $\tau$ & 0.15, 0.65 \\
& & Top-$k$ for steering & 200 & & \\
\bottomrule
\end{tabular}}
\caption{Full hyperparameter settings. Steering strength $\alpha$ is tuned per property on a held-out validation set; optimal values are listed in Table~\ref{tab:steering_strengths}.}
\label{tab:hyperparams}
\end{table*}

\paragraph{Steering Strength Selection.}
The steering strength $\alpha$ is tuned per property by evaluating on a held-out 100-molecule validation split. We search over $\alpha \in \{0.01, 0.05, 0.1, 0.2, 0.3, 0.5, 0.7, 1.0, 1.5, 2.0, 5.0, 10.0\}$ and select the $\alpha$ maximizing $\accat{0.15}$ on the validation set. Optimal values vary by model and property type: composite properties (QED, SA) prefer smaller strengths ($\alpha \leq 0.5$), while counting properties (HBA, HBD, RotBond) benefit from larger strengths ($\alpha \geq 1.0$). MolGen (BART, $d\!=\!1024$) requires $\sim$10$\times$ larger $\alpha$ than the 4096-dimensional causal LMs. Table~\ref{tab:steering_strengths} lists the selected values.

\begin{table}[t]
\centering
\small
\setlength{\tabcolsep}{3pt}
\begin{tabular}{l cccc}
\toprule
\textbf{Property} & \textbf{DA} & \textbf{L3} & \textbf{Mi} & \textbf{MG} \\
\midrule
QED      & 0.07 & 1.5 & 0.5 & 5.0 \\
DRD2     & 0.7  & 0.7 & 0.3 & 10.0 \\
logP     & 0.7  & 0.7 & 0.5 & 10.0 \\
MW       & 0.5  & 1.5 & 0.3 & 10.0 \\
RotBond  & 1.0  & 1.0 & 1.0 & 10.0 \\
SA       & 0.2  & 0.1 & 0.1 & 5.0 \\
HBA      & 2.0  & 1.0 & 1.0 & 10.0 \\
HBD      & 2.0  & 2.0 & 1.0 & 10.0 \\
\bottomrule
\end{tabular}
\caption{Per-property steering strength $\alpha$ selected on validation set. Composite properties (QED, SA) use small $\alpha$; counting properties (HBA, HBD) use large $\alpha$. MolGen ($d\!=\!1024$) requires larger $\alpha$ than 4096-dim models.}
\label{tab:steering_strengths}
\end{table}

\paragraph{Contrastive Loss Details.}
The InfoNCE contrastive loss $\mathcal{L}_\text{contrast}$ operates per property. For each property $p$, molecules in a training batch are ranked by their oracle-computed property value. The top 25\% molecules form the positive set and the bottom 25\% form the negative set. For each positive anchor, its importance-gated code $\mathbf{z}_p = \mathbf{w}_p \odot \mathbf{z}$ is projected through a learned 2-layer MLP to a 128-dimensional embedding space. The InfoNCE loss with temperature $\tau_c = 0.07$ is:
\begin{equation}
\mathcal{L}_\text{contrast}^{(p)} = -\!\!\!\!\sum_{i \in \text{pos}} \log \frac{\sum_{j \in \text{pos}, j \neq i} \exp(\text{sim}_{ij} / \tau_c)}{\sum_{k \neq i} \exp(\text{sim}_{ik} / \tau_c)}
\end{equation}
where $\text{sim}_{ij}$ is the cosine similarity between the projected embeddings of molecules $i$ and $j$. The total contrastive loss averages over all properties: $\mathcal{L}_\text{contrast} = \frac{1}{|P|} \sum_p \mathcal{L}_\text{contrast}^{(p)}$.

\paragraph{Supervised Loss Details.}
The supervised loss $\mathcal{L}_\text{sup}$ uses a per-property 2-layer MLP (hidden dim 256, ReLU activation) that takes the importance-gated code $\mathbf{z}_p$ as input and predicts the normalized property value. The loss is MSE averaged over properties: $\mathcal{L}_\text{sup} = \frac{1}{|P|} \sum_p \text{MSE}(\hat{y}_p, y_p)$.

\paragraph{Instruction Templates.}
For causal LMs (DrugAssist, LLaMA-3, Mistral), we use the LLaMA-2 chat format:

\smallskip
\noindent\texttt{<s>[INST] Modify this molecule to \{increase/decrease\} its \{property description\}: \{SMILES\} [/INST]}
\smallskip

\noindent Property descriptions use full natural language names to match SFT training prompts:

\begin{table}[t]
\centering
\small
\begin{tabular}{ll}
\toprule
\textbf{Property} & \textbf{Prompt Description} \\
\midrule
QED & ``drug-likeness (QED)'' \\
DRD2 & ``DRD2 binding activity'' \\
logP & ``lipophilicity (logP)'' \\
MW & ``molecular weight'' \\
RotBond & ``number of rotatable bonds'' \\
SA & ``synthetic accessibility'' \\
HBA & ``number of hydrogen bond acceptors'' \\
HBD & ``number of hydrogen bond donors'' \\
\bottomrule
\end{tabular}
\caption{Property descriptions used in instruction prompts.}
\label{tab:prompts}
\end{table}

For MolGen (BART encoder-decoder), the input is the source SMILES directly, and the model generates the target SMILES via beam search (beam width 5).

\paragraph{Computational Infrastructure.}
All experiments were conducted on NVIDIA A100 80GB GPUs. SFT training uses 2 GPUs with DeepSpeed ZeRO-2. SAE training and evaluation use a single GPU. Gradient direction computation requires fp32 precision and takes approximately 2 hours per model per property on a single A100. SAE training converges in approximately 30 minutes. Inference-time steering adds negligible overhead ($<$1\% latency increase) as it only involves a single vector addition per forward pass.


\section{Case Studies}
\label{sec:app_case}

\subsection{Single-Feature Steering}
\label{sec:app_single_feature}

Tables~\ref{tab:case_logp}--\ref{tab:case_acceptor} show molecular editing examples using \textit{single SAE feature} steering, demonstrating that individual features capture interpretable chemical semantics.
All cases use the DrugAssist model with Layer-3 SAE steering.
The instruction prompt for each property group is shown above the table; \texttt{\{SMILES\}} is replaced with the source molecule at inference time.

\begin{table*}[ht]
\centering
\small
\texttt{[INST] Modify this molecule to increase its lipophilicity (logP): \{SMILES\} [/INST]}\\[4pt]
{\scriptsize SAE Feature F3714, semantics: phenyl 94\%, halogen 79\%, logP$\uparrow$165\%}\\[6pt]
\setlength{\tabcolsep}{3pt}
\begin{tabular}{c c c m{5.2cm}}
\toprule
\textbf{Source} & \textbf{SFT} & \textbf{+\method} & \textbf{Property / Similarity} \\
\midrule
\includegraphics[height=1.8cm]{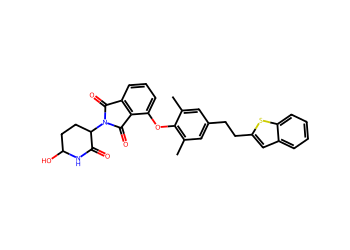} & \includegraphics[height=1.8cm]{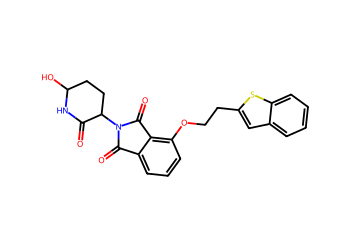} & \includegraphics[height=1.8cm]{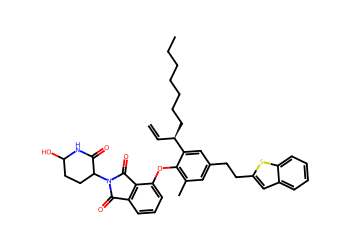} & SFT: 2.72 ($\Delta${-}2.57, \textcolor{red}{\ding{55}}) Sim 0.72 \newline \method: 8.61 ($\Delta$+3.32, \textcolor{green!50!black}{\ding{51}}) Sim 0.75 \\[2pt]
\includegraphics[height=1.8cm]{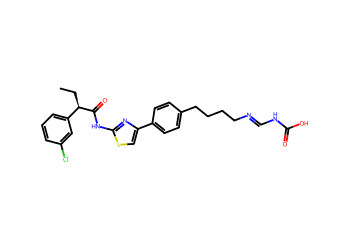} & \includegraphics[height=1.8cm]{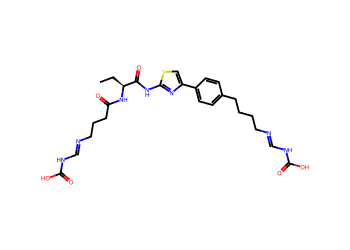} & \includegraphics[height=1.8cm]{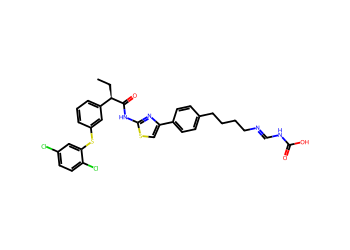} & SFT: 3.34 ($\Delta${-}2.88, \textcolor{red}{\ding{55}}) Sim 0.68 \newline \method: 9.02 ($\Delta$+2.81, \textcolor{green!50!black}{\ding{51}}) Sim 0.81 \\[2pt]
\includegraphics[height=1.8cm]{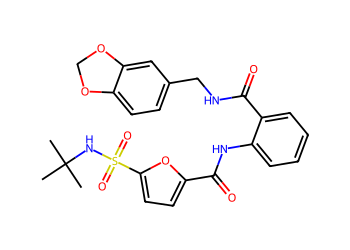} & \includegraphics[height=1.8cm]{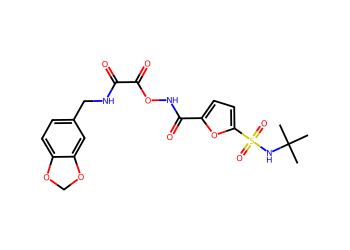} & \includegraphics[height=1.8cm]{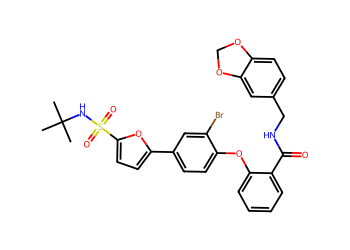} & SFT: 0.59 ($\Delta${-}2.68, \textcolor{red}{\ding{55}}) Sim 0.65 \newline \method: 6.24 ($\Delta$+2.97, \textcolor{green!50!black}{\ding{51}}) Sim 0.62 \\[2pt]
\includegraphics[height=1.8cm]{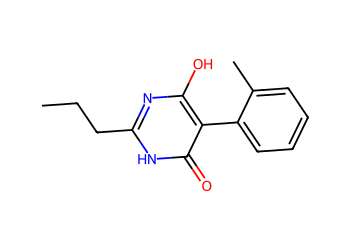} & \includegraphics[height=1.8cm]{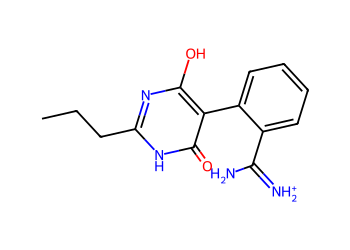} & \includegraphics[height=1.8cm]{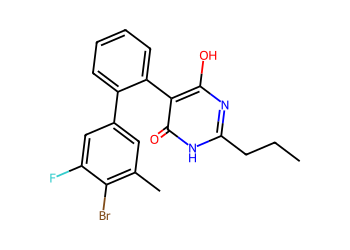} & SFT: {-}0.44 ($\Delta${-}2.84, \textcolor{red}{\ding{55}}) Sim 0.68 \newline \method: 4.97 ($\Delta$+2.57, \textcolor{green!50!black}{\ding{51}}) Sim 0.64 \\[2pt]
\includegraphics[height=1.8cm]{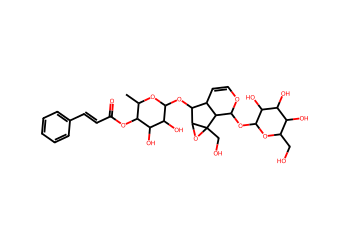} & \includegraphics[height=1.8cm]{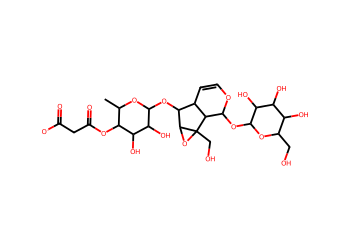} & \includegraphics[height=1.8cm]{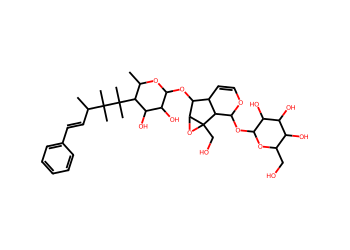} & SFT: {-}6.05 ($\Delta${-}3.57, \textcolor{red}{\ding{55}}) Sim 0.69 \newline \method: 0.92 ($\Delta$+3.39, \textcolor{green!50!black}{\ding{51}}) Sim 0.67 \\
\bottomrule
\end{tabular}
\caption{Case study: steering logP with Feature F3714. Source logP values: 5.29, 6.21, 3.27, 2.40, {-}2.47. SFT \textit{decreases} logP in every case; \method steering \textit{increases} it by +2.6 to +3.4.}
\label{tab:case_logp}
\end{table*}

\begin{table*}[ht]
\centering
\small
\texttt{[INST] Modify this molecule to increase its drug-likeness (QED): \{SMILES\} [/INST]}\\[4pt]
{\scriptsize SAE Feature F4829, semantics: compact heterocyclic scaffolds, MW$\sim$200, QED$\uparrow$, MW$\downarrow$95\%}\\[6pt]
\setlength{\tabcolsep}{3pt}
\begin{tabular}{c c c m{5.2cm}}
\toprule
\textbf{Source} & \textbf{SFT} & \textbf{+\method} & \textbf{Property / Similarity} \\
\midrule
\includegraphics[height=1.8cm]{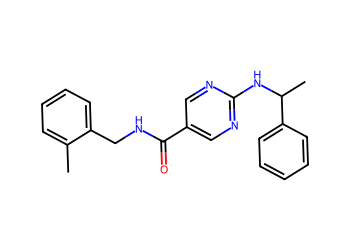} & \includegraphics[height=1.8cm]{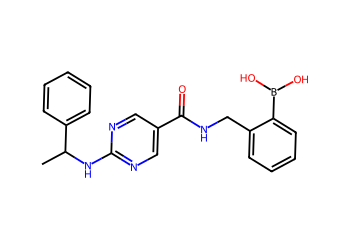} & \includegraphics[height=1.8cm]{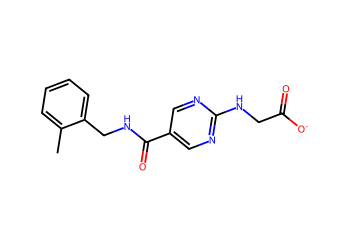} & SFT: 0.46 ($\Delta${-}0.25, \textcolor{red}{\ding{55}}) Sim 0.76 \newline \method: 0.77 ($\Delta$+0.06, \textcolor{green!50!black}{\ding{51}}) Sim 0.62 \\[2pt]
\includegraphics[height=1.8cm]{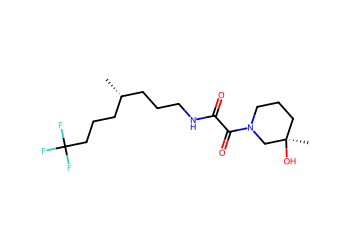} & \includegraphics[height=1.8cm]{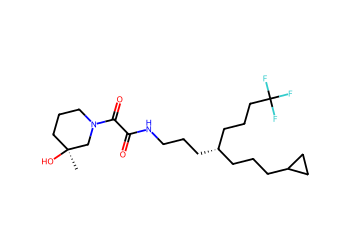} & \includegraphics[height=1.8cm]{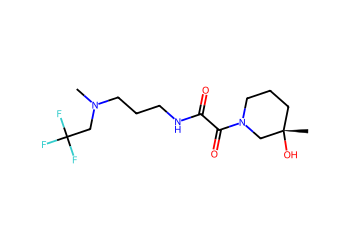} & SFT: 0.38 ($\Delta${-}0.16, \textcolor{red}{\ding{55}}) Sim 0.75 \newline \method: 0.56 ($\Delta$+0.02, \textcolor{green!50!black}{\ding{51}}) Sim 0.64 \\[2pt]
\includegraphics[height=1.8cm]{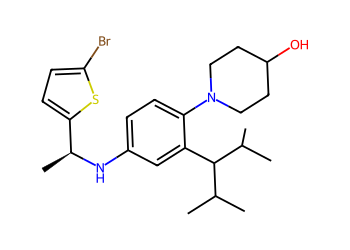} & \includegraphics[height=1.8cm]{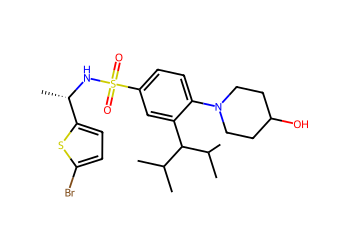} & \includegraphics[height=1.8cm]{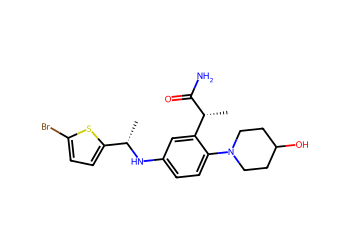} & SFT: 0.44 ($\Delta${-}0.01, \textcolor{red}{\ding{55}}) Sim 0.69 \newline \method: 0.61 ($\Delta$+0.17, \textcolor{green!50!black}{\ding{51}}) Sim 0.76 \\[2pt]
\includegraphics[height=1.8cm]{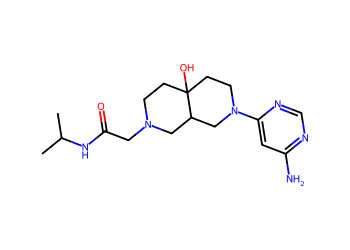} & \includegraphics[height=1.8cm]{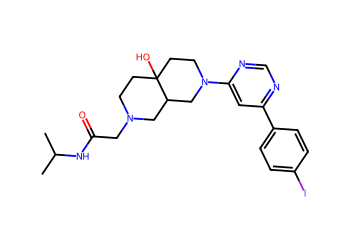} & \includegraphics[height=1.8cm]{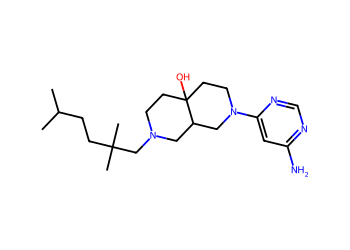} & SFT: 0.57 ($\Delta${-}0.12, \textcolor{red}{\ding{55}}) Sim 0.71 \newline \method: 0.80 ($\Delta$+0.10, \textcolor{green!50!black}{\ding{51}}) Sim 0.65 \\[2pt]
\includegraphics[height=1.8cm]{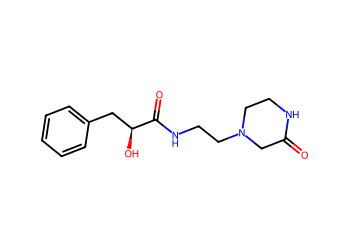} & \includegraphics[height=1.8cm]{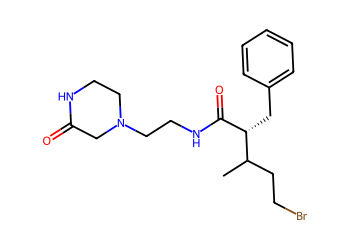} & \includegraphics[height=1.8cm]{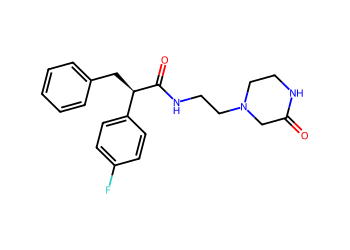} & SFT: 0.61 ($\Delta${-}0.03, \textcolor{red}{\ding{55}}) Sim 0.68 \newline \method: 0.78 ($\Delta$+0.14, \textcolor{green!50!black}{\ding{51}}) Sim 0.72 \\
\bottomrule
\end{tabular}
\caption{Case study: steering QED with Feature F4829. Source QED values: 0.71, 0.54, 0.45, 0.69, 0.64. SFT \textit{decreases} QED in every case; \method steering \textit{increases} it by +0.02 to +0.17.}
\label{tab:case_qed}
\end{table*}

\begin{table*}[ht]
\centering
\small
\texttt{[INST] Modify this molecule to increase its number of hydrogen bond donors: \{SMILES\} [/INST]}\\[4pt]
{\scriptsize SAE Feature F28089, semantics: NH 72\%, phenyl only 40\%, low logP}\\[6pt]
\setlength{\tabcolsep}{3pt}
\begin{tabular}{c c c m{5.2cm}}
\toprule
\textbf{Source} & \textbf{SFT} & \textbf{+\method} & \textbf{Property / Similarity} \\
\midrule
\includegraphics[height=1.8cm]{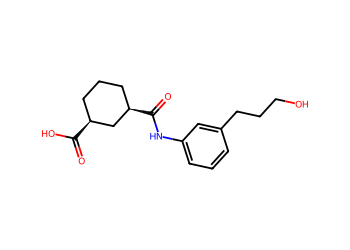} & \includegraphics[height=1.8cm]{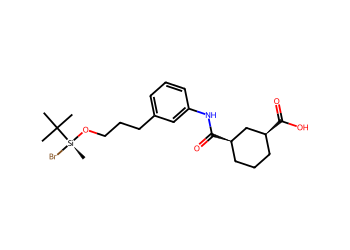} & \includegraphics[height=1.8cm]{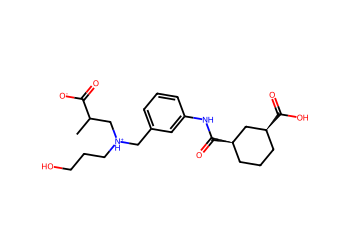} & SFT: 2 ($\Delta${-}1, \textcolor{red}{\ding{55}}) Sim 0.65 \newline \method: 4 ($\Delta$+1, \textcolor{green!50!black}{\ding{51}}) Sim 0.64 \\[2pt]
\includegraphics[height=1.8cm]{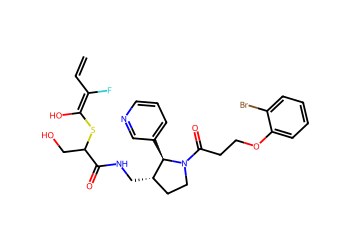} & \includegraphics[height=1.8cm]{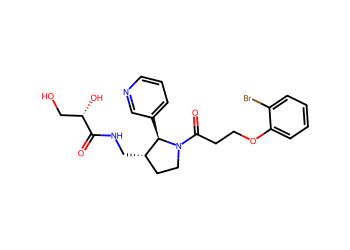} & \includegraphics[height=1.8cm]{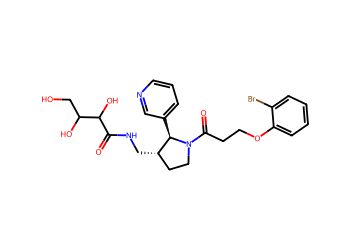} & SFT: 3 ($\Delta$0, \textcolor{red}{\ding{55}}) Sim 0.71 \newline \method: 4 ($\Delta$+1, \textcolor{green!50!black}{\ding{51}}) Sim 0.69 \\[2pt]
\includegraphics[height=1.8cm]{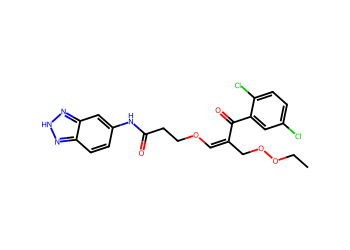} & \includegraphics[height=1.8cm]{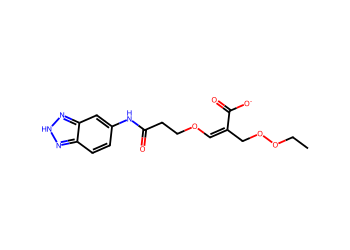} & \includegraphics[height=1.8cm]{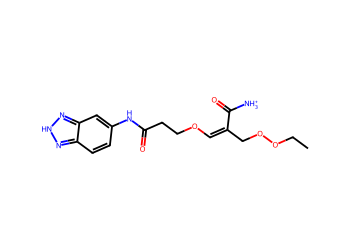} & SFT: 2 ($\Delta$0, \textcolor{red}{\ding{55}}) Sim 0.71 \newline \method: 3 ($\Delta$+1, \textcolor{green!50!black}{\ding{51}}) Sim 0.71 \\[2pt]
\includegraphics[height=1.8cm]{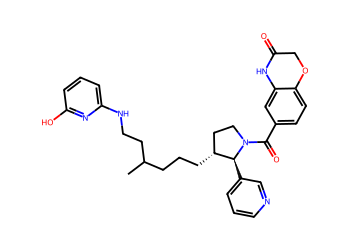} & \includegraphics[height=1.8cm]{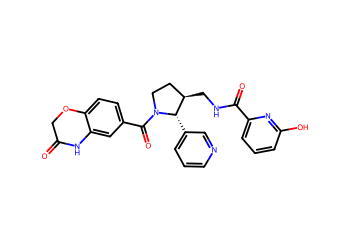} & \includegraphics[height=1.8cm]{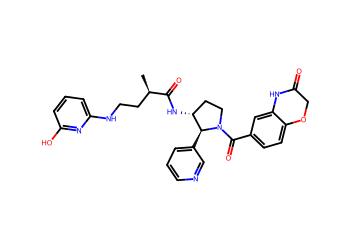} & SFT: 3 ($\Delta$0, \textcolor{red}{\ding{55}}) Sim 0.66 \newline \method: 4 ($\Delta$+1, \textcolor{green!50!black}{\ding{51}}) Sim 0.77 \\[2pt]
\includegraphics[height=1.8cm]{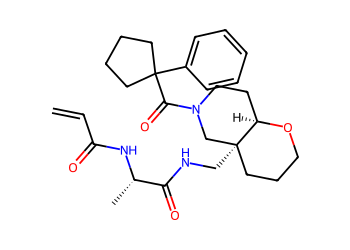} & \includegraphics[height=1.8cm]{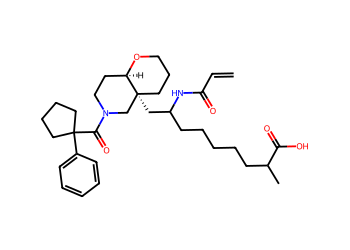} & \includegraphics[height=1.8cm]{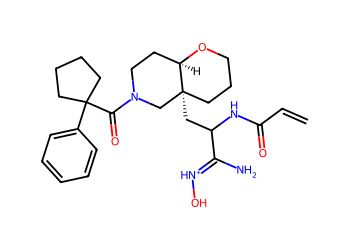} & SFT: 2 ($\Delta$0, \textcolor{red}{\ding{55}}) Sim 0.68 \newline \method: 4 ($\Delta$+2, \textcolor{green!50!black}{\ding{51}}) Sim 0.67 \\
\bottomrule
\end{tabular}
\caption{Case study: steering H-bond donors with Feature F28089. Source HBD values: 3, 3, 2, 3, 2. SFT fails to increase donors (unchanged or decreased); \method steering adds +1 to +2.}
\label{tab:case_donor}
\end{table*}

\begin{table*}[ht]
\centering
\small
\texttt{[INST] Modify this molecule to increase its number of hydrogen bond acceptors: \{SMILES\} [/INST]}\\[4pt]
{\scriptsize SAE Feature F18533, semantics: piperidine 33\%, ester 26\%, phenyl 84\%}\\[6pt]
\setlength{\tabcolsep}{3pt}
\begin{tabular}{c c c m{5.2cm}}
\toprule
\textbf{Source} & \textbf{SFT} & \textbf{+\method} & \textbf{Property / Similarity} \\
\midrule
\includegraphics[height=1.8cm]{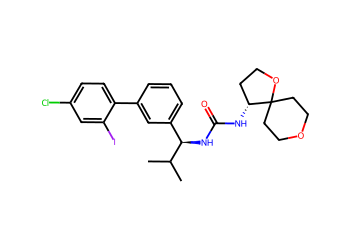} & \includegraphics[height=1.8cm]{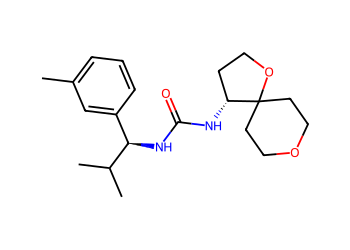} & \includegraphics[height=1.8cm]{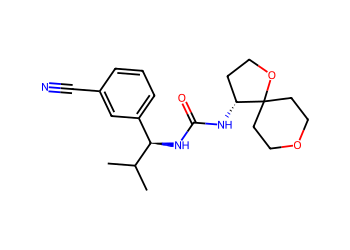} & SFT: 3 ($\Delta$0, \textcolor{red}{\ding{55}}) Sim 0.67 \newline \method: 4 ($\Delta$+1, \textcolor{green!50!black}{\ding{51}}) Sim 0.63 \\[2pt]
\includegraphics[height=1.8cm]{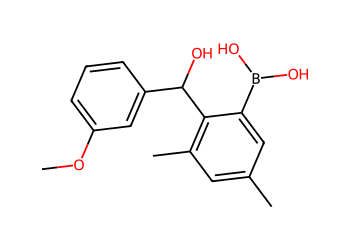} & \includegraphics[height=1.8cm]{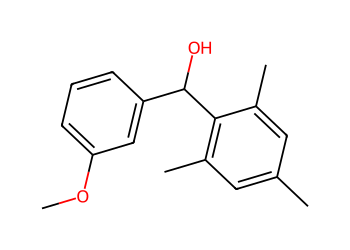} & \includegraphics[height=1.8cm]{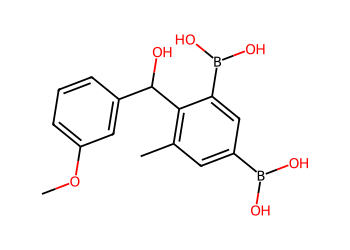} & SFT: 2 ($\Delta${-}2, \textcolor{red}{\ding{55}}) Sim 0.79 \newline \method: 6 ($\Delta$+2, \textcolor{green!50!black}{\ding{51}}) Sim 0.79 \\[2pt]
\includegraphics[height=1.8cm]{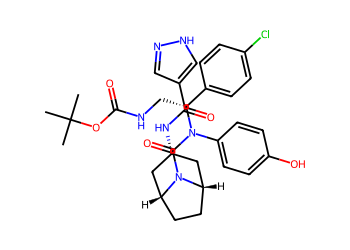} & \includegraphics[height=1.8cm]{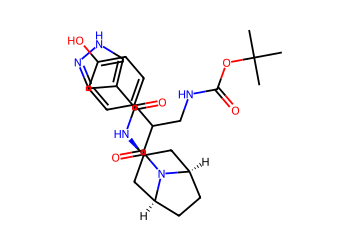} & \includegraphics[height=1.8cm]{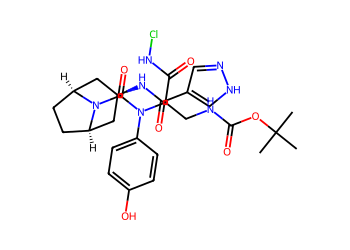} & SFT: 6 ($\Delta$0, \textcolor{red}{\ding{55}}) Sim 0.68 \newline \method: 7 ($\Delta$+1, \textcolor{green!50!black}{\ding{51}}) Sim 0.75 \\[2pt]
\includegraphics[height=1.8cm]{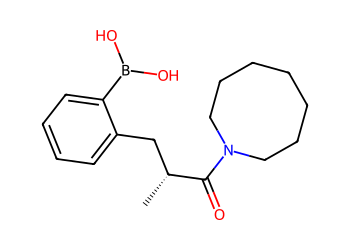} & \includegraphics[height=1.8cm]{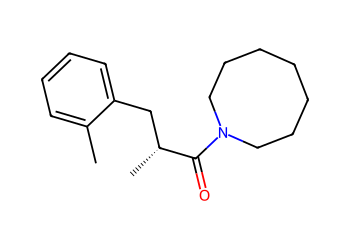} & \includegraphics[height=1.8cm]{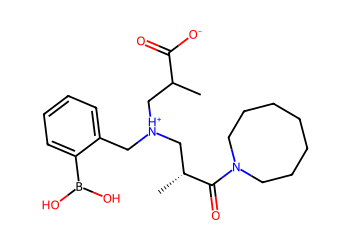} & SFT: 1 ($\Delta${-}2, \textcolor{red}{\ding{55}}) Sim 0.72 \newline \method: 5 ($\Delta$+2, \textcolor{green!50!black}{\ding{51}}) Sim 0.63 \\[2pt]
\includegraphics[height=1.8cm]{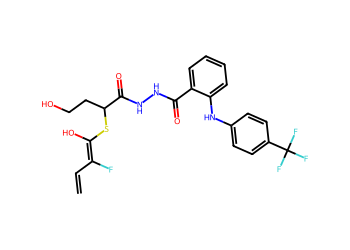} & \includegraphics[height=1.8cm]{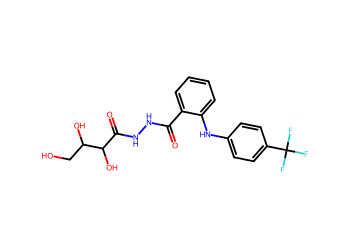} & \includegraphics[height=1.8cm]{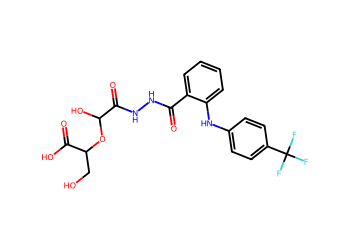} & SFT: 6 ($\Delta$0, \textcolor{red}{\ding{55}}) Sim 0.56 \newline \method: 7 ($\Delta$+1, \textcolor{green!50!black}{\ding{51}}) Sim 0.54 \\
\bottomrule
\end{tabular}
\caption{Case study: steering H-bond acceptors with Feature F18533. Source HBA values: 3, 4, 6, 3, 6. SFT fails to increase acceptors (unchanged or decreased); \method steering adds +1 to +2.}
\label{tab:case_acceptor}
\end{table*}

\subsection{\method Steering Examples}
\label{sec:app_csae_case}

\begin{figure*}[ht]
\centering
\setlength{\tabcolsep}{1pt}
\begin{tabular}{ccc@{\hspace{8pt}}ccc}
\textbf{Source} & \textbf{SFT} & \textbf{+\method} & \textbf{Source} & \textbf{SFT} & \textbf{+\method} \\[2pt]
\includegraphics[height=1.6cm]{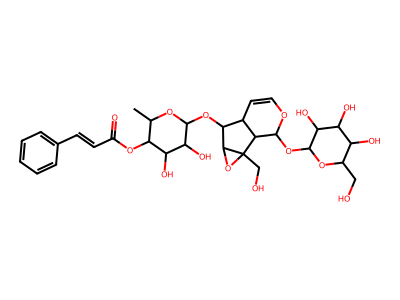} & \includegraphics[height=1.6cm]{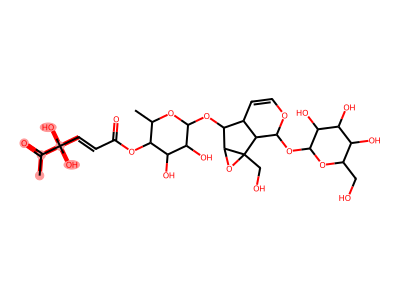} & \includegraphics[height=1.6cm]{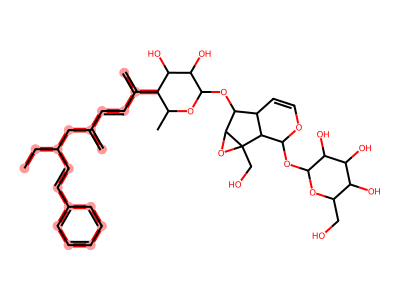} & \includegraphics[height=1.6cm]{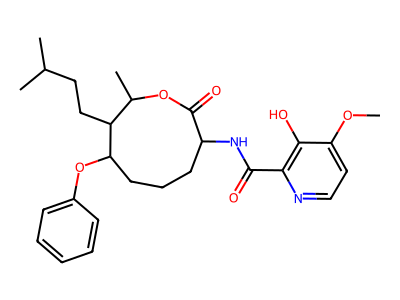} & \includegraphics[height=1.6cm]{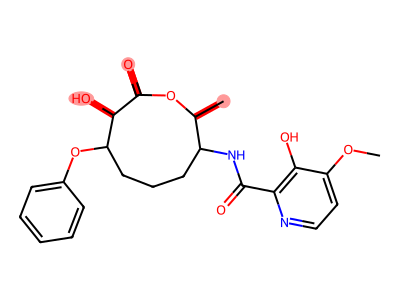} & \includegraphics[height=1.6cm]{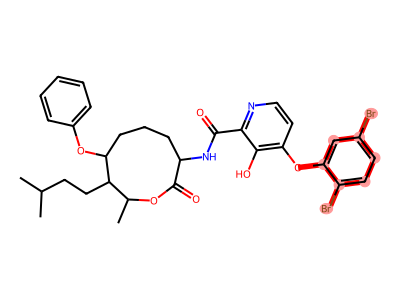} \\[-2pt]
{\scriptsize -2.47} & {\scriptsize -5.36 (\textcolor{red}{-2.89}) Sim 0.75} & {\scriptsize \textbf{1.32} (\textcolor{green!50!black}{+3.79}) Sim 0.64} & {\scriptsize 4.51} & {\scriptsize 1.82 (\textcolor{red}{-2.69}) Sim 0.65} & {\scriptsize \textbf{7.82} (\textcolor{green!50!black}{+3.31}) Sim 0.79} \\[6pt]
\includegraphics[height=1.6cm]{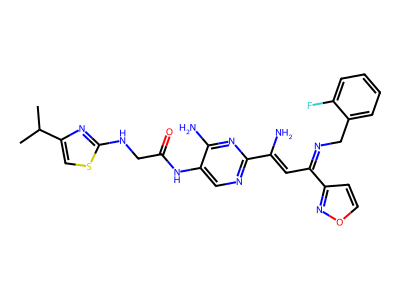} & \includegraphics[height=1.6cm]{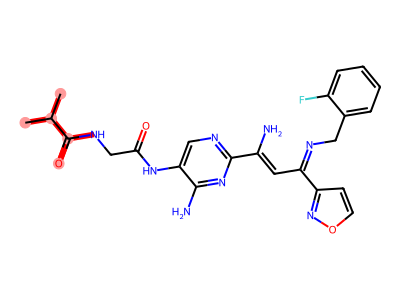} & \includegraphics[height=1.6cm]{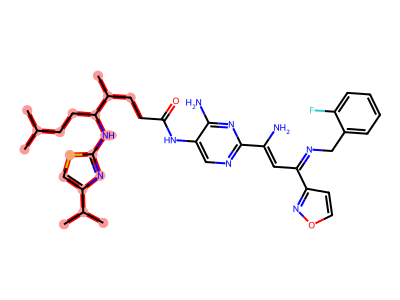} & \includegraphics[height=1.6cm]{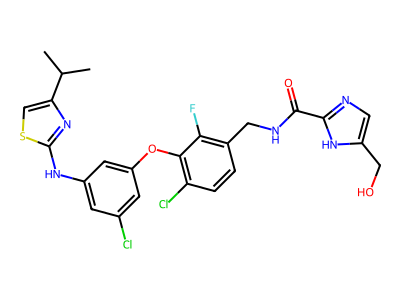} & \includegraphics[height=1.6cm]{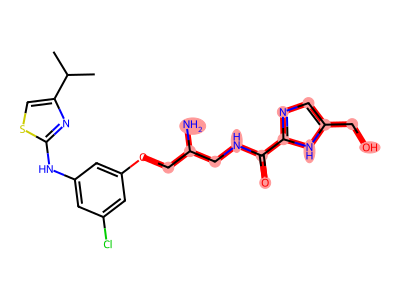} & \includegraphics[height=1.6cm]{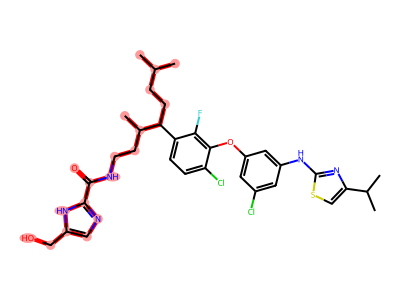} \\[-2pt]
{\scriptsize 3.81} & {\scriptsize 1.89 (\textcolor{red}{-1.92}) Sim 0.75} & {\scriptsize \textbf{7.03} (\textcolor{green!50!black}{+3.22}) Sim 0.77} & {\scriptsize 6.39} & {\scriptsize 3.02 (\textcolor{red}{-3.38}) Sim 0.67} & {\scriptsize \textbf{9.44} (\textcolor{green!50!black}{+3.05}) Sim 0.74} \\[6pt]
\includegraphics[height=1.6cm]{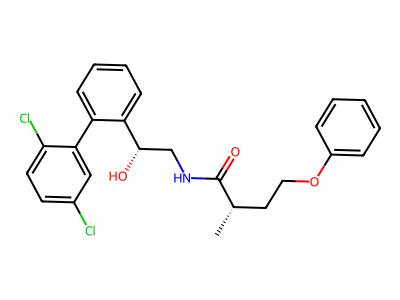} & \includegraphics[height=1.6cm]{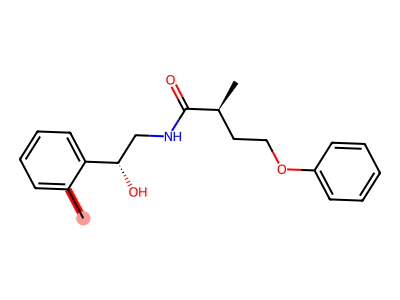} & \includegraphics[height=1.6cm]{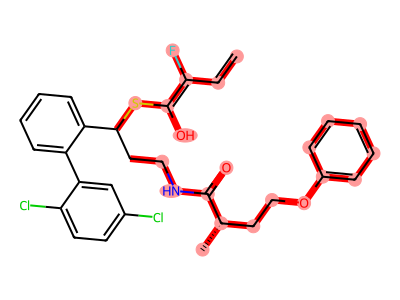} & \includegraphics[height=1.6cm]{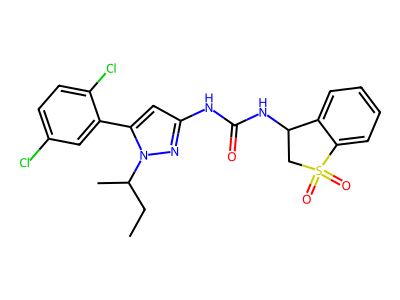} & \includegraphics[height=1.6cm]{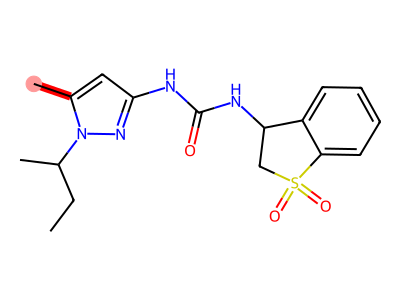} & \includegraphics[height=1.6cm]{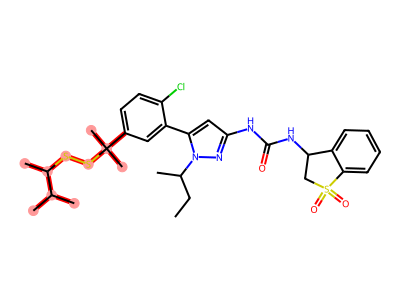} \\
\end{tabular}
\caption{\method steering examples: logP$\uparrow$ (DrugAssist). Red highlights indicate structural modifications. Each case shows SFT failing while \method succeeds.}
\label{fig:csae_logp}
\end{figure*}

\begin{figure*}[ht]
\centering
\setlength{\tabcolsep}{1pt}
\begin{tabular}{ccc@{\hspace{8pt}}ccc}
\textbf{Source} & \textbf{SFT} & \textbf{+\method} & \textbf{Source} & \textbf{SFT} & \textbf{+\method} \\[2pt]
\includegraphics[height=1.6cm]{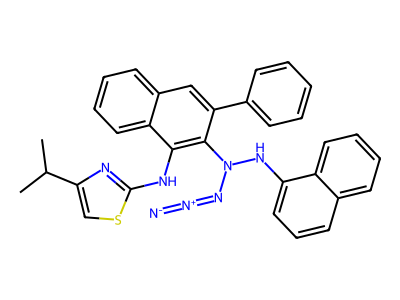} & \includegraphics[height=1.6cm]{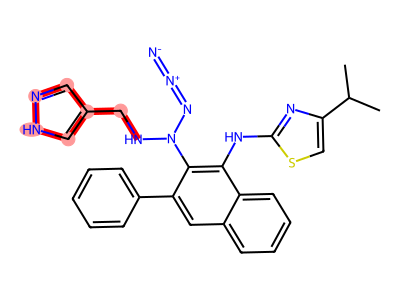} & \includegraphics[height=1.6cm]{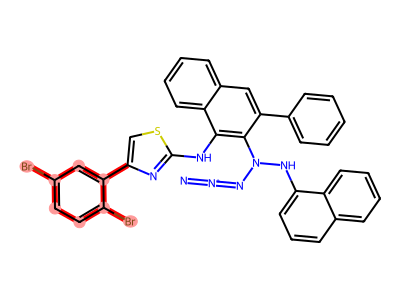} & \includegraphics[height=1.6cm]{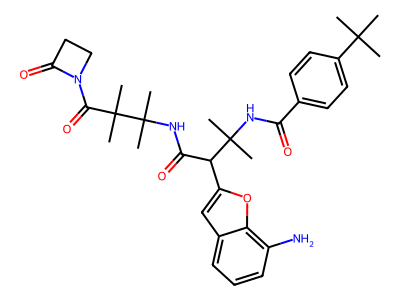} & \includegraphics[height=1.6cm]{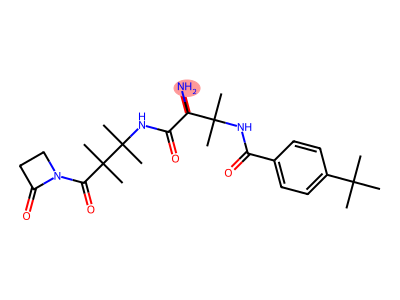} & \includegraphics[height=1.6cm]{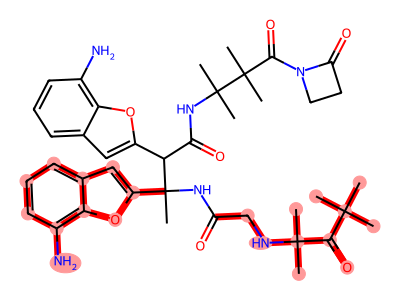} \\[-2pt]
{\scriptsize 542} & {\scriptsize 496 (\textcolor{red}{-46}) Sim 0.67} & {\scriptsize \textbf{733} (\textcolor{green!50!black}{+192}) Sim 0.69} & {\scriptsize 589} & {\scriptsize 473 (\textcolor{red}{-116}) Sim 0.62} & {\scriptsize \textbf{729} (\textcolor{green!50!black}{+140}) Sim 0.61} \\[6pt]
\includegraphics[height=1.6cm]{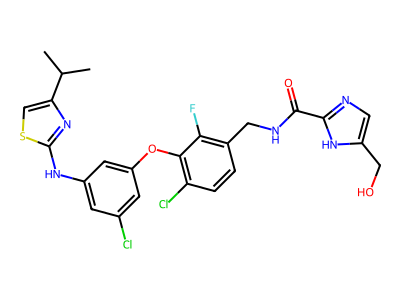} & \includegraphics[height=1.6cm]{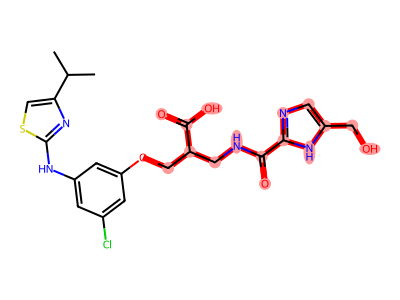} & \includegraphics[height=1.6cm]{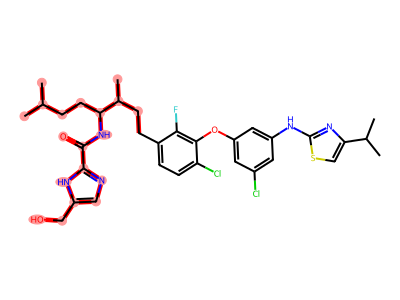} & \includegraphics[height=1.6cm]{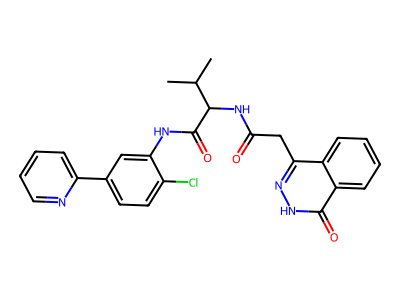} & \includegraphics[height=1.6cm]{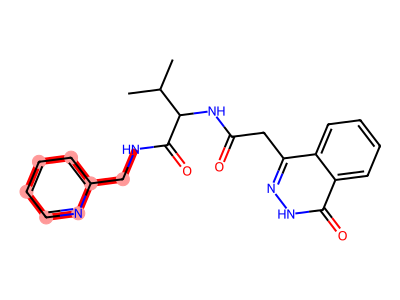} & \includegraphics[height=1.6cm]{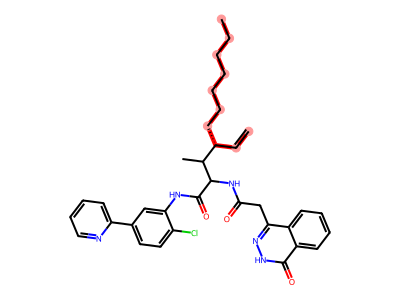} \\[-2pt]
{\scriptsize 550} & {\scriptsize 494 (\textcolor{red}{-56}) Sim 0.67} & {\scriptsize \textbf{677} (\textcolor{green!50!black}{+126}) Sim 0.77} & {\scriptsize 490} & {\scriptsize 393 (\textcolor{red}{-97}) Sim 0.63} & {\scriptsize \textbf{614} (\textcolor{green!50!black}{+124}) Sim 0.73} \\[6pt]
\includegraphics[height=1.6cm]{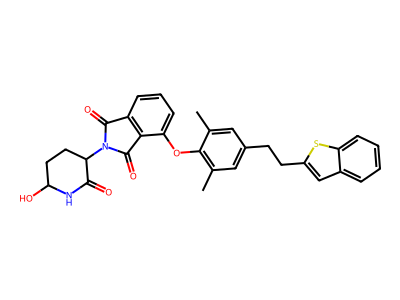} & \includegraphics[height=1.6cm]{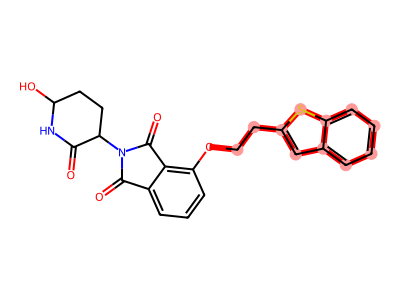} & \includegraphics[height=1.6cm]{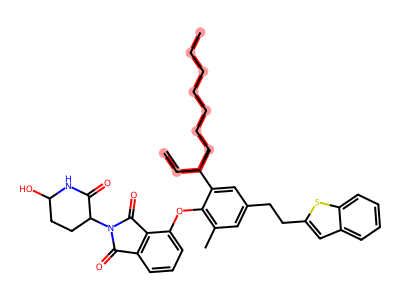} & \includegraphics[height=1.6cm]{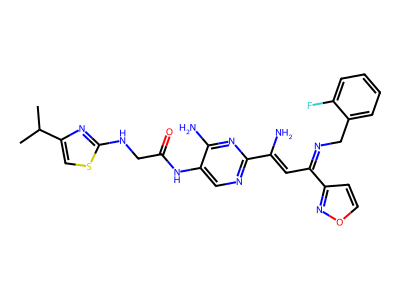} & \includegraphics[height=1.6cm]{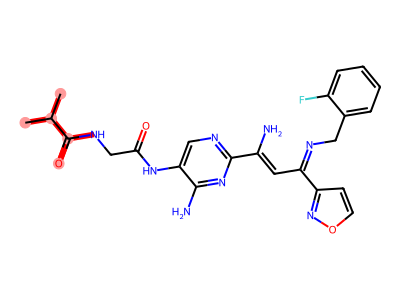} & \includegraphics[height=1.6cm]{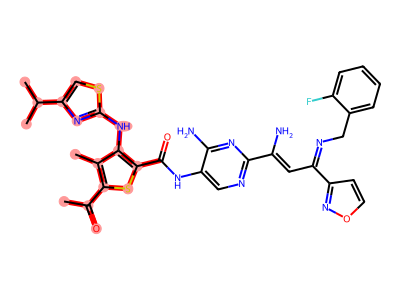} \\
\end{tabular}
\caption{\method steering examples: MW$\uparrow$ (DrugAssist). Red highlights indicate structural modifications. Each case shows SFT failing while \method succeeds.}
\label{fig:csae_mw}
\end{figure*}

\begin{figure*}[ht]
\centering
\setlength{\tabcolsep}{1pt}
\begin{tabular}{ccc@{\hspace{8pt}}ccc}
\textbf{Source} & \textbf{SFT} & \textbf{+\method} & \textbf{Source} & \textbf{SFT} & \textbf{+\method} \\[2pt]
\includegraphics[height=1.6cm]{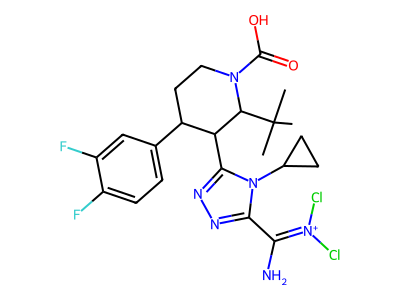} & \includegraphics[height=1.6cm]{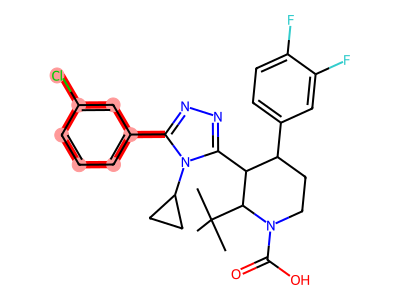} & \includegraphics[height=1.6cm]{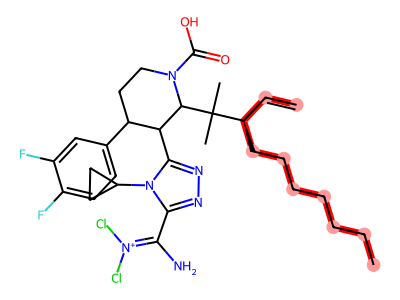} & \includegraphics[height=1.6cm]{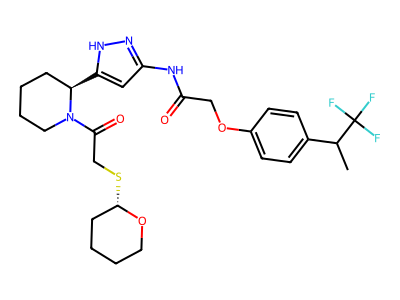} & \includegraphics[height=1.6cm]{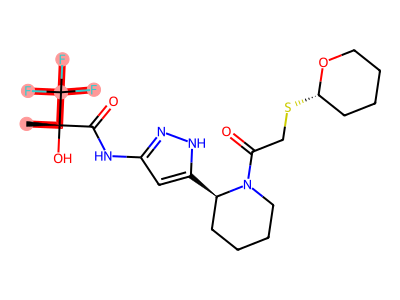} & \includegraphics[height=1.6cm]{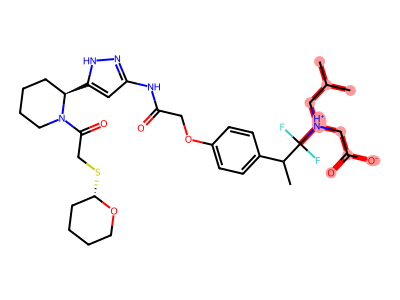} \\[-2pt]
{\scriptsize 4} & {\scriptsize 4 (\textcolor{red}{+0}) Sim 0.67} & {\scriptsize \textbf{13} (\textcolor{green!50!black}{+9}) Sim 0.72} & {\scriptsize 9} & {\scriptsize 6 (\textcolor{red}{-3}) Sim 0.65} & {\scriptsize \textbf{15} (\textcolor{green!50!black}{+6}) Sim 0.81} \\[6pt]
\includegraphics[height=1.6cm]{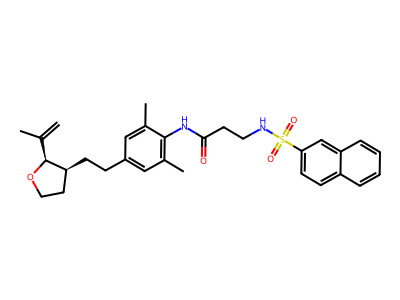} & \includegraphics[height=1.6cm]{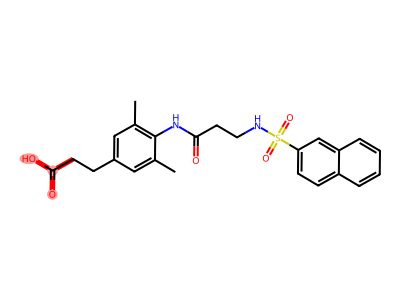} & \includegraphics[height=1.6cm]{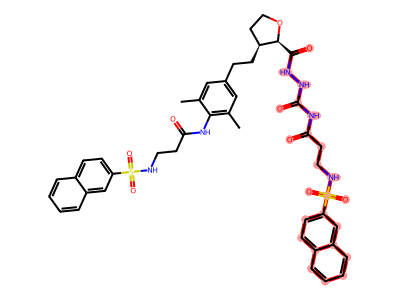} & \includegraphics[height=1.6cm]{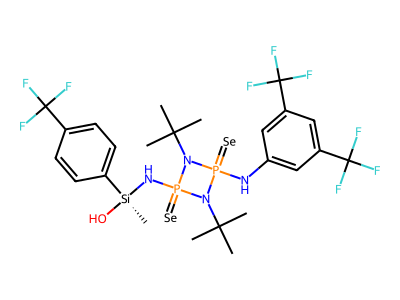} & \includegraphics[height=1.6cm]{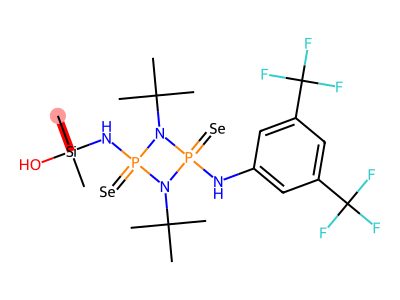} & \includegraphics[height=1.6cm]{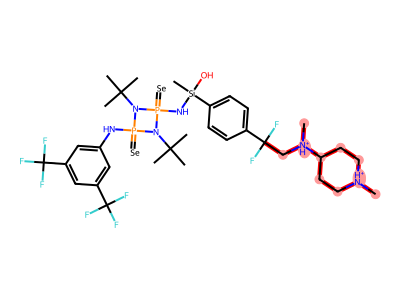} \\[-2pt]
{\scriptsize 10} & {\scriptsize 9 (\textcolor{red}{-1}) Sim 0.62} & {\scriptsize \textbf{15} (\textcolor{green!50!black}{+5}) Sim 0.77} & {\scriptsize 5} & {\scriptsize 4 (\textcolor{red}{-1}) Sim 0.76} & {\scriptsize \textbf{9} (\textcolor{green!50!black}{+4}) Sim 0.65} \\[6pt]
\includegraphics[height=1.6cm]{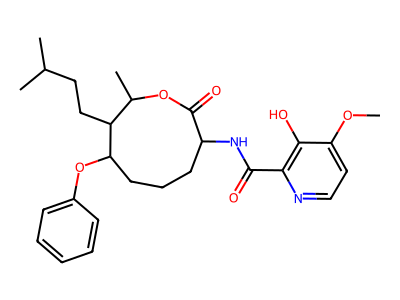} & \includegraphics[height=1.6cm]{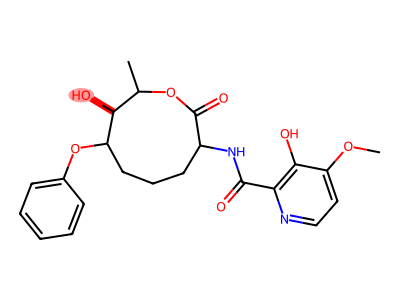} & \includegraphics[height=1.6cm]{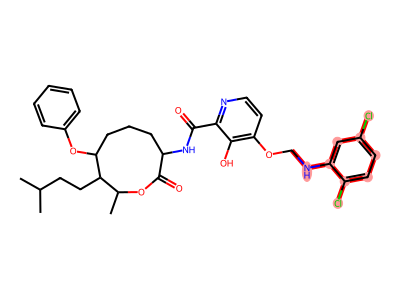} & \includegraphics[height=1.6cm]{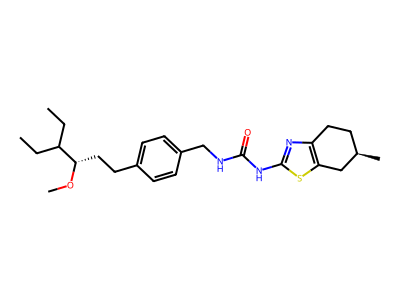} & \includegraphics[height=1.6cm]{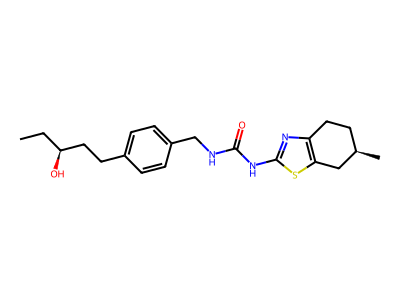} & \includegraphics[height=1.6cm]{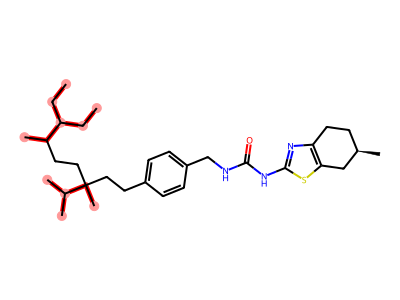} \\
\end{tabular}
\caption{\method steering examples: RotBond$\uparrow$ (DrugAssist). Red highlights indicate structural modifications. Each case shows SFT failing while \method succeeds.}
\label{fig:csae_rotbond}
\end{figure*}

\begin{figure*}[ht]
\centering
\setlength{\tabcolsep}{1pt}
\begin{tabular}{ccc@{\hspace{8pt}}ccc}
\textbf{Source} & \textbf{SFT} & \textbf{+\method} & \textbf{Source} & \textbf{SFT} & \textbf{+\method} \\[2pt]
\includegraphics[height=1.6cm]{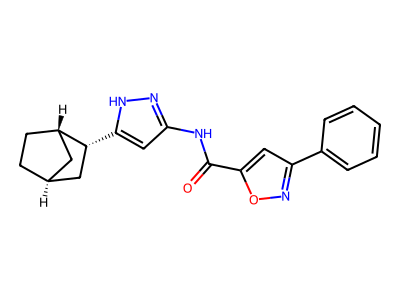} & \includegraphics[height=1.6cm]{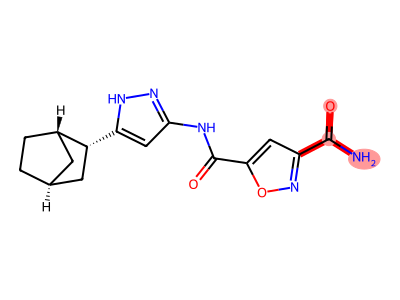} & \includegraphics[height=1.6cm]{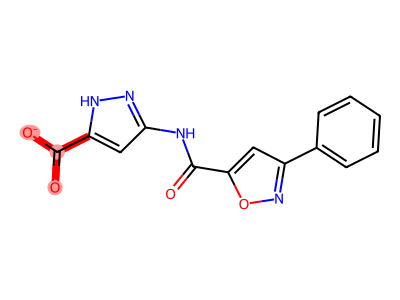} & \includegraphics[height=1.6cm]{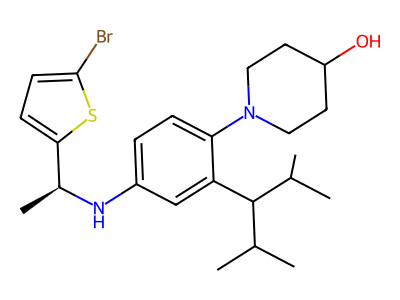} & \includegraphics[height=1.6cm]{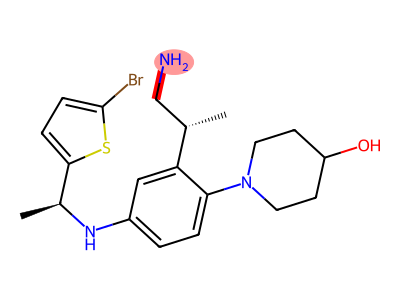} & \includegraphics[height=1.6cm]{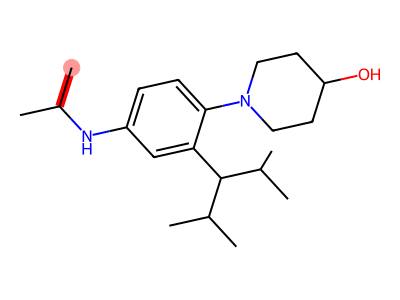} \\[-2pt]
{\scriptsize 4.19} & {\scriptsize 4.50 (\textcolor{red}{-0.31}) Sim 0.70} & {\scriptsize \textbf{2.67} (\textcolor{green!50!black}{+1.52}) Sim 0.60} & {\scriptsize 3.24} & {\scriptsize 3.42 (\textcolor{red}{-0.17}) Sim 0.77} & {\scriptsize \textbf{2.62} (\textcolor{green!50!black}{+0.62}) Sim 0.67} \\[6pt]
\includegraphics[height=1.6cm]{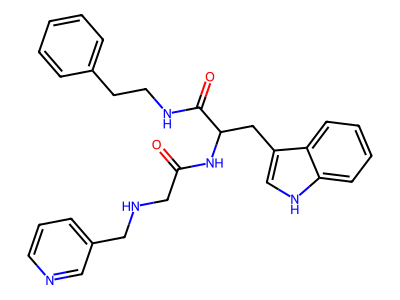} & \includegraphics[height=1.6cm]{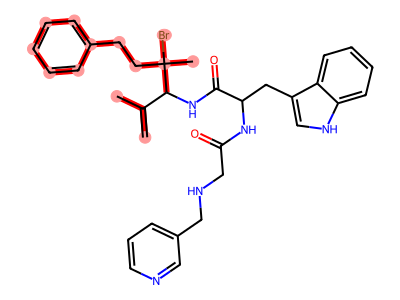} & \includegraphics[height=1.6cm]{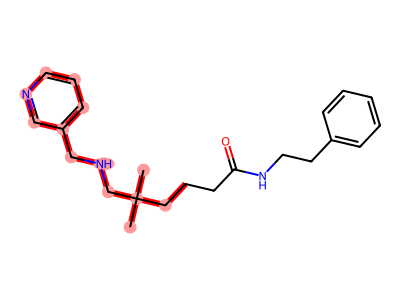} & \includegraphics[height=1.6cm]{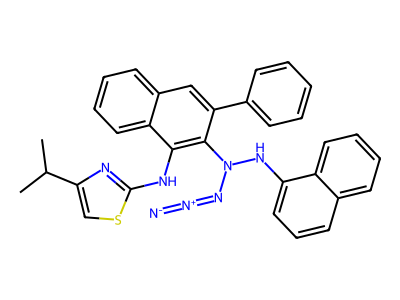} & \includegraphics[height=1.6cm]{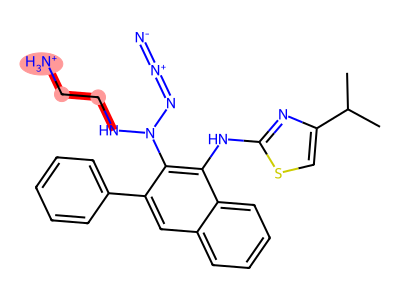} & \includegraphics[height=1.6cm]{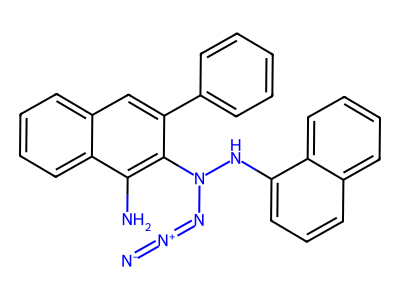} \\[-2pt]
{\scriptsize 2.66} & {\scriptsize 3.77 (\textcolor{red}{-1.11}) Sim 0.71} & {\scriptsize \textbf{2.16} (\textcolor{green!50!black}{+0.51}) Sim 0.43} & {\scriptsize 3.14} & {\scriptsize 3.57 (\textcolor{red}{-0.43}) Sim 0.74} & {\scriptsize \textbf{2.83} (\textcolor{green!50!black}{+0.30}) Sim 0.58} \\[6pt]
\includegraphics[height=1.6cm]{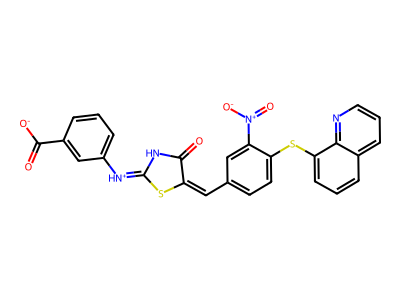} & \includegraphics[height=1.6cm]{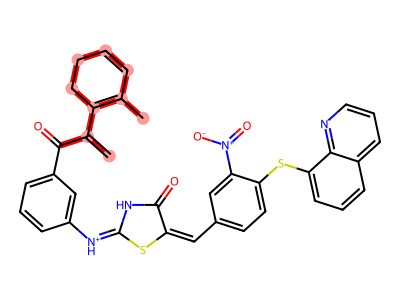} & \includegraphics[height=1.6cm]{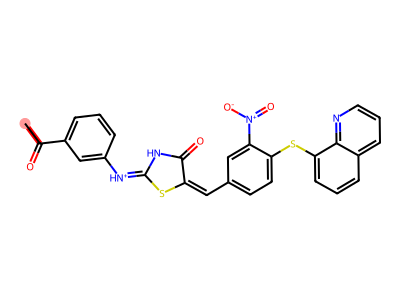} & \includegraphics[height=1.6cm]{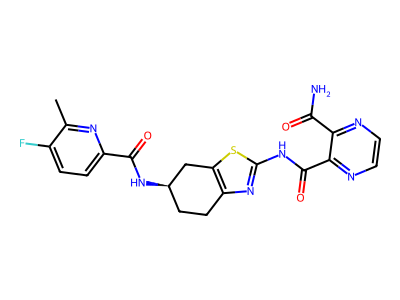} & \includegraphics[height=1.6cm]{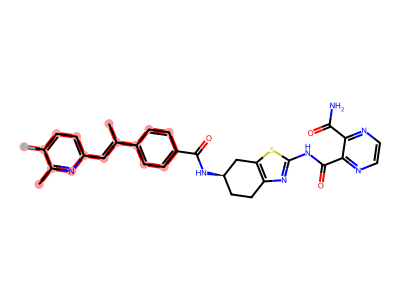} & \includegraphics[height=1.6cm]{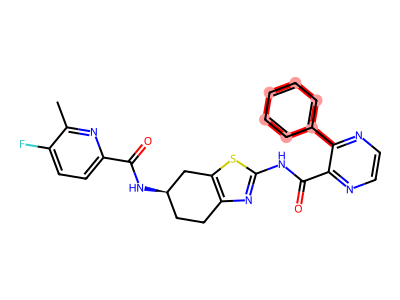} \\
\end{tabular}
\caption{\method steering examples: SA$\downarrow$ (DrugAssist). Red highlights indicate structural modifications. Each case shows SFT failing while \method succeeds.}
\label{fig:csae_sa}
\end{figure*}

\begin{figure*}[ht]
\centering
\setlength{\tabcolsep}{1pt}
\begin{tabular}{ccc@{\hspace{8pt}}ccc}
\textbf{Source} & \textbf{SFT} & \textbf{+\method} & \textbf{Source} & \textbf{SFT} & \textbf{+\method} \\[2pt]
\includegraphics[height=1.6cm]{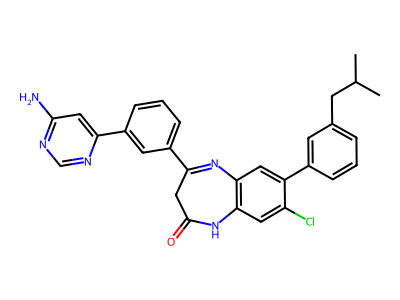} & \includegraphics[height=1.6cm]{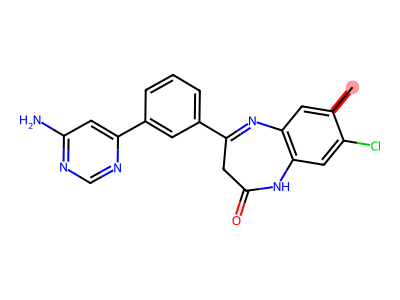} & \includegraphics[height=1.6cm]{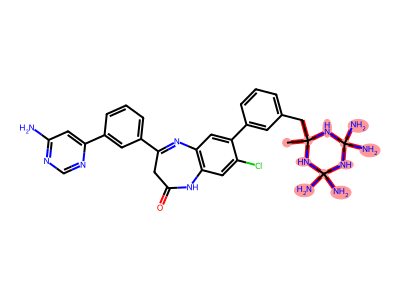} & \includegraphics[height=1.6cm]{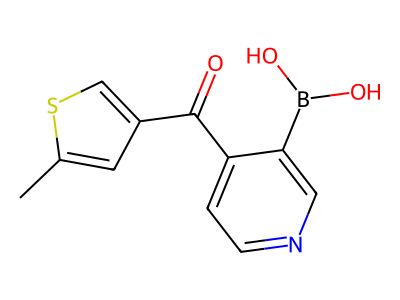} & \includegraphics[height=1.6cm]{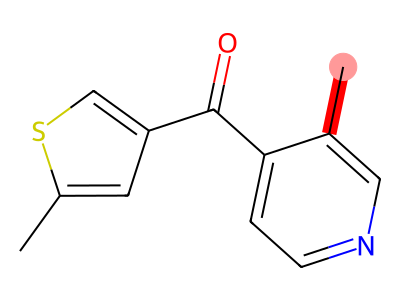} & \includegraphics[height=1.6cm]{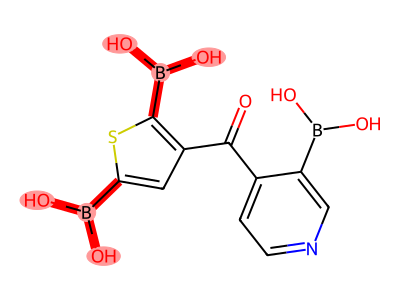} \\[-2pt]
{\scriptsize 5} & {\scriptsize 5 (\textcolor{red}{+0}) Sim 0.69} & {\scriptsize \textbf{12} (\textcolor{green!50!black}{+7}) Sim 0.75} & {\scriptsize 5} & {\scriptsize 3 (\textcolor{red}{-2}) Sim 0.72} & {\scriptsize \textbf{9} (\textcolor{green!50!black}{+4}) Sim 0.53} \\[6pt]
\includegraphics[height=1.6cm]{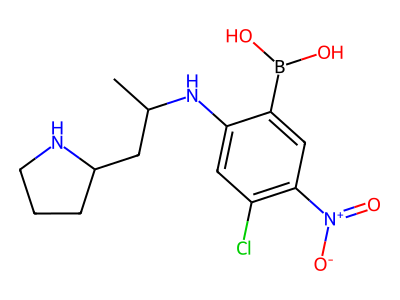} & \includegraphics[height=1.6cm]{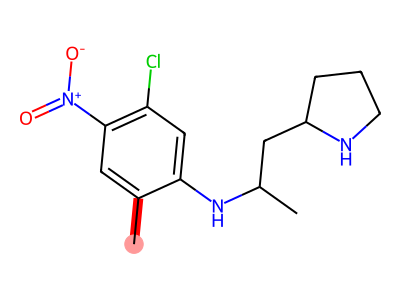} & \includegraphics[height=1.6cm]{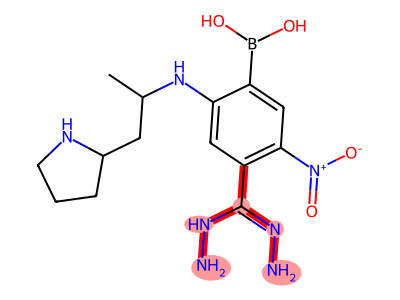} & \includegraphics[height=1.6cm]{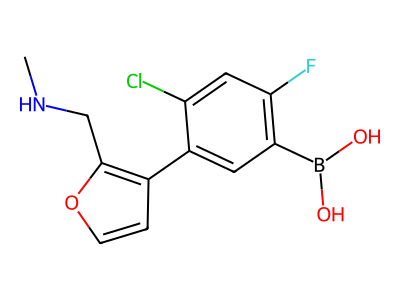} & \includegraphics[height=1.6cm]{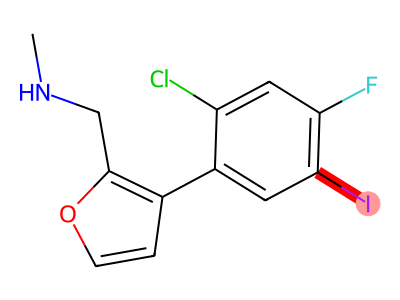} & \includegraphics[height=1.6cm]{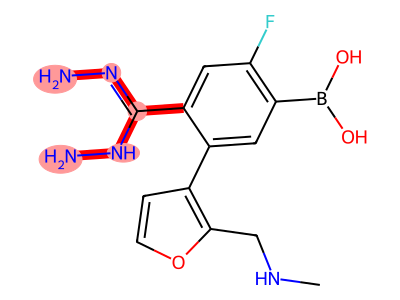} \\[-2pt]
{\scriptsize 6} & {\scriptsize 4 (\textcolor{red}{-2}) Sim 0.76} & {\scriptsize \textbf{9} (\textcolor{green!50!black}{+3}) Sim 0.69} & {\scriptsize 4} & {\scriptsize 2 (\textcolor{red}{-2}) Sim 0.69} & {\scriptsize \textbf{7} (\textcolor{green!50!black}{+3}) Sim 0.62} \\[6pt]
\includegraphics[height=1.6cm]{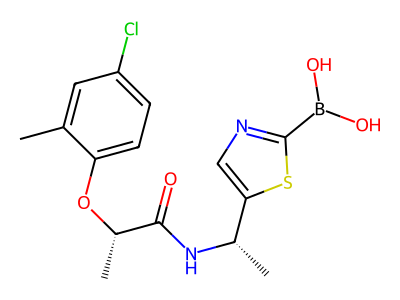} & \includegraphics[height=1.6cm]{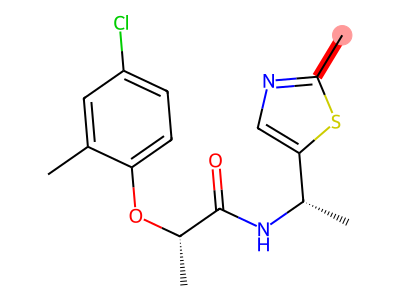} & \includegraphics[height=1.6cm]{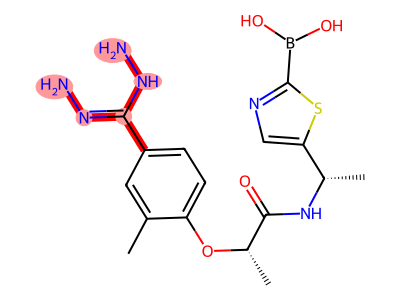} & \includegraphics[height=1.6cm]{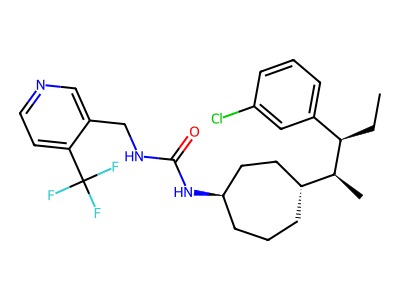} & \includegraphics[height=1.6cm]{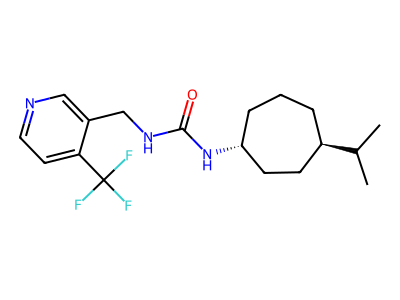} & \includegraphics[height=1.6cm]{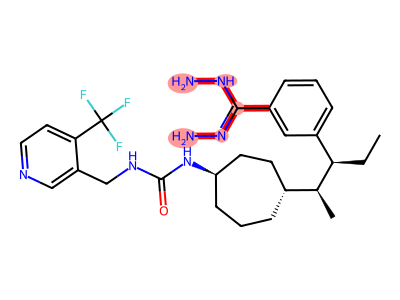} \\
\end{tabular}
\caption{\method steering examples: HBA$\uparrow$ (DrugAssist). Red highlights indicate structural modifications. Each case shows SFT failing while \method succeeds.}
\label{fig:csae_acceptor}
\end{figure*}

\begin{figure*}[ht]
\centering
\setlength{\tabcolsep}{1pt}
\begin{tabular}{ccc@{\hspace{8pt}}ccc}
\textbf{Source} & \textbf{SFT} & \textbf{+\method} & \textbf{Source} & \textbf{SFT} & \textbf{+\method} \\[2pt]
\includegraphics[height=1.6cm]{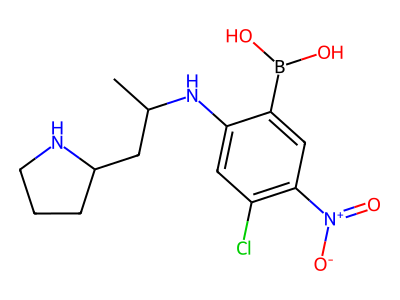} & \includegraphics[height=1.6cm]{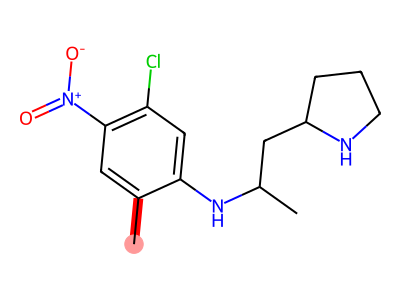} & \includegraphics[height=1.6cm]{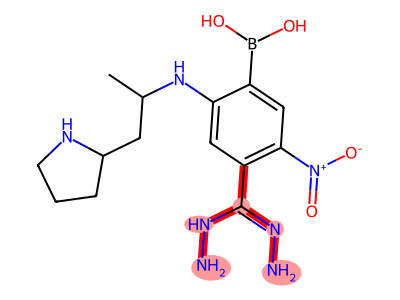} & \includegraphics[height=1.6cm]{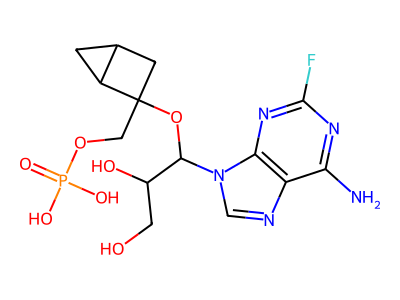} & \includegraphics[height=1.6cm]{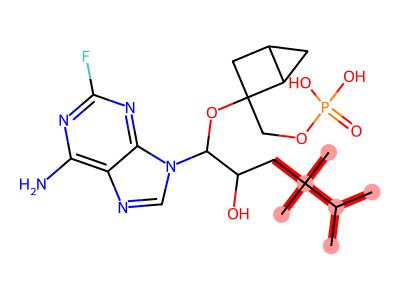} & \includegraphics[height=1.6cm]{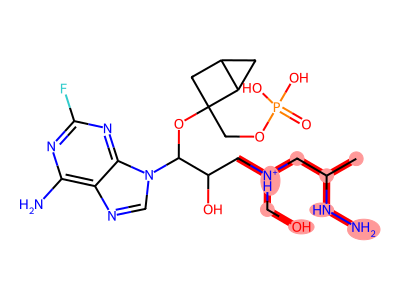} \\[-2pt]
{\scriptsize 4} & {\scriptsize 2 (\textcolor{red}{-2}) Sim 0.76} & {\scriptsize \textbf{7} (\textcolor{green!50!black}{+3}) Sim 0.69} & {\scriptsize 5} & {\scriptsize 4 (\textcolor{red}{-1}) Sim 0.81} & {\scriptsize \textbf{8} (\textcolor{green!50!black}{+3}) Sim 0.76} \\[6pt]
\includegraphics[height=1.6cm]{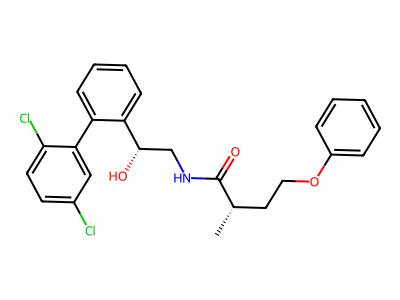} & \includegraphics[height=1.6cm]{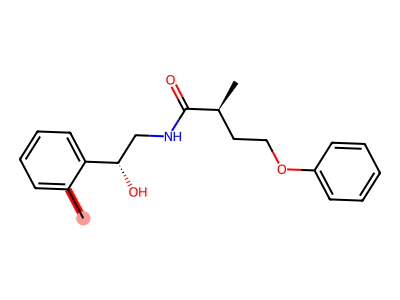} & \includegraphics[height=1.6cm]{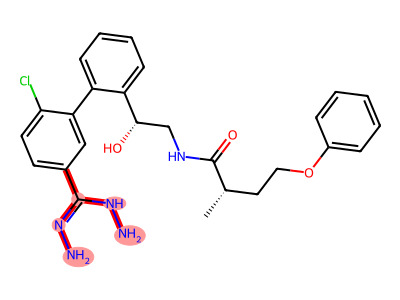} & \includegraphics[height=1.6cm]{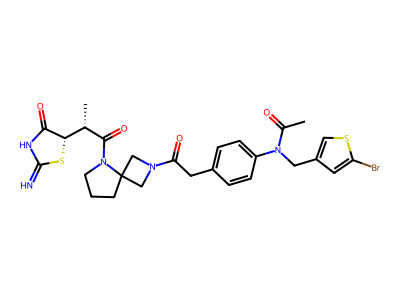} & \includegraphics[height=1.6cm]{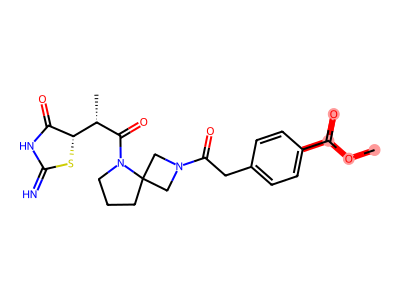} & \includegraphics[height=1.6cm]{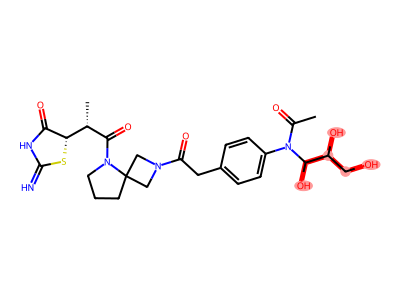} \\[-2pt]
{\scriptsize 2} & {\scriptsize 2 (\textcolor{red}{+0}) Sim 0.65} & {\scriptsize \textbf{5} (\textcolor{green!50!black}{+3}) Sim 0.73} & {\scriptsize 2} & {\scriptsize 2 (\textcolor{red}{+0}) Sim 0.63} & {\scriptsize \textbf{5} (\textcolor{green!50!black}{+3}) Sim 0.67} \\[6pt]
\includegraphics[height=1.6cm]{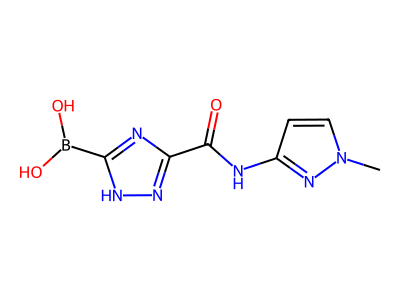} & \includegraphics[height=1.6cm]{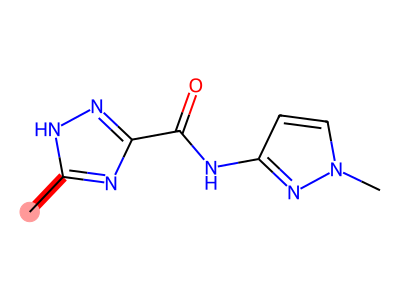} & \includegraphics[height=1.6cm]{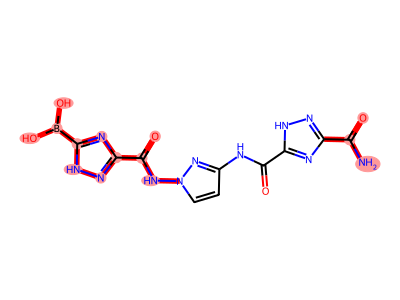} & \includegraphics[height=1.6cm]{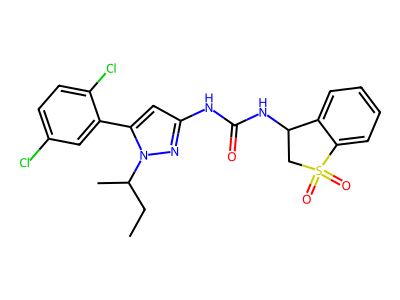} & \includegraphics[height=1.6cm]{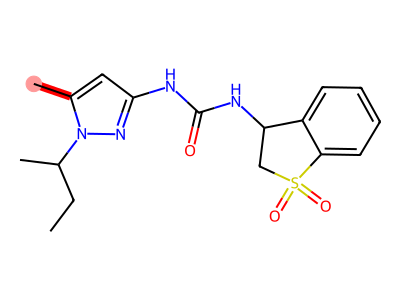} & \includegraphics[height=1.6cm]{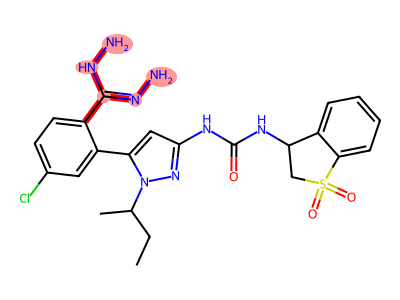} \\
\end{tabular}
\caption{\method steering examples: HBD$\uparrow$ (DrugAssist). Red highlights indicate structural modifications. Each case shows SFT failing while \method succeeds.}
\label{fig:csae_donor}
\end{figure*}

Figures~\ref{fig:csae_logp}--\ref{fig:csae_donor} show molecular editing examples using the full \method steering pipeline on DrugAssist. Red-highlighted regions indicate structural modifications relative to the source molecule. Each figure shows cases where SFT fails to improve the target property but \method succeeds. For each case, we report the property value, change ($\Delta$), and Tanimoto similarity (Sim) to the source molecule.

\begin{figure*}[ht]
\centering
\setlength{\tabcolsep}{1pt}
\begin{tabular}{ccc@{\hspace{8pt}}ccc}
\textbf{Source} & \textbf{SFT} & \textbf{+\method} & \textbf{Source} & \textbf{SFT} & \textbf{+\method} \\[2pt]
\includegraphics[height=1.6cm]{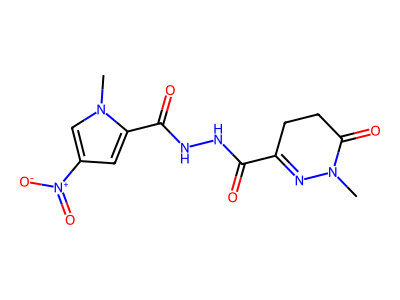} & \includegraphics[height=1.6cm]{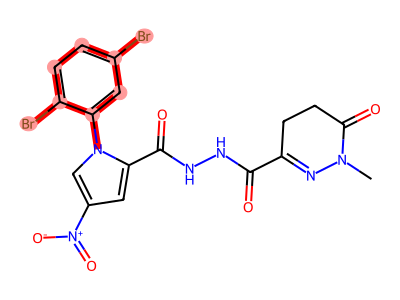} & \includegraphics[height=1.6cm]{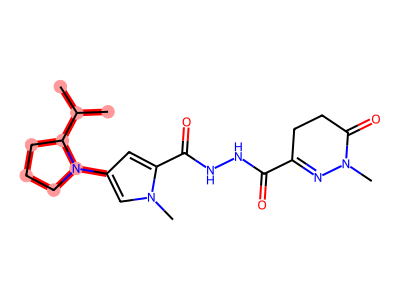} & \includegraphics[height=1.6cm]{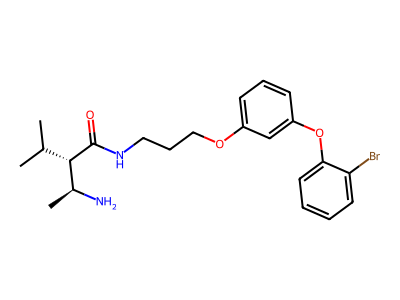} & \includegraphics[height=1.6cm]{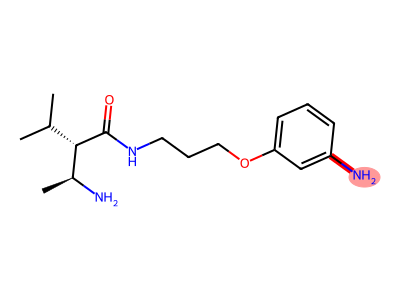} & \includegraphics[height=1.6cm]{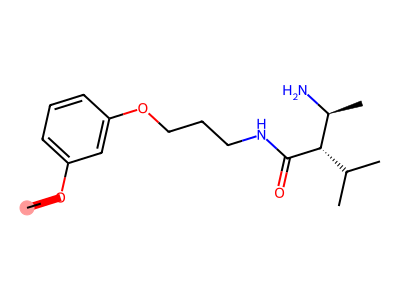} \\[-2pt]
{\scriptsize 0.56} & {\scriptsize 0.45 (\textcolor{red}{-0.11}) Sim 0.68} & {\scriptsize \textbf{0.75} (\textcolor{green!50!black}{+0.19}) Sim 0.54} & {\scriptsize 0.51} & {\scriptsize 0.50 (\textcolor{red}{-0.01}) Sim 0.67} & {\scriptsize \textbf{0.69} (\textcolor{green!50!black}{+0.17}) Sim 0.70} \\[6pt]
\includegraphics[height=1.6cm]{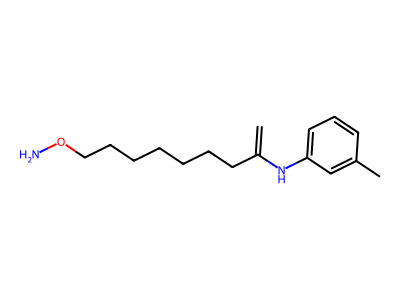} & \includegraphics[height=1.6cm]{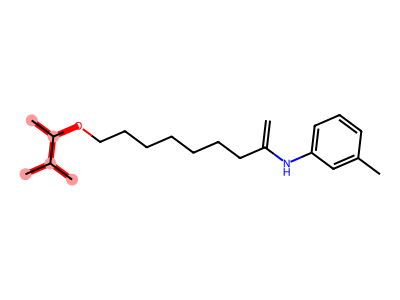} & \includegraphics[height=1.6cm]{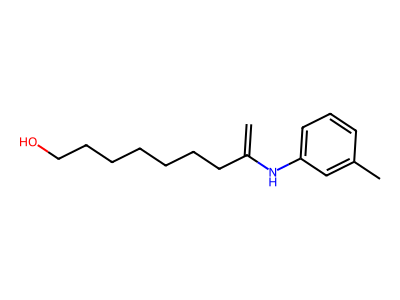} & \includegraphics[height=1.6cm]{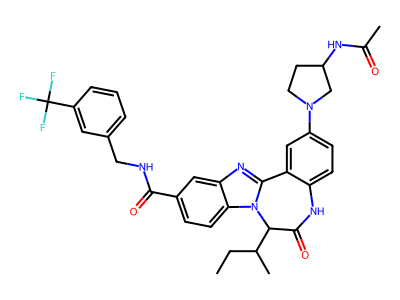} & \includegraphics[height=1.6cm]{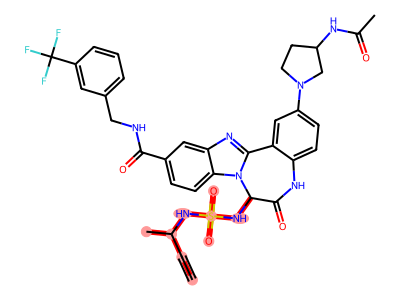} & \includegraphics[height=1.6cm]{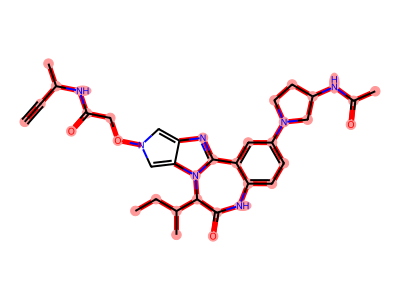} \\[-2pt]
{\scriptsize 0.49} & {\scriptsize 0.46 (\textcolor{red}{-0.04}) Sim 0.70} & {\scriptsize \textbf{0.64} (\textcolor{green!50!black}{+0.15}) Sim 0.71} & {\scriptsize 0.24} & {\scriptsize 0.17 (\textcolor{red}{-0.07}) Sim 0.75} & {\scriptsize \textbf{0.37} (\textcolor{green!50!black}{+0.14}) Sim 0.56} \\[6pt]
\includegraphics[height=1.6cm]{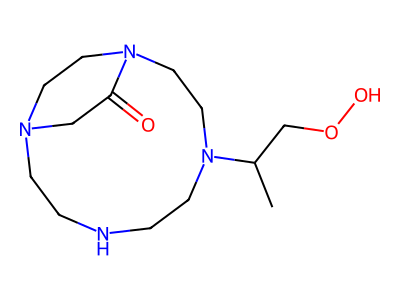} & \includegraphics[height=1.6cm]{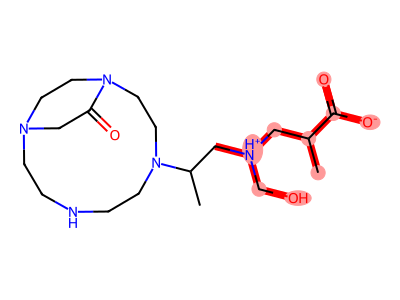} & \includegraphics[height=1.6cm]{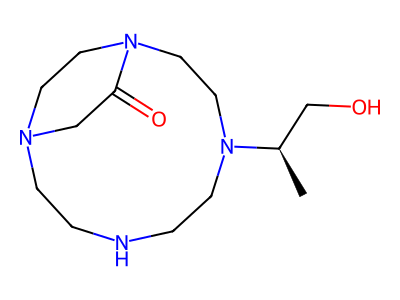} & \includegraphics[height=1.6cm]{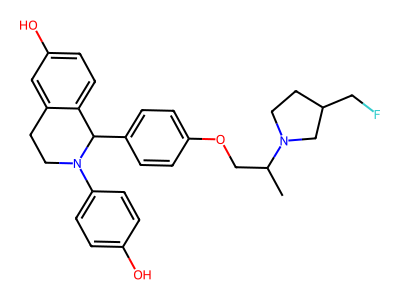} & \includegraphics[height=1.6cm]{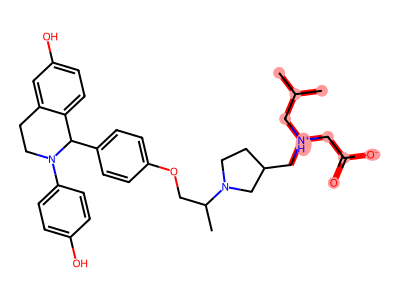} & \includegraphics[height=1.6cm]{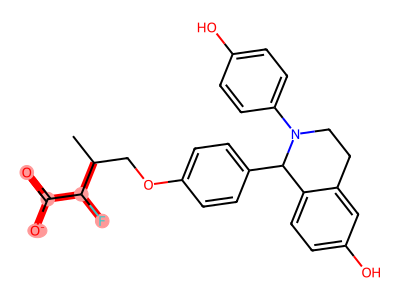} \\
\end{tabular}
\caption{\method steering examples: QED$\uparrow$ (DrugAssist). Red highlights indicate structural modifications. Each case shows SFT failing while \method succeeds.}
\label{fig:csae_qed}
\end{figure*}

\begin{figure*}[ht]
\centering
\setlength{\tabcolsep}{1pt}
\begin{tabular}{ccc@{\hspace{8pt}}ccc}
\textbf{Source} & \textbf{SFT} & \textbf{+\method} & \textbf{Source} & \textbf{SFT} & \textbf{+\method} \\[2pt]
\includegraphics[height=1.6cm]{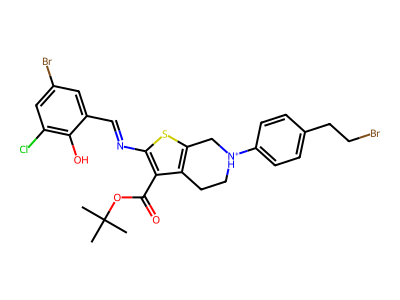} & \includegraphics[height=1.6cm]{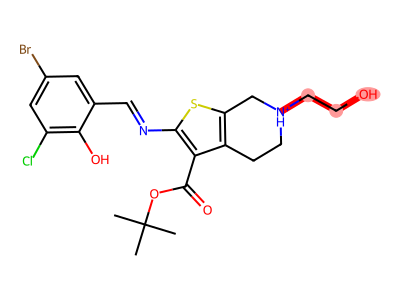} & \includegraphics[height=1.6cm]{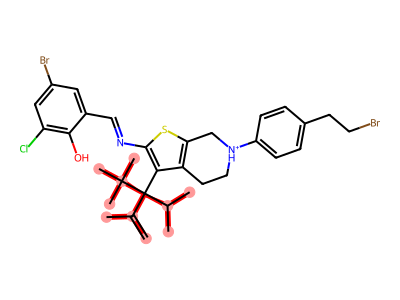} & \includegraphics[height=1.6cm]{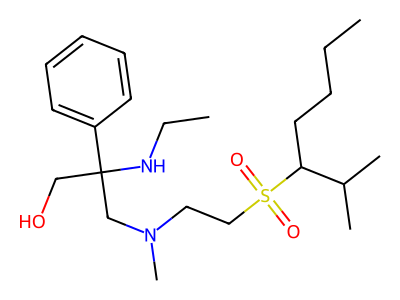} & \includegraphics[height=1.6cm]{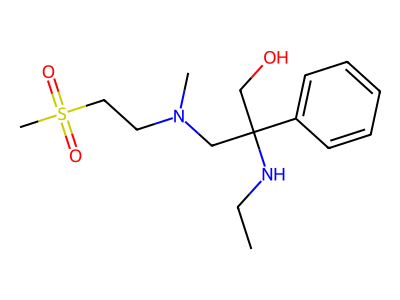} & \includegraphics[height=1.6cm]{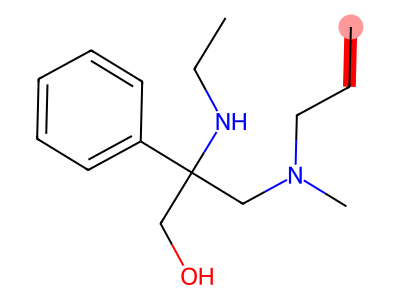} \\[-2pt]
{\scriptsize 0.16} & {\scriptsize 0.11 (\textcolor{red}{-0.05}) Sim 0.69} & {\scriptsize \textbf{0.29} (\textcolor{green!50!black}{+0.13}) Sim 0.69} & {\scriptsize 0.14} & {\scriptsize 0.08 (\textcolor{red}{-0.06}) Sim 0.65} & {\scriptsize \textbf{0.26} (\textcolor{green!50!black}{+0.12}) Sim 0.60} \\[6pt]
\includegraphics[height=1.6cm]{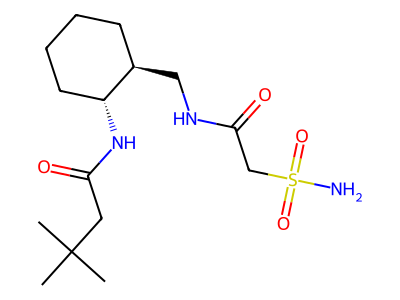} & \includegraphics[height=1.6cm]{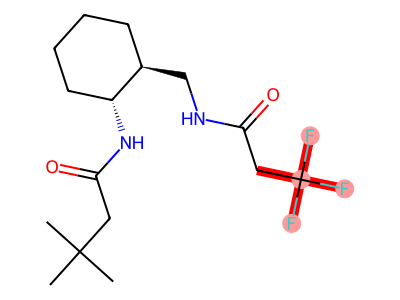} & \includegraphics[height=1.6cm]{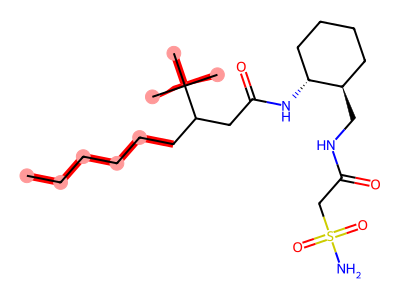} & \includegraphics[height=1.6cm]{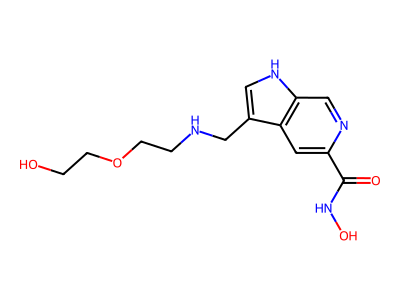} & \includegraphics[height=1.6cm]{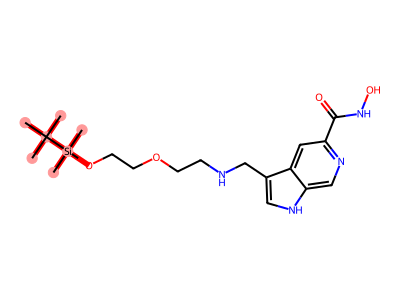} & \includegraphics[height=1.6cm]{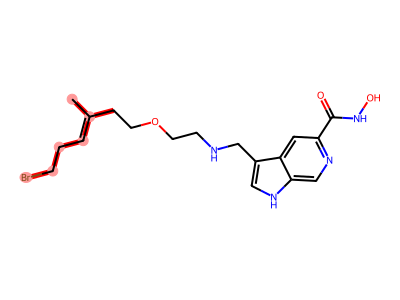} \\[-2pt]
{\scriptsize 0.02} & {\scriptsize 0.01 (\textcolor{red}{-0.01}) Sim 0.67} & {\scriptsize \textbf{0.13} (\textcolor{green!50!black}{+0.11}) Sim 0.61} & {\scriptsize 0.10} & {\scriptsize 0.10 (\textcolor{red}{+0.00}) Sim 0.73} & {\scriptsize \textbf{0.17} (\textcolor{green!50!black}{+0.07}) Sim 0.68} \\[6pt]
\includegraphics[height=1.6cm]{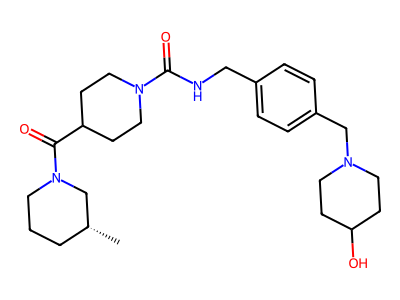} & \includegraphics[height=1.6cm]{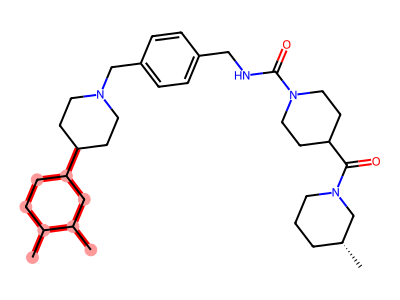} & \includegraphics[height=1.6cm]{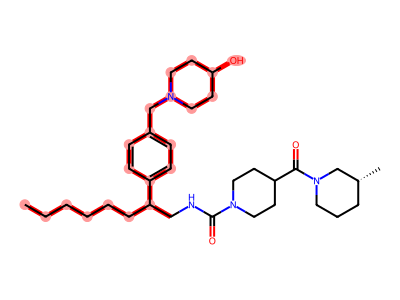} & \includegraphics[height=1.6cm]{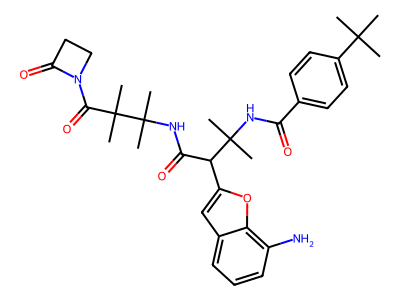} & \includegraphics[height=1.6cm]{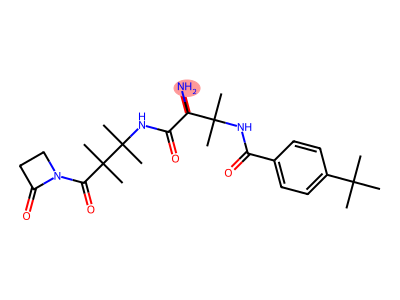} & \includegraphics[height=1.6cm]{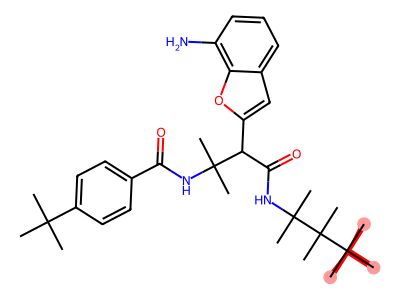} \\
\end{tabular}
\caption{\method steering examples: DRD2$\uparrow$ (DrugAssist). Red highlights indicate structural modifications. Each case shows SFT failing while \method succeeds.}
\label{fig:csae_drd2}
\end{figure*}

\subsubsection*{GeLLM3O-Mistral}

\begin{figure*}[ht]
\centering
\setlength{\tabcolsep}{1pt}
\begin{tabular}{ccc@{\hspace{8pt}}ccc}
\textbf{Source} & \textbf{SFT} & \textbf{+\method} & \textbf{Source} & \textbf{SFT} & \textbf{+\method} \\[2pt]
\includegraphics[height=1.6cm]{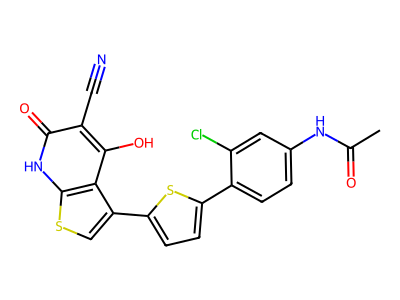} & \includegraphics[height=1.6cm]{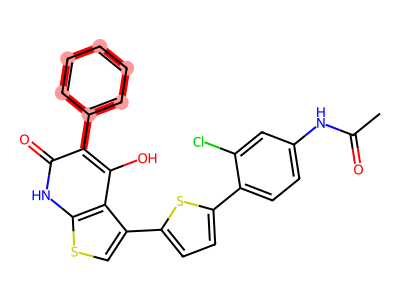} & \includegraphics[height=1.6cm]{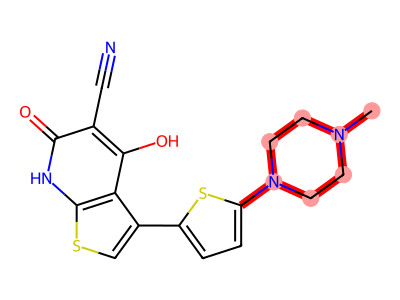} & \includegraphics[height=1.6cm]{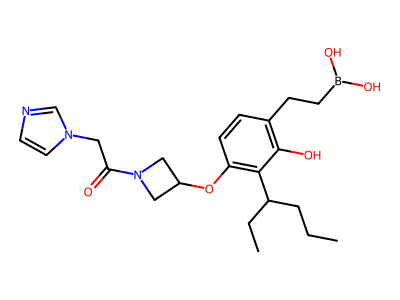} & \includegraphics[height=1.6cm]{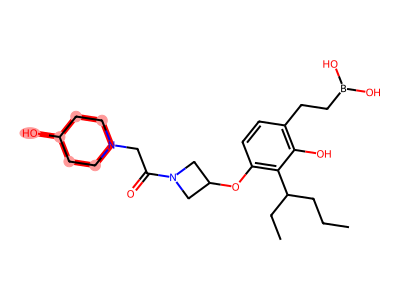} & \includegraphics[height=1.6cm]{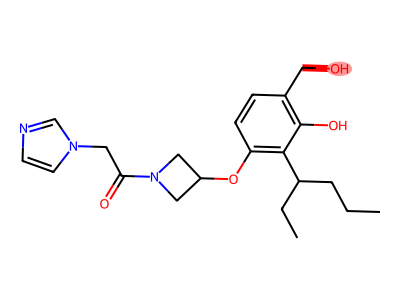} \\[-2pt]
{\scriptsize 0.41} & {\scriptsize 0.25 (\textcolor{red}{-0.15}) Sim 0.71} & {\scriptsize \textbf{0.72} (\textcolor{green!50!black}{+0.32}) Sim 0.53} & {\scriptsize 0.47} & {\scriptsize 0.37 (\textcolor{red}{-0.10}) Sim 0.70} & {\scriptsize \textbf{0.69} (\textcolor{green!50!black}{+0.22}) Sim 0.81} \\[6pt]
\includegraphics[height=1.6cm]{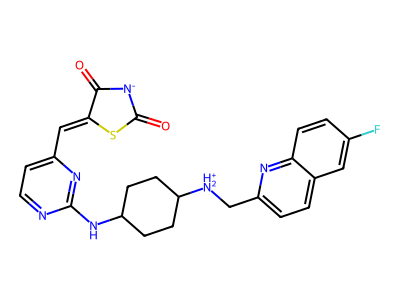} & \includegraphics[height=1.6cm]{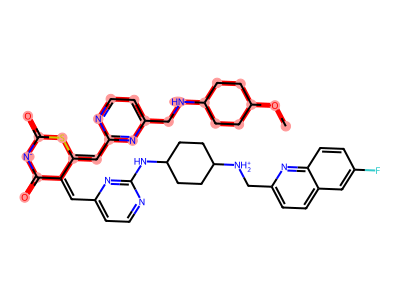} & \includegraphics[height=1.6cm]{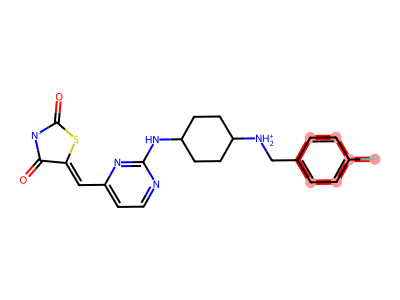} & \includegraphics[height=1.6cm]{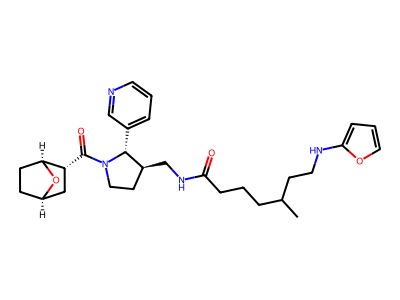} & \includegraphics[height=1.6cm]{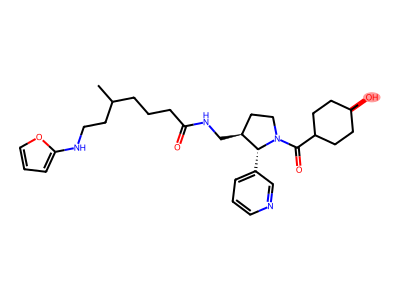} & \includegraphics[height=1.6cm]{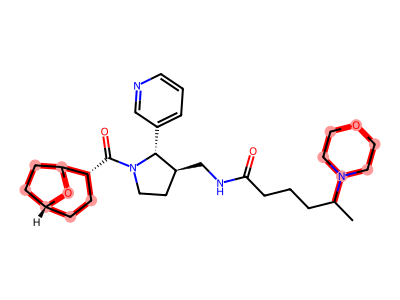} \\[-2pt]
{\scriptsize 0.52} & {\scriptsize 0.18 (\textcolor{red}{-0.34}) Sim 0.57} & {\scriptsize \textbf{0.69} (\textcolor{green!50!black}{+0.17}) Sim 0.75} & {\scriptsize 0.44} & {\scriptsize 0.39 (\textcolor{red}{-0.05}) Sim 0.76} & {\scriptsize \textbf{0.56} (\textcolor{green!50!black}{+0.13}) Sim 0.66} \\[6pt]
\includegraphics[height=1.6cm]{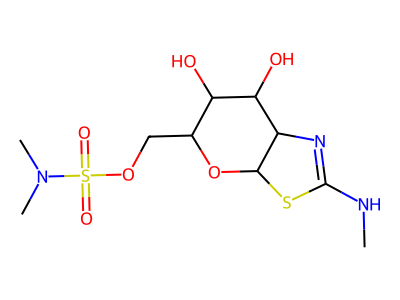} & \includegraphics[height=1.6cm]{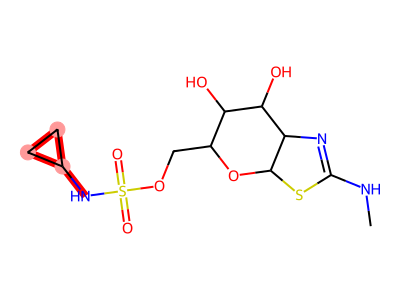} & \includegraphics[height=1.6cm]{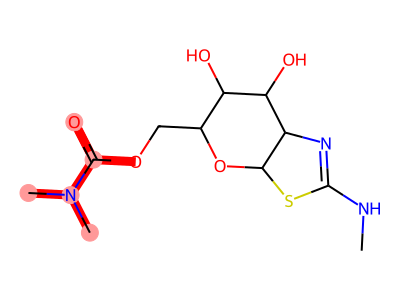} & \includegraphics[height=1.6cm]{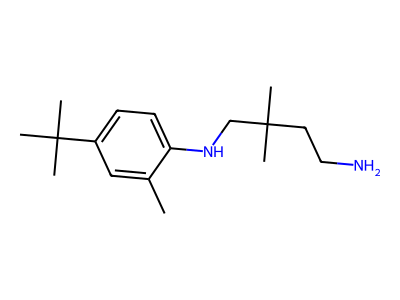} & \includegraphics[height=1.6cm]{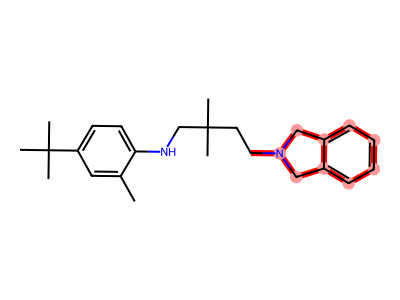} & \includegraphics[height=1.6cm]{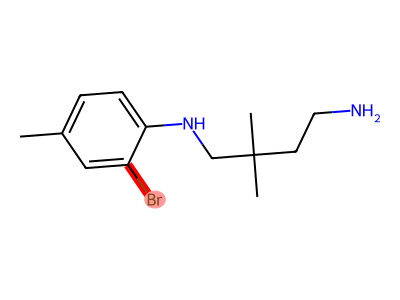} \\
\end{tabular}
\caption{\method steering examples: QED$\uparrow$ (GeLLM3O-Mistral).}
\label{fig:mi_qed}
\end{figure*}

\begin{figure*}[ht]
\centering
\setlength{\tabcolsep}{1pt}
\begin{tabular}{ccc@{\hspace{8pt}}ccc}
\textbf{Source} & \textbf{SFT} & \textbf{+\method} & \textbf{Source} & \textbf{SFT} & \textbf{+\method} \\[2pt]
\includegraphics[height=1.6cm]{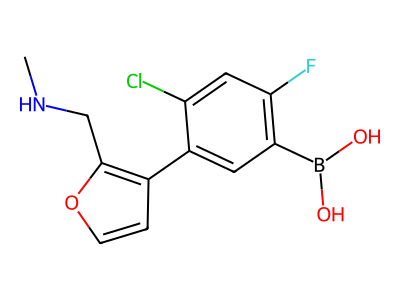} & \includegraphics[height=1.6cm]{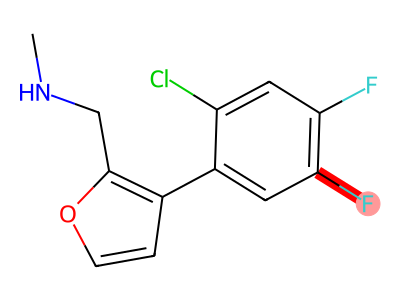} & \includegraphics[height=1.6cm]{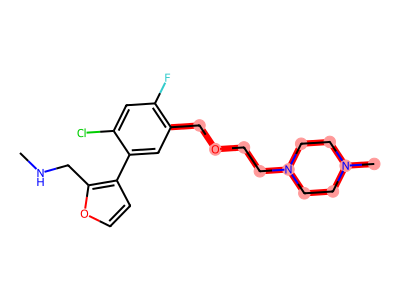} & \includegraphics[height=1.6cm]{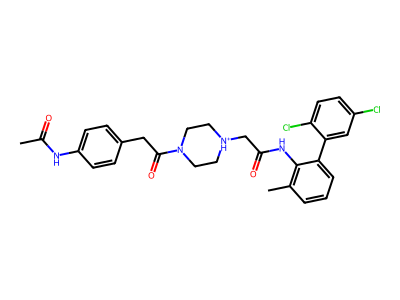} & \includegraphics[height=1.6cm]{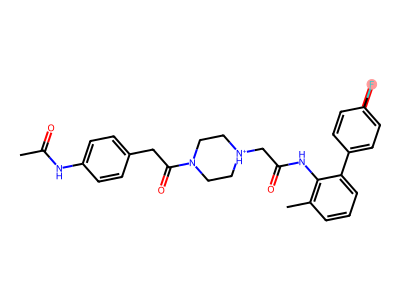} & \includegraphics[height=1.6cm]{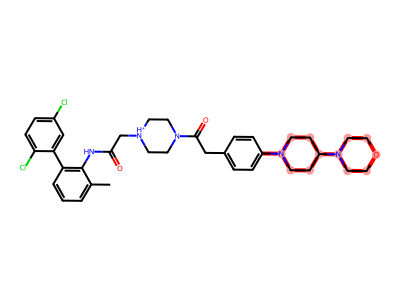} \\[-2pt]
{\scriptsize 0.17} & {\scriptsize 0.04 (\textcolor{red}{-0.13}) Sim 0.76} & {\scriptsize \textbf{0.46} (\textcolor{green!50!black}{+0.29}) Sim 0.56} & {\scriptsize 0.05} & {\scriptsize 0.02 (\textcolor{red}{-0.03}) Sim 0.71} & {\scriptsize \textbf{0.24} (\textcolor{green!50!black}{+0.19}) Sim 0.70} \\[6pt]
\includegraphics[height=1.6cm]{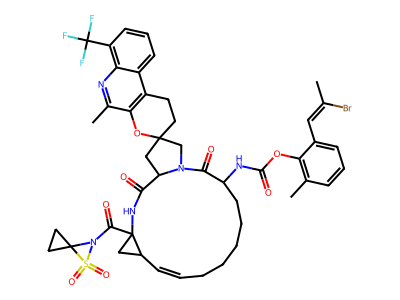} & \includegraphics[height=1.6cm]{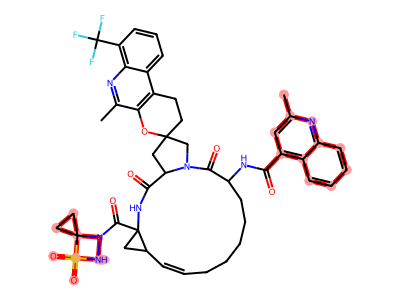} & \includegraphics[height=1.6cm]{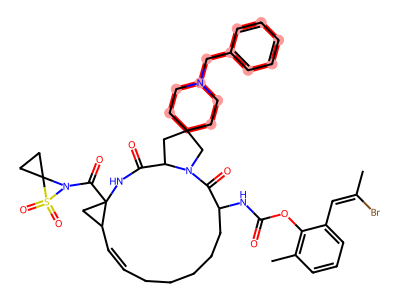} & \includegraphics[height=1.6cm]{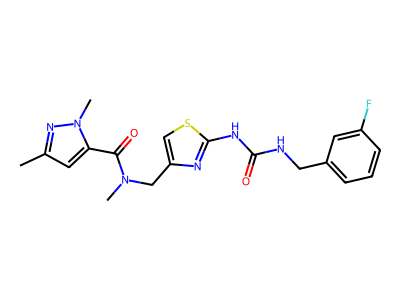} & \includegraphics[height=1.6cm]{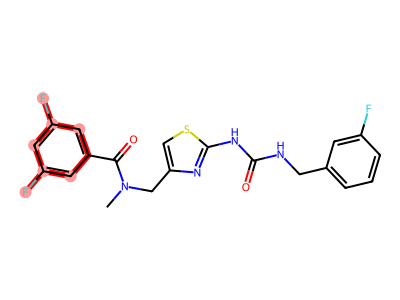} & \includegraphics[height=1.6cm]{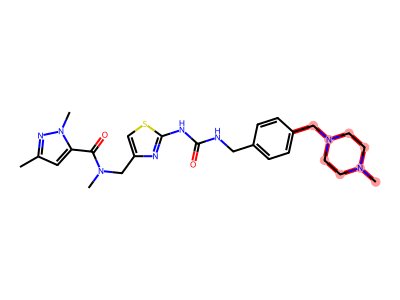} \\[-2pt]
{\scriptsize 0.20} & {\scriptsize 0.19 (\textcolor{red}{-0.01}) Sim 0.62} & {\scriptsize \textbf{0.36} (\textcolor{green!50!black}{+0.16}) Sim 0.63} & {\scriptsize 0.04} & {\scriptsize 0.02 (\textcolor{red}{-0.02}) Sim 0.72} & {\scriptsize \textbf{0.16} (\textcolor{green!50!black}{+0.12}) Sim 0.69} \\[6pt]
\includegraphics[height=1.6cm]{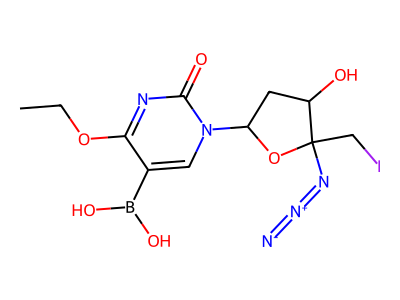} & \includegraphics[height=1.6cm]{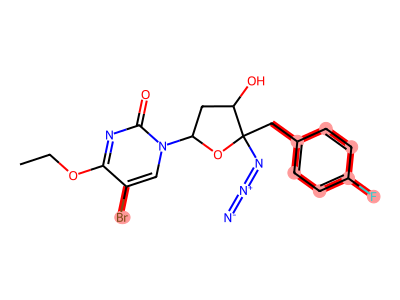} & \includegraphics[height=1.6cm]{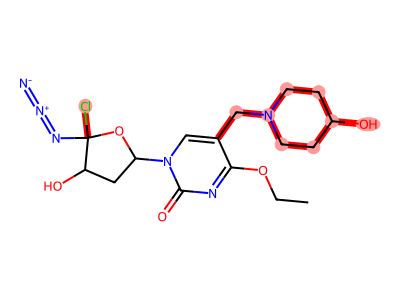} & \includegraphics[height=1.6cm]{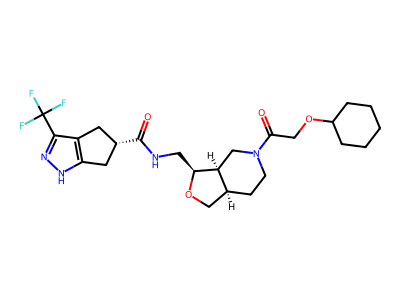} & \includegraphics[height=1.6cm]{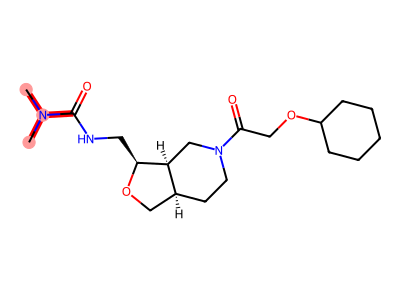} & \includegraphics[height=1.6cm]{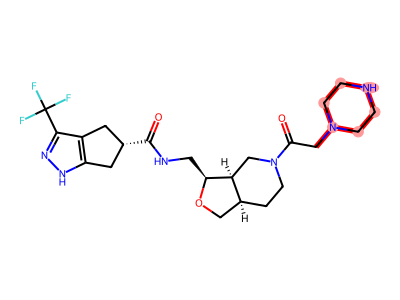} \\
\end{tabular}
\caption{\method steering examples: DRD2$\uparrow$ (GeLLM3O-Mistral).}
\label{fig:mi_drd2}
\end{figure*}

\begin{figure*}[ht]
\centering
\setlength{\tabcolsep}{1pt}
\begin{tabular}{ccc@{\hspace{8pt}}ccc}
\textbf{Source} & \textbf{SFT} & \textbf{+\method} & \textbf{Source} & \textbf{SFT} & \textbf{+\method} \\[2pt]
\includegraphics[height=1.6cm]{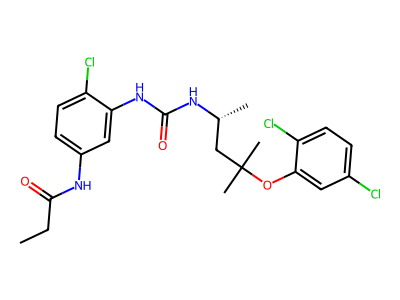} & \includegraphics[height=1.6cm]{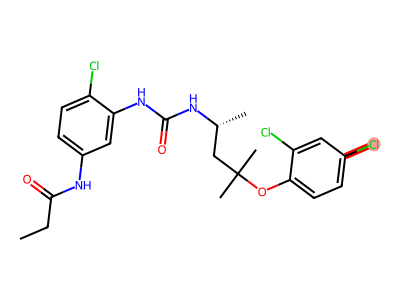} & \includegraphics[height=1.6cm]{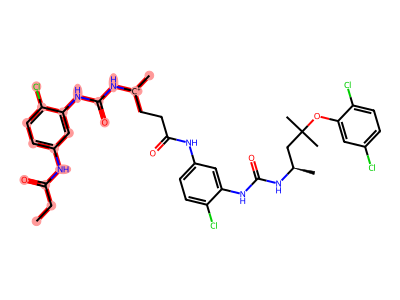} & \includegraphics[height=1.6cm]{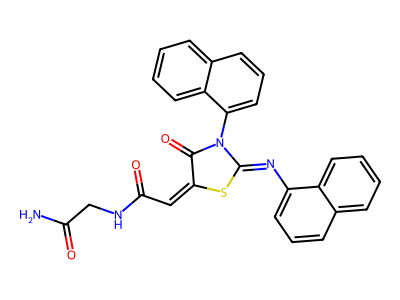} & \includegraphics[height=1.6cm]{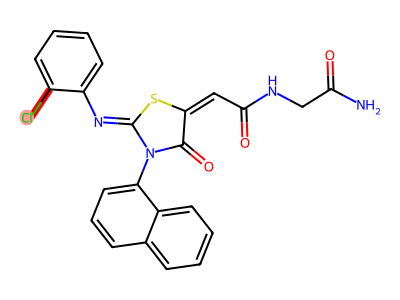} & \includegraphics[height=1.6cm]{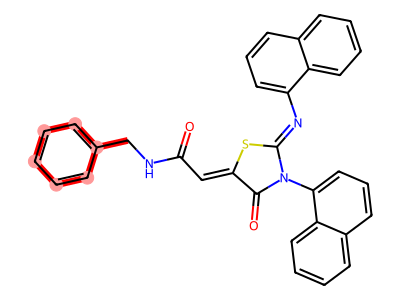} \\[-2pt]
{\scriptsize 6.75} & {\scriptsize 6.75 (\textcolor{red}{-0.00}) Sim 0.89} & {\scriptsize \textbf{9.50} (\textcolor{green!50!black}{+2.75}) Sim 0.81} & {\scriptsize 4.25} & {\scriptsize 3.75 (\textcolor{red}{-0.50}) Sim 0.88} & {\scriptsize \textbf{6.96} (\textcolor{green!50!black}{+2.71}) Sim 0.78} \\[6pt]
\includegraphics[height=1.6cm]{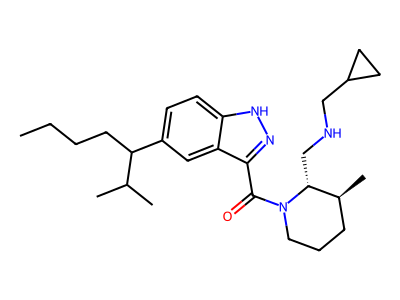} & \includegraphics[height=1.6cm]{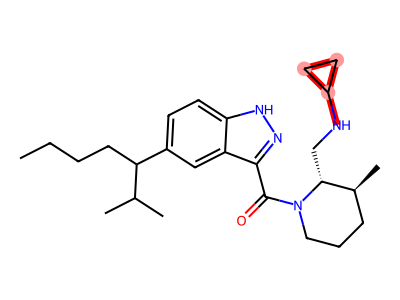} & \includegraphics[height=1.6cm]{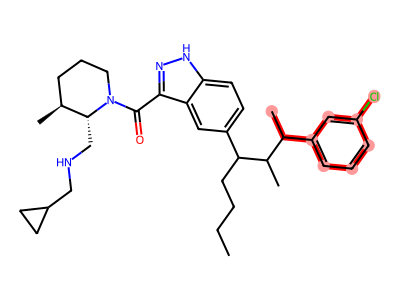} & \includegraphics[height=1.6cm]{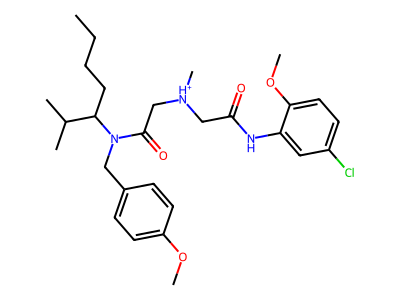} & \includegraphics[height=1.6cm]{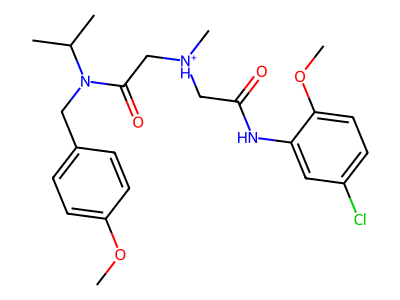} & \includegraphics[height=1.6cm]{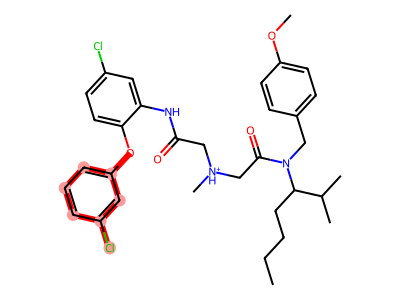} \\[-2pt]
{\scriptsize 5.73} & {\scriptsize 5.49 (\textcolor{red}{-0.25}) Sim 0.83} & {\scriptsize \textbf{8.17} (\textcolor{green!50!black}{+2.44}) Sim 0.78} & {\scriptsize 4.05} & {\scriptsize 2.25 (\textcolor{red}{-1.81}) Sim 0.77} & {\scriptsize \textbf{6.49} (\textcolor{green!50!black}{+2.44}) Sim 0.85} \\[6pt]
\includegraphics[height=1.6cm]{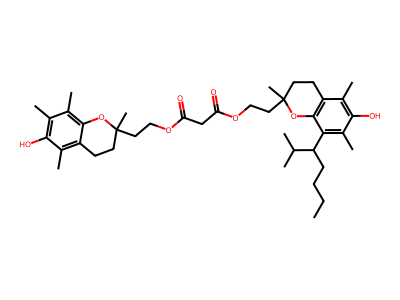} & \includegraphics[height=1.6cm]{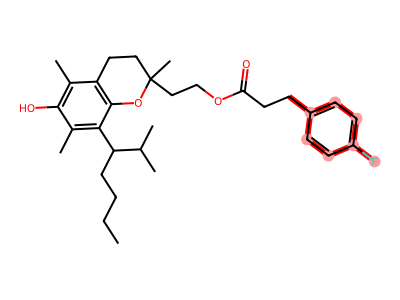} & \includegraphics[height=1.6cm]{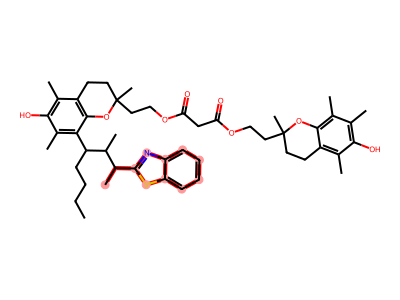} & \includegraphics[height=1.6cm]{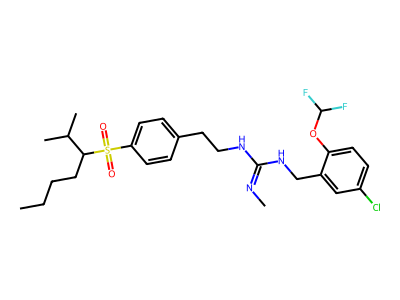} & \includegraphics[height=1.6cm]{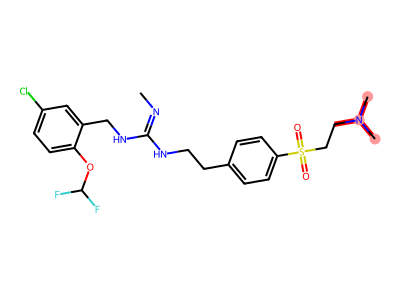} & \includegraphics[height=1.6cm]{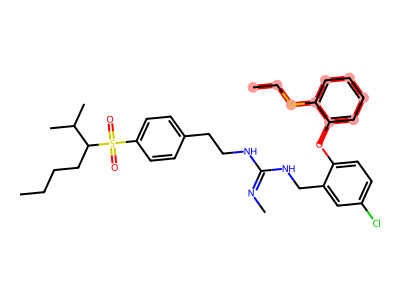} \\
\end{tabular}
\caption{\method steering examples: logP$\uparrow$ (GeLLM3O-Mistral).}
\label{fig:mi_logp}
\end{figure*}

\begin{figure*}[ht]
\centering
\setlength{\tabcolsep}{1pt}
\begin{tabular}{ccc@{\hspace{8pt}}ccc}
\textbf{Source} & \textbf{SFT} & \textbf{+\method} & \textbf{Source} & \textbf{SFT} & \textbf{+\method} \\[2pt]
\includegraphics[height=1.6cm]{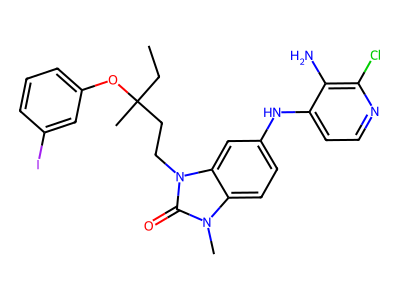} & \includegraphics[height=1.6cm]{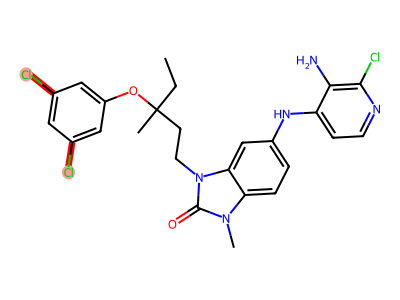} & \includegraphics[height=1.6cm]{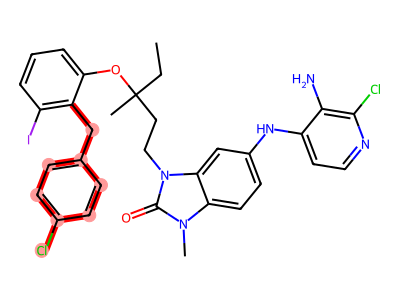} & \includegraphics[height=1.6cm]{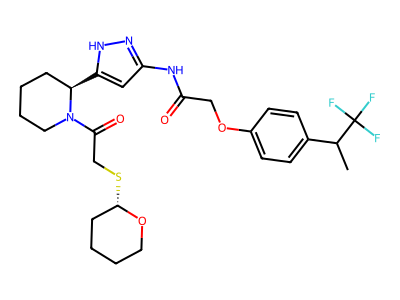} & \includegraphics[height=1.6cm]{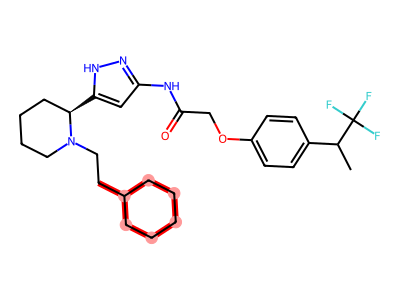} & \includegraphics[height=1.6cm]{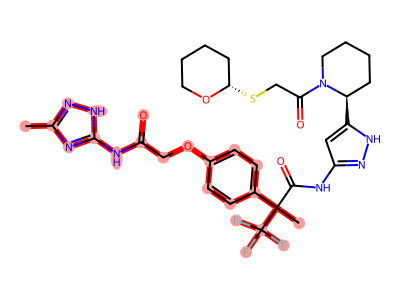} \\[-2pt]
{\scriptsize 592} & {\scriptsize 535 (\textcolor{red}{-57}) Sim 0.79} & {\scriptsize \textbf{716} (\textcolor{green!50!black}{+125}) Sim 0.71} & {\scriptsize 555} & {\scriptsize 507 (\textcolor{red}{-48}) Sim 0.64} & {\scriptsize \textbf{679} (\textcolor{green!50!black}{+124}) Sim 0.68} \\[6pt]
\includegraphics[height=1.6cm]{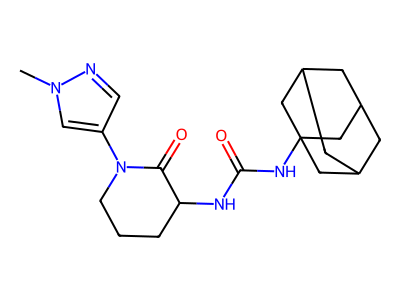} & \includegraphics[height=1.6cm]{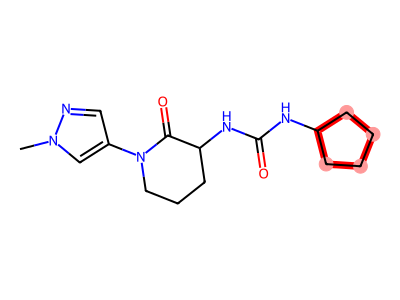} & \includegraphics[height=1.6cm]{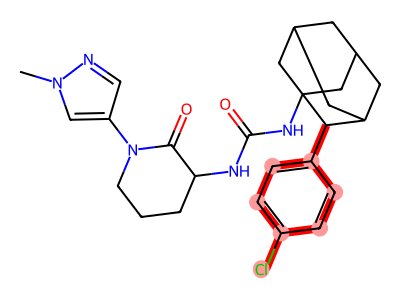} & \includegraphics[height=1.6cm]{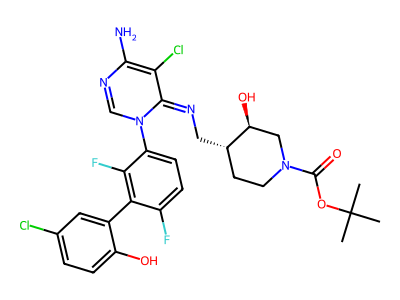} & \includegraphics[height=1.6cm]{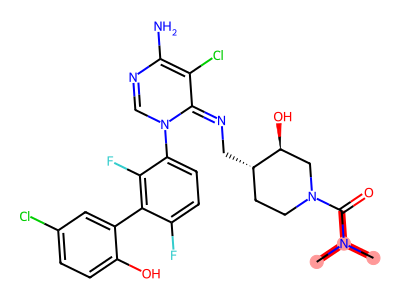} & \includegraphics[height=1.6cm]{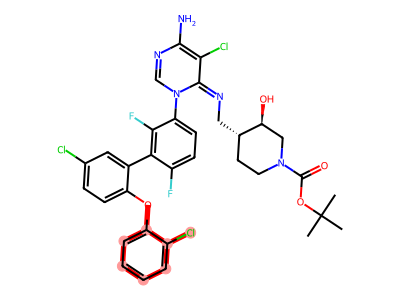} \\[-2pt]
{\scriptsize 371} & {\scriptsize 305 (\textcolor{red}{-66}) Sim 0.67} & {\scriptsize \textbf{482} (\textcolor{green!50!black}{+111}) Sim 0.68} & {\scriptsize 596} & {\scriptsize 567 (\textcolor{red}{-29}) Sim 0.81} & {\scriptsize \textbf{707} (\textcolor{green!50!black}{+111}) Sim 0.80} \\[6pt]
\includegraphics[height=1.6cm]{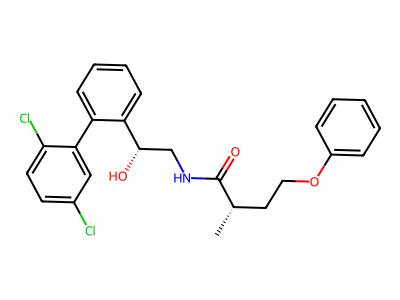} & \includegraphics[height=1.6cm]{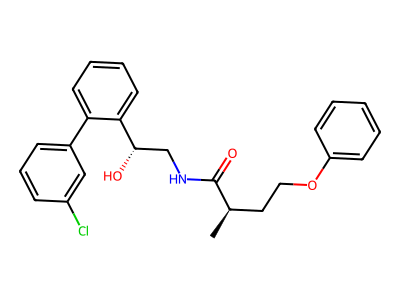} & \includegraphics[height=1.6cm]{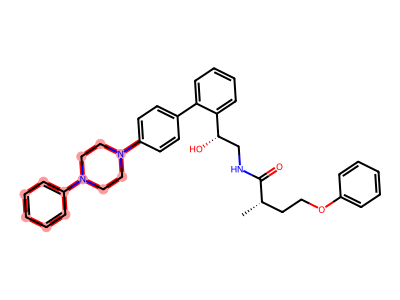} & \includegraphics[height=1.6cm]{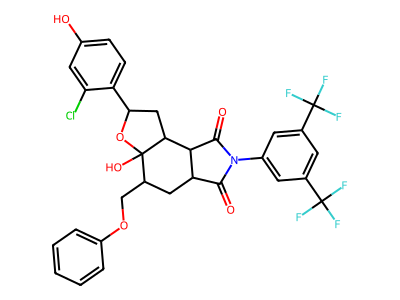} & \includegraphics[height=1.6cm]{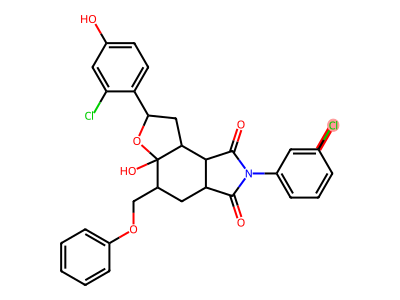} & \includegraphics[height=1.6cm]{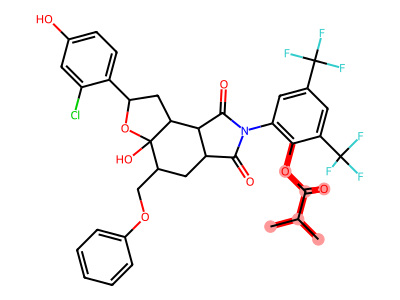} \\[-2pt]
{\scriptsize 458} & {\scriptsize 424 (\textcolor{red}{-34}) Sim 0.76} & {\scriptsize \textbf{550} (\textcolor{green!50!black}{+91}) Sim 0.61} & {\scriptsize 656} & {\scriptsize 554 (\textcolor{red}{-102}) Sim 0.80} & {\scriptsize \textbf{742} (\textcolor{green!50!black}{+86}) Sim 0.72} \\
\end{tabular}
\caption{\method steering examples: MW$\uparrow$ (GeLLM3O-Mistral).}
\label{fig:mi_mw}
\end{figure*}

\begin{figure*}[ht]
\centering
\setlength{\tabcolsep}{1pt}
\begin{tabular}{ccc@{\hspace{8pt}}ccc}
\textbf{Source} & \textbf{SFT} & \textbf{+\method} & \textbf{Source} & \textbf{SFT} & \textbf{+\method} \\[2pt]
\includegraphics[height=1.6cm]{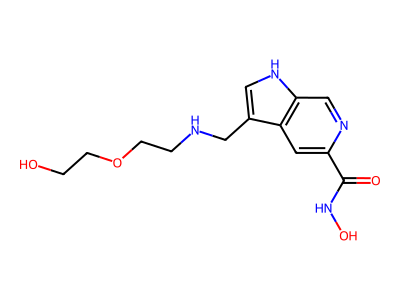} & \includegraphics[height=1.6cm]{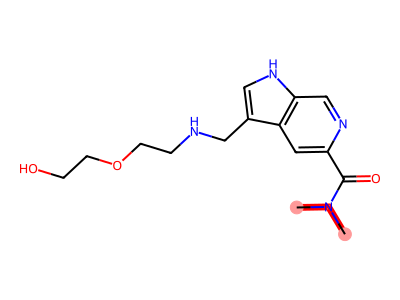} & \includegraphics[height=1.6cm]{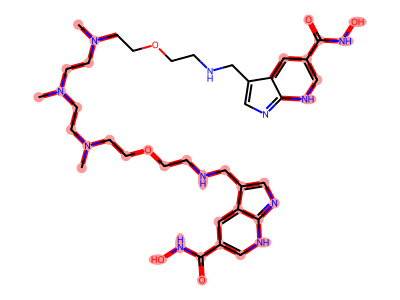} & \includegraphics[height=1.6cm]{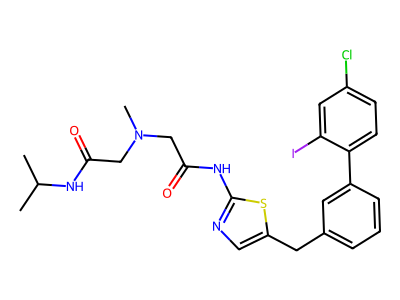} & \includegraphics[height=1.6cm]{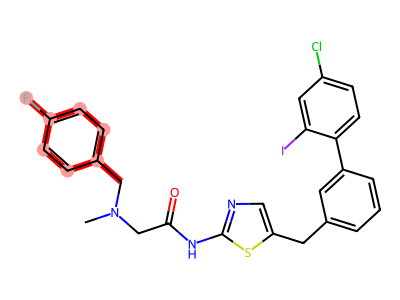} & \includegraphics[height=1.6cm]{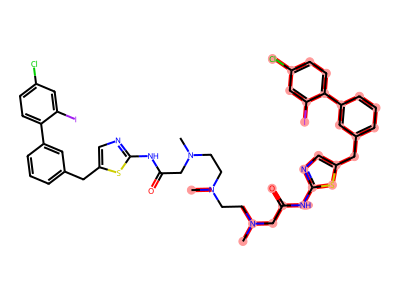} \\[-2pt]
{\scriptsize 8} & {\scriptsize 8 (\textcolor{red}{+0}) Sim 0.75} & {\scriptsize \textbf{24} (\textcolor{green!50!black}{+16}) Sim 0.43} & {\scriptsize 9} & {\scriptsize 8 (\textcolor{red}{-1}) Sim 0.76} & {\scriptsize \textbf{18} (\textcolor{green!50!black}{+9}) Sim 0.82} \\[6pt]
\includegraphics[height=1.6cm]{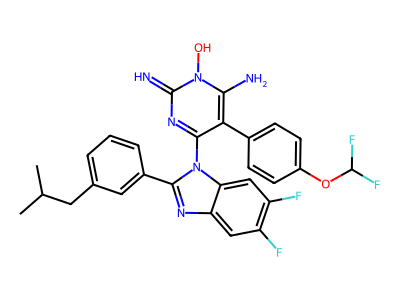} & \includegraphics[height=1.6cm]{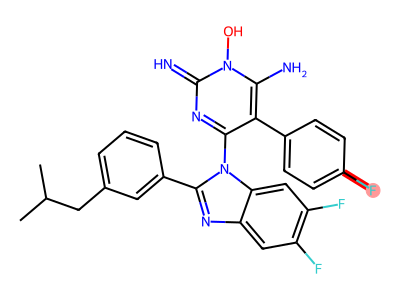} & \includegraphics[height=1.6cm]{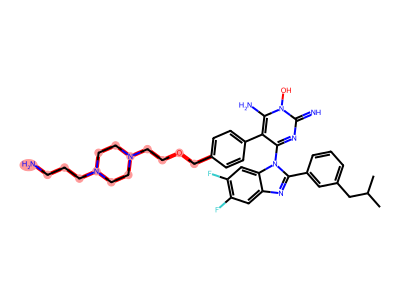} & \includegraphics[height=1.6cm]{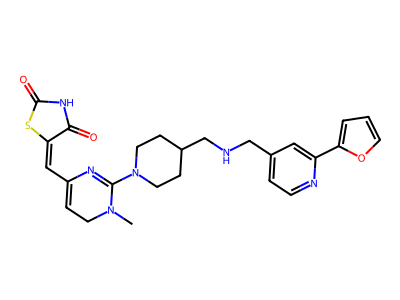} & \includegraphics[height=1.6cm]{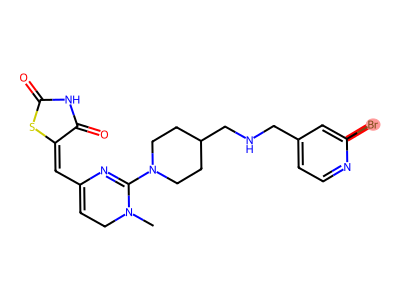} & \includegraphics[height=1.6cm]{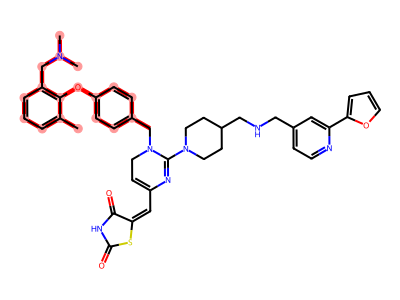} \\[-2pt]
{\scriptsize 7} & {\scriptsize 5 (\textcolor{red}{-2}) Sim 0.83} & {\scriptsize \textbf{13} (\textcolor{green!50!black}{+6}) Sim 0.69} & {\scriptsize 6} & {\scriptsize 5 (\textcolor{red}{-1}) Sim 0.77} & {\scriptsize \textbf{12} (\textcolor{green!50!black}{+6}) Sim 0.66} \\[6pt]
\includegraphics[height=1.6cm]{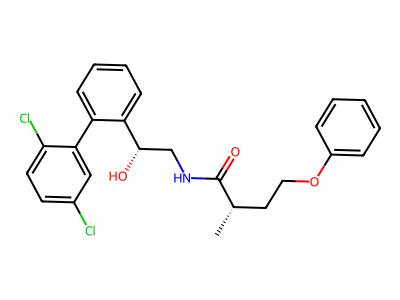} & \includegraphics[height=1.6cm]{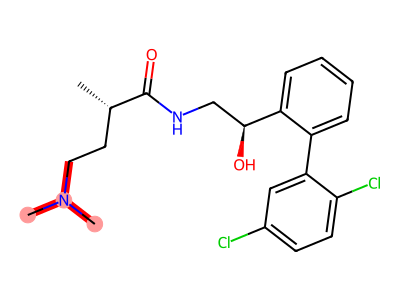} & \includegraphics[height=1.6cm]{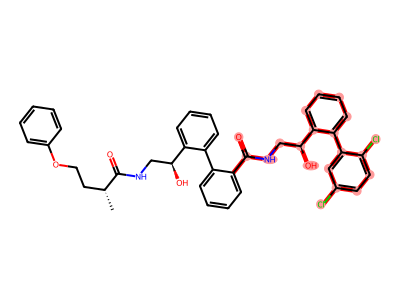} & \includegraphics[height=1.6cm]{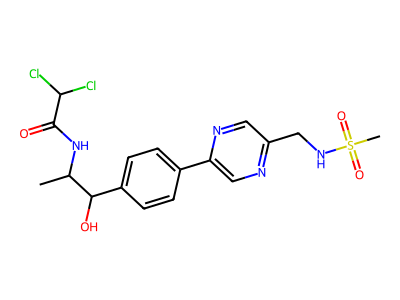} & \includegraphics[height=1.6cm]{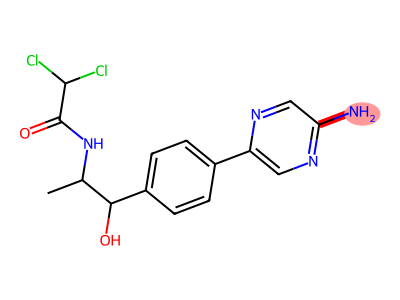} & \includegraphics[height=1.6cm]{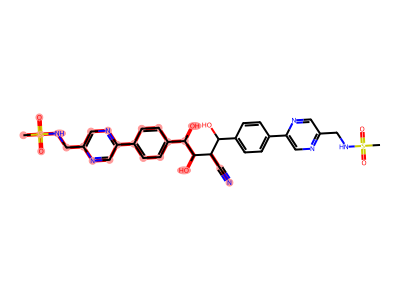} \\
\end{tabular}
\caption{\method steering examples: RotBond$\uparrow$ (GeLLM3O-Mistral).}
\label{fig:mi_rotbond}
\end{figure*}

\begin{figure*}[ht]
\centering
\setlength{\tabcolsep}{1pt}
\begin{tabular}{ccc@{\hspace{8pt}}ccc}
\textbf{Source} & \textbf{SFT} & \textbf{+\method} & \textbf{Source} & \textbf{SFT} & \textbf{+\method} \\[2pt]
\includegraphics[height=1.6cm]{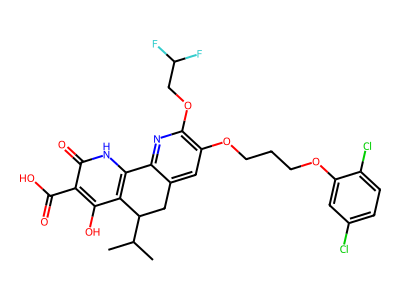} & \includegraphics[height=1.6cm]{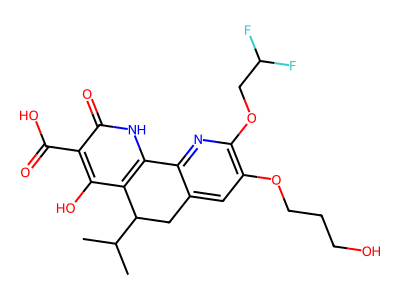} & \includegraphics[height=1.6cm]{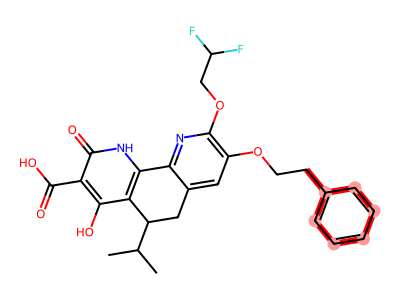} & \includegraphics[height=1.6cm]{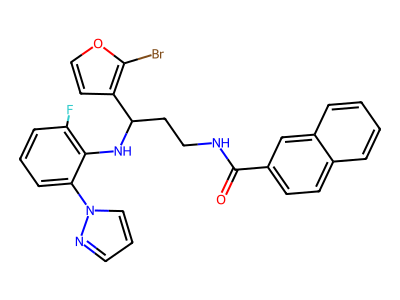} & \includegraphics[height=1.6cm]{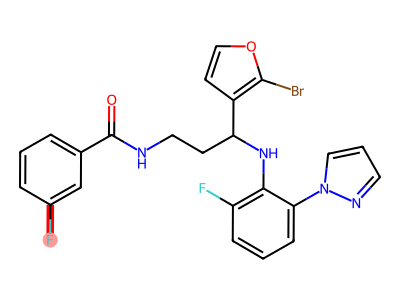} & \includegraphics[height=1.6cm]{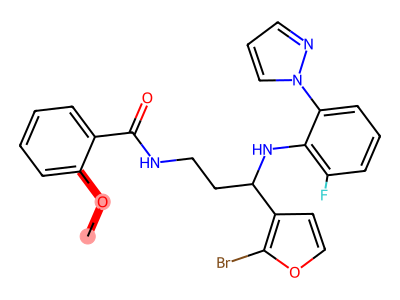} \\[-2pt]
{\scriptsize 3.80} & {\scriptsize 3.82 (\textcolor{red}{-0.02}) Sim 0.79} & {\scriptsize \textbf{3.65} (\textcolor{green!50!black}{+0.15}) Sim 0.75} & {\scriptsize 3.28} & {\scriptsize 3.30 (\textcolor{red}{-0.02}) Sim 0.83} & {\scriptsize \textbf{3.27} (\textcolor{green!50!black}{+0.00}) Sim 0.70} \\
\end{tabular}
\caption{\method steering examples: SA$\downarrow$ (GeLLM3O-Mistral).}
\label{fig:mi_sa}
\end{figure*}

\begin{figure*}[ht]
\centering
\setlength{\tabcolsep}{1pt}
\begin{tabular}{ccc@{\hspace{8pt}}ccc}
\textbf{Source} & \textbf{SFT} & \textbf{+\method} & \textbf{Source} & \textbf{SFT} & \textbf{+\method} \\[2pt]
\includegraphics[height=1.6cm]{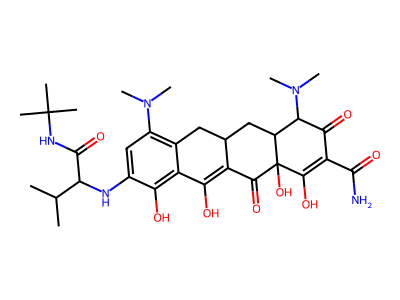} & \includegraphics[height=1.6cm]{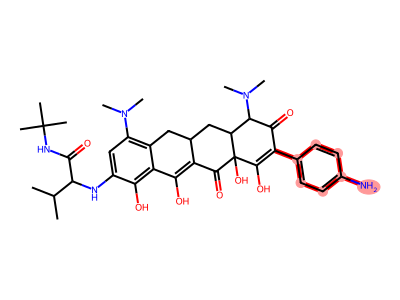} & \includegraphics[height=1.6cm]{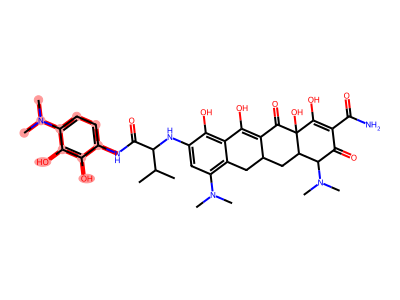} & \includegraphics[height=1.6cm]{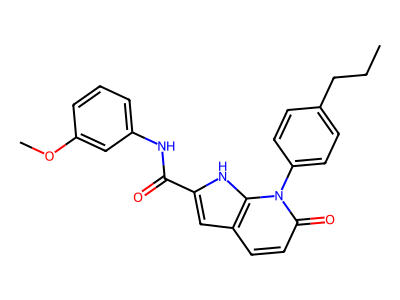} & \includegraphics[height=1.6cm]{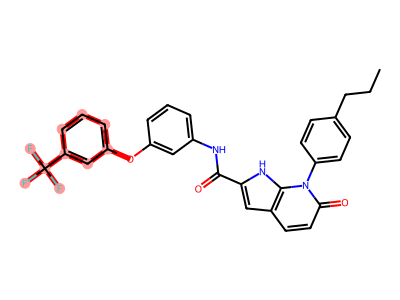} & \includegraphics[height=1.6cm]{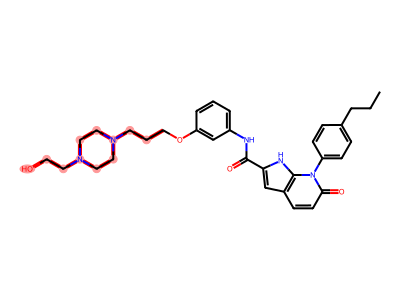} \\[-2pt]
{\scriptsize 11} & {\scriptsize 11 (\textcolor{red}{+0}) Sim 0.79} & {\scriptsize \textbf{14} (\textcolor{green!50!black}{+3}) Sim 0.80} & {\scriptsize 4} & {\scriptsize 4 (\textcolor{red}{+0}) Sim 0.76} & {\scriptsize \textbf{7} (\textcolor{green!50!black}{+3}) Sim 0.71} \\[6pt]
\includegraphics[height=1.6cm]{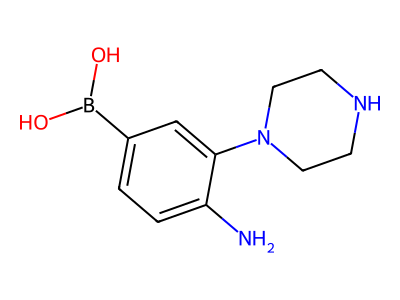} & \includegraphics[height=1.6cm]{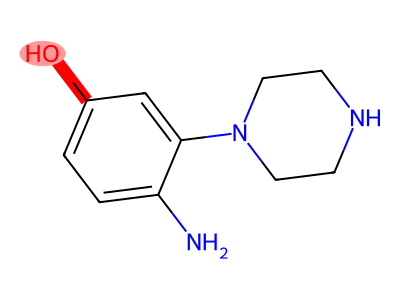} & \includegraphics[height=1.6cm]{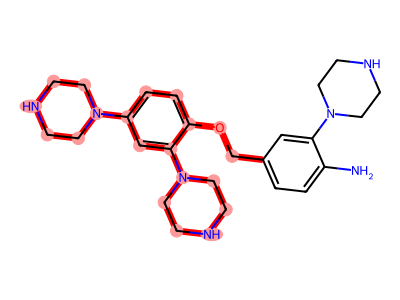} & \includegraphics[height=1.6cm]{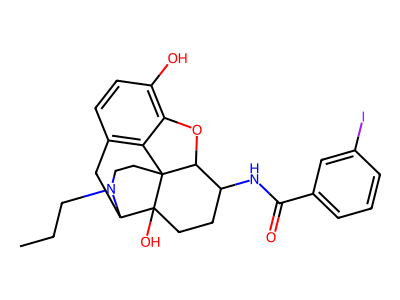} & \includegraphics[height=1.6cm]{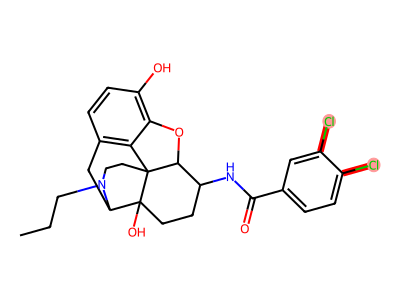} & \includegraphics[height=1.6cm]{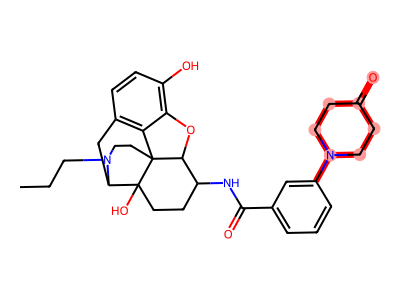} \\[-2pt]
{\scriptsize 5} & {\scriptsize 4 (\textcolor{red}{-1}) Sim 0.64} & {\scriptsize \textbf{8} (\textcolor{green!50!black}{+3}) Sim 0.44} & {\scriptsize 5} & {\scriptsize 5 (\textcolor{red}{+0}) Sim 0.79} & {\scriptsize \textbf{7} (\textcolor{green!50!black}{+2}) Sim 0.78} \\[6pt]
\includegraphics[height=1.6cm]{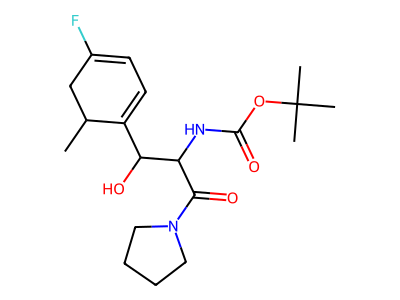} & \includegraphics[height=1.6cm]{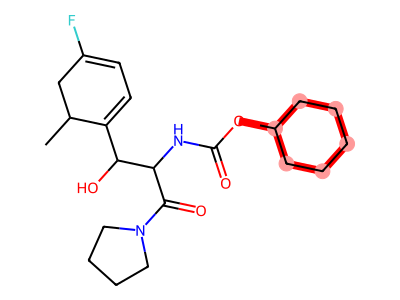} & \includegraphics[height=1.6cm]{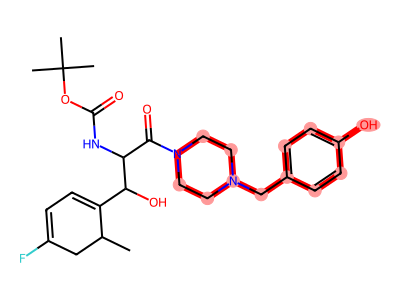} & \includegraphics[height=1.6cm]{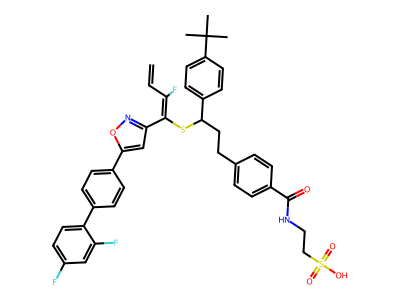} & \includegraphics[height=1.6cm]{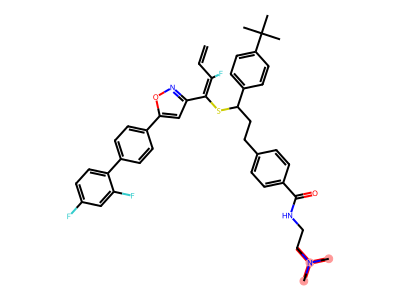} & \includegraphics[height=1.6cm]{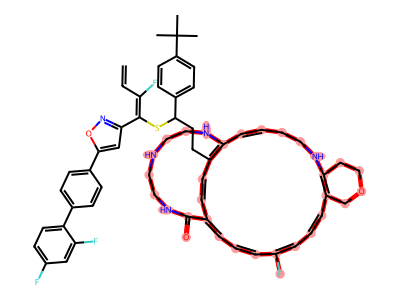} \\
\end{tabular}
\caption{\method steering examples: HBA$\uparrow$ (GeLLM3O-Mistral).}
\label{fig:mi_acceptor}
\end{figure*}

\begin{figure*}[ht]
\centering
\setlength{\tabcolsep}{1pt}
\begin{tabular}{ccc@{\hspace{8pt}}ccc}
\textbf{Source} & \textbf{SFT} & \textbf{+\method} & \textbf{Source} & \textbf{SFT} & \textbf{+\method} \\[2pt]
\includegraphics[height=1.6cm]{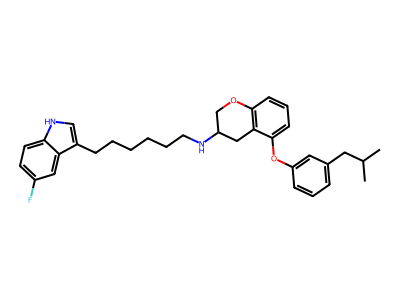} & \includegraphics[height=1.6cm]{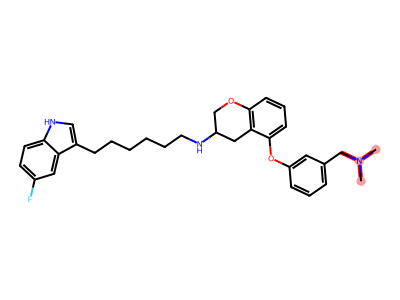} & \includegraphics[height=1.6cm]{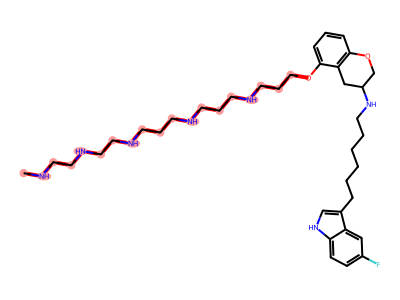} & \includegraphics[height=1.6cm]{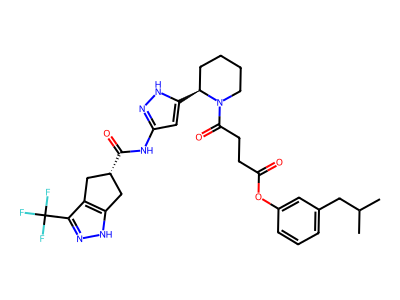} & \includegraphics[height=1.6cm]{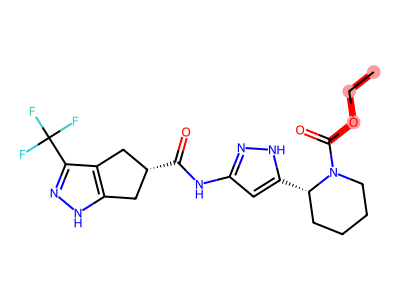} & \includegraphics[height=1.6cm]{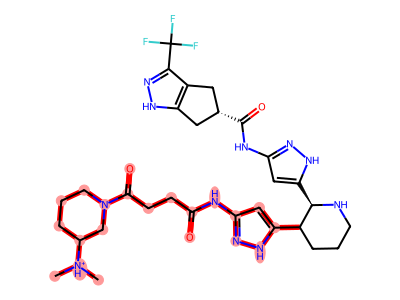} \\[-2pt]
{\scriptsize 2} & {\scriptsize 2 (\textcolor{red}{+0}) Sim 0.82} & {\scriptsize \textbf{7} (\textcolor{green!50!black}{+5}) Sim 0.65} & {\scriptsize 3} & {\scriptsize 3 (\textcolor{red}{+0}) Sim 0.62} & {\scriptsize \textbf{7} (\textcolor{green!50!black}{+4}) Sim 0.45} \\[6pt]
\includegraphics[height=1.6cm]{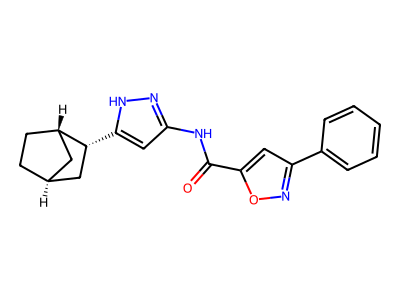} & \includegraphics[height=1.6cm]{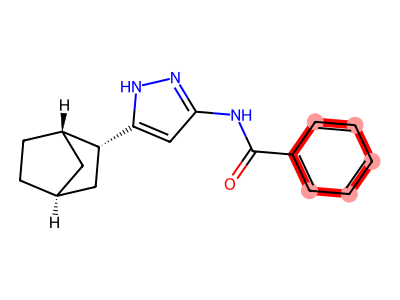} & \includegraphics[height=1.6cm]{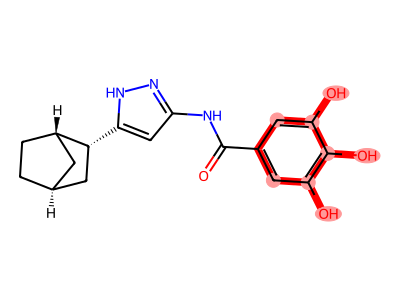} & \includegraphics[height=1.6cm]{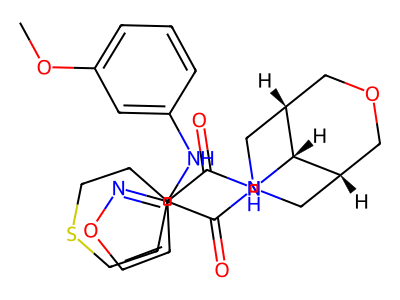} & \includegraphics[height=1.6cm]{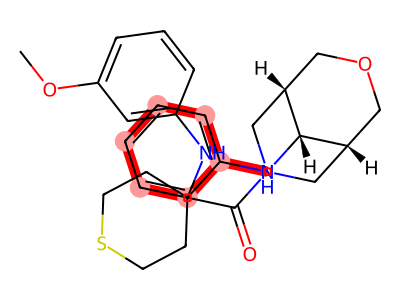} & \includegraphics[height=1.6cm]{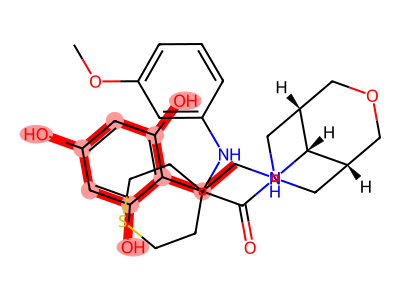} \\[-2pt]
{\scriptsize 2} & {\scriptsize 2 (\textcolor{red}{+0}) Sim 0.68} & {\scriptsize \textbf{5} (\textcolor{green!50!black}{+3}) Sim 0.57} & {\scriptsize 2} & {\scriptsize 2 (\textcolor{red}{+0}) Sim 0.70} & {\scriptsize \textbf{5} (\textcolor{green!50!black}{+3}) Sim 0.62} \\[6pt]
\includegraphics[height=1.6cm]{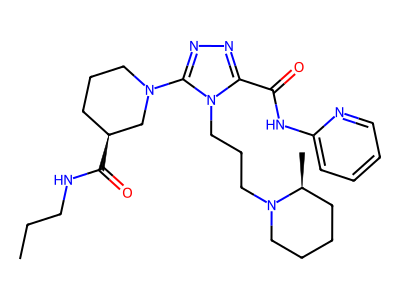} & \includegraphics[height=1.6cm]{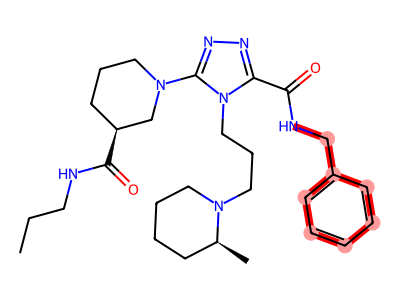} & \includegraphics[height=1.6cm]{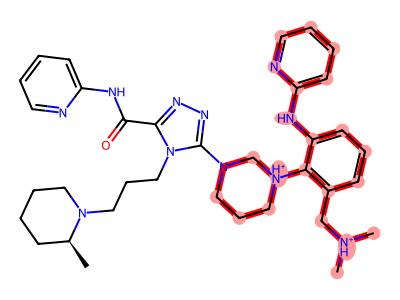} & \includegraphics[height=1.6cm]{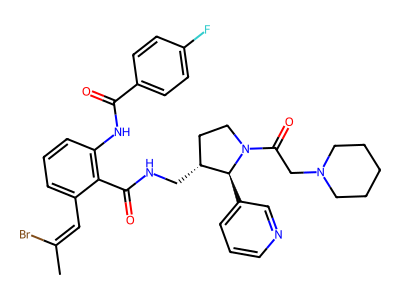} & \includegraphics[height=1.6cm]{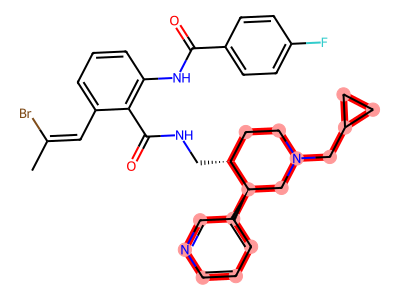} & \includegraphics[height=1.6cm]{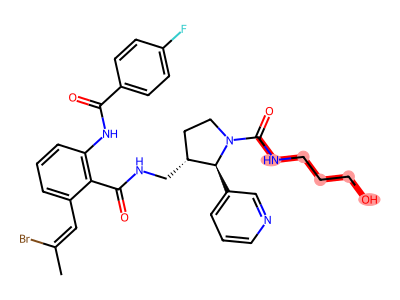} \\[-2pt]
{\scriptsize 2} & {\scriptsize 2 (\textcolor{red}{+0}) Sim 0.76} & {\scriptsize \textbf{4} (\textcolor{green!50!black}{+2}) Sim 0.57} & {\scriptsize 2} & {\scriptsize 2 (\textcolor{red}{+0}) Sim 0.69} & {\scriptsize \textbf{4} (\textcolor{green!50!black}{+2}) Sim 0.75} \\
\end{tabular}
\caption{\method steering examples: HBD$\uparrow$ (GeLLM3O-Mistral).}
\label{fig:mi_donor}
\end{figure*}

\subsubsection*{GeLLM3O-LLaMA3}

\begin{figure*}[ht]
\centering
\setlength{\tabcolsep}{1pt}
\begin{tabular}{ccc@{\hspace{8pt}}ccc}
\textbf{Source} & \textbf{SFT} & \textbf{+\method} & \textbf{Source} & \textbf{SFT} & \textbf{+\method} \\[2pt]
\includegraphics[height=1.6cm]{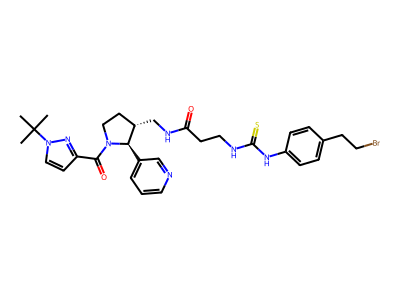} & \includegraphics[height=1.6cm]{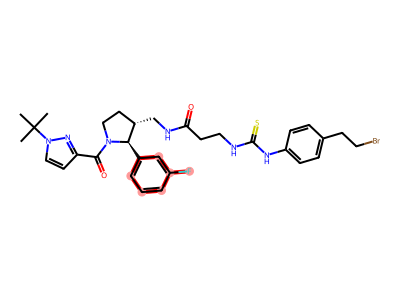} & \includegraphics[height=1.6cm]{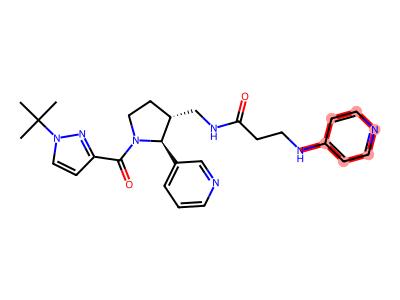} & \includegraphics[height=1.6cm]{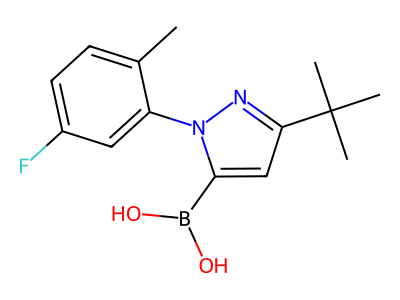} & \includegraphics[height=1.6cm]{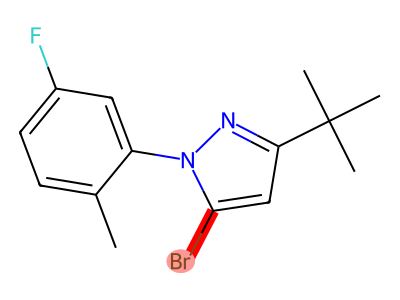} & \includegraphics[height=1.6cm]{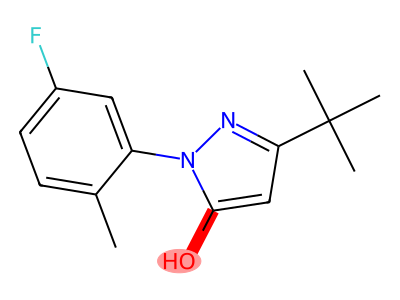} \\[-2pt]
{\scriptsize 0.22} & {\scriptsize 0.20 (\textcolor{red}{-0.02}) Sim 0.80} & {\scriptsize \textbf{0.52} (\textcolor{green!50!black}{+0.30}) Sim 0.69} & {\scriptsize 0.81} & {\scriptsize 0.76 (\textcolor{red}{-0.05}) Sim 0.65} & {\scriptsize \textbf{0.84} (\textcolor{green!50!black}{+0.03}) Sim 0.69} \\[6pt]
\includegraphics[height=1.6cm]{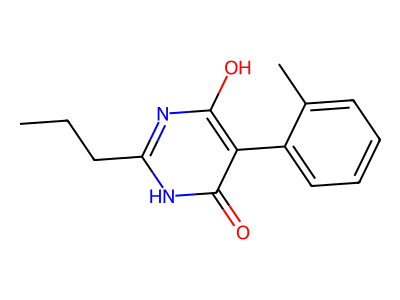} & \includegraphics[height=1.6cm]{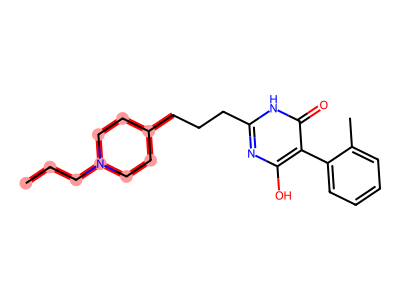} & \includegraphics[height=1.6cm]{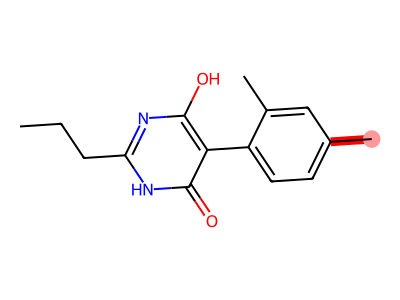} & \includegraphics[height=1.6cm]{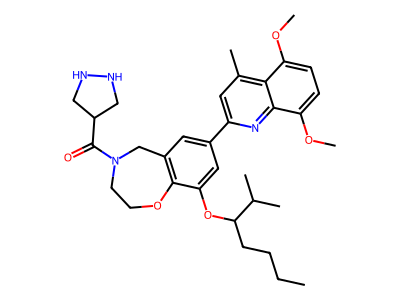} & \includegraphics[height=1.6cm]{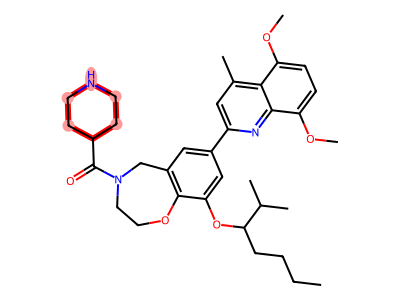} & \includegraphics[height=1.6cm]{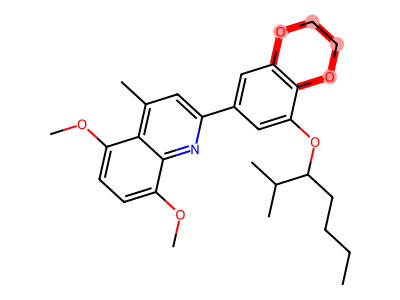} \\[-2pt]
{\scriptsize 0.87} & {\scriptsize 0.78 (\textcolor{red}{-0.09}) Sim 0.62} & {\scriptsize \textbf{0.89} (\textcolor{green!50!black}{+0.02}) Sim 0.73} & {\scriptsize 0.34} & {\scriptsize 0.29 (\textcolor{red}{-0.05}) Sim 0.87} & {\scriptsize \textbf{0.36} (\textcolor{green!50!black}{+0.02}) Sim 0.63} \\[6pt]
\includegraphics[height=1.6cm]{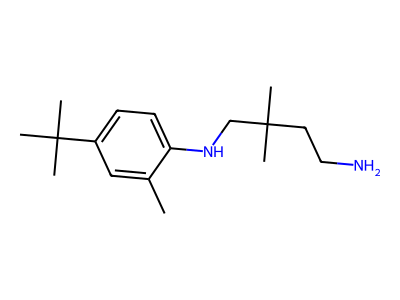} & \includegraphics[height=1.6cm]{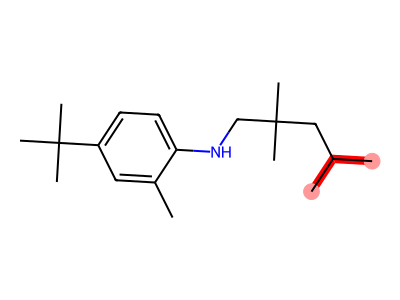} & \includegraphics[height=1.6cm]{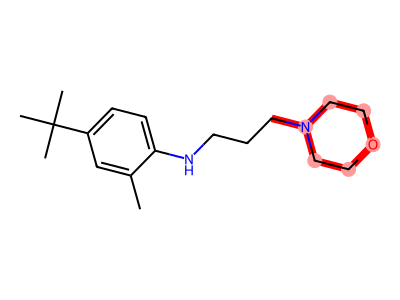} & \includegraphics[height=1.6cm]{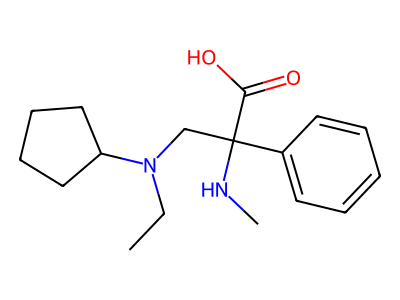} & \includegraphics[height=1.6cm]{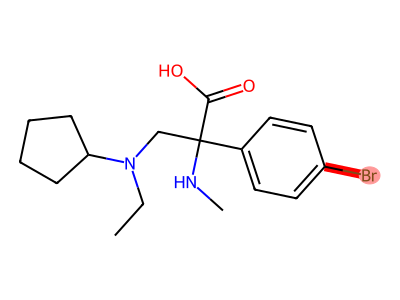} & \includegraphics[height=1.6cm]{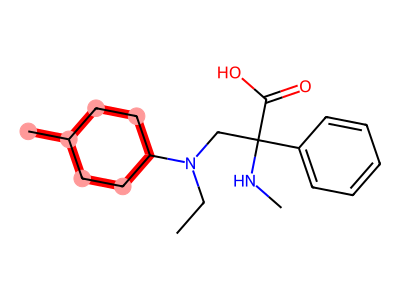} \\
\end{tabular}
\caption{\method steering examples: QED$\uparrow$ (GeLLM3O-LLaMA3).}
\label{fig:l3_qed}
\end{figure*}

\begin{figure*}[ht]
\centering
\setlength{\tabcolsep}{1pt}
\begin{tabular}{ccc@{\hspace{8pt}}ccc}
\textbf{Source} & \textbf{SFT} & \textbf{+\method} & \textbf{Source} & \textbf{SFT} & \textbf{+\method} \\[2pt]
\includegraphics[height=1.6cm]{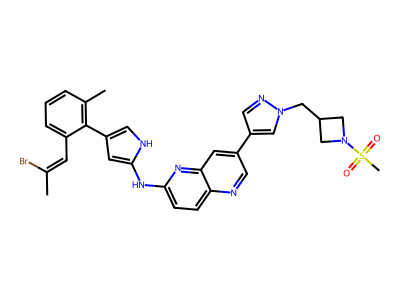} & \includegraphics[height=1.6cm]{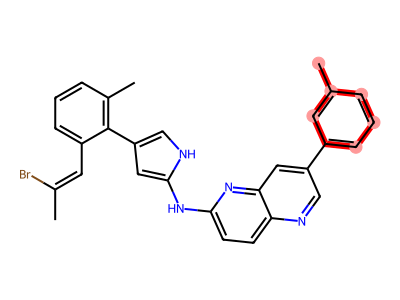} & \includegraphics[height=1.6cm]{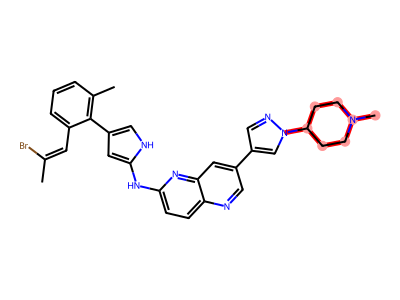} & \includegraphics[height=1.6cm]{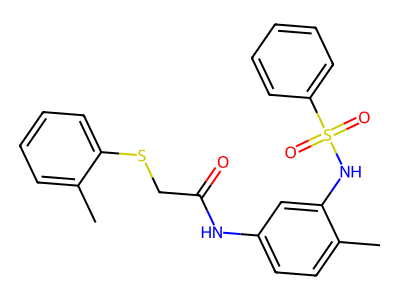} & \includegraphics[height=1.6cm]{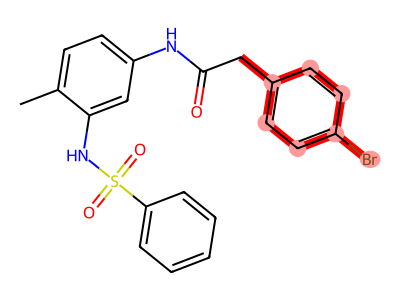} & \includegraphics[height=1.6cm]{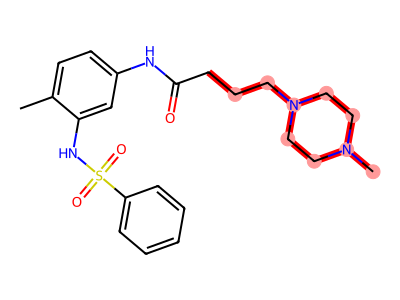} \\[-2pt]
{\scriptsize 0.16} & {\scriptsize 0.06 (\textcolor{red}{-0.10}) Sim 0.61} & {\scriptsize \textbf{0.43} (\textcolor{green!50!black}{+0.27}) Sim 0.64} & {\scriptsize 0.02} & {\scriptsize 0.00 (\textcolor{red}{-0.02}) Sim 0.61} & {\scriptsize \textbf{0.24} (\textcolor{green!50!black}{+0.22}) Sim 0.56} \\[6pt]
\includegraphics[height=1.6cm]{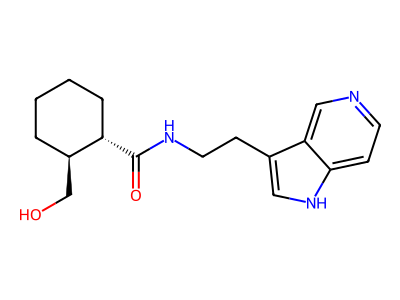} & \includegraphics[height=1.6cm]{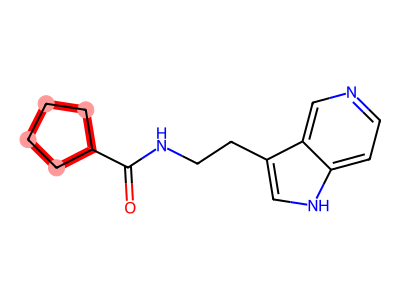} & \includegraphics[height=1.6cm]{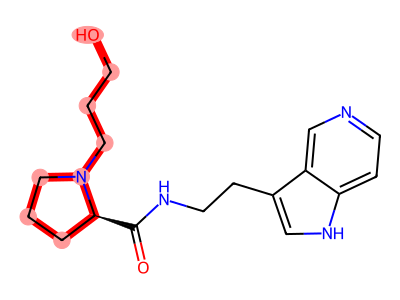} & \includegraphics[height=1.6cm]{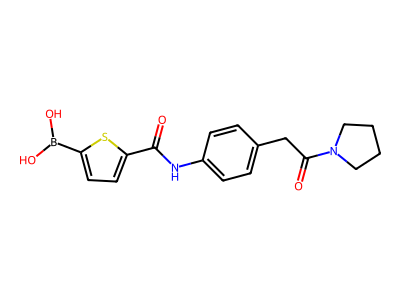} & \includegraphics[height=1.6cm]{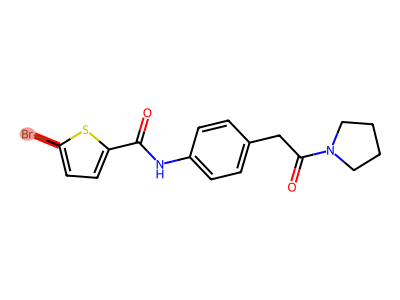} & \includegraphics[height=1.6cm]{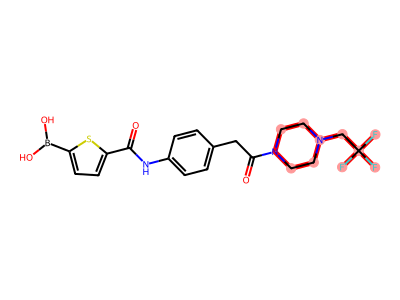} \\[-2pt]
{\scriptsize 0.03} & {\scriptsize 0.03 (\textcolor{red}{+0.00}) Sim 0.74} & {\scriptsize \textbf{0.22} (\textcolor{green!50!black}{+0.19}) Sim 0.62} & {\scriptsize 0.03} & {\scriptsize 0.01 (\textcolor{red}{-0.02}) Sim 0.71} & {\scriptsize \textbf{0.16} (\textcolor{green!50!black}{+0.13}) Sim 0.74} \\[6pt]
\includegraphics[height=1.6cm]{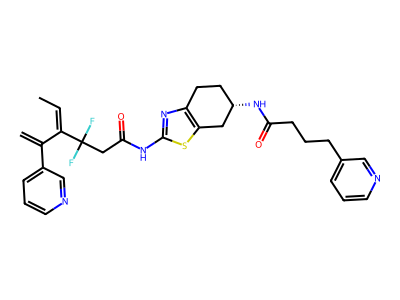} & \includegraphics[height=1.6cm]{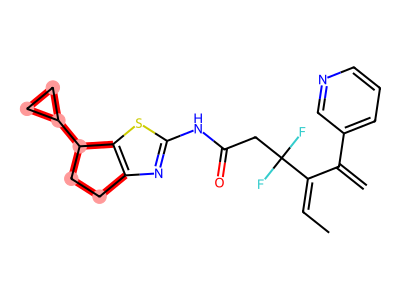} & \includegraphics[height=1.6cm]{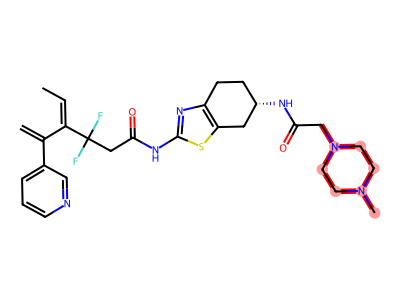} & \includegraphics[height=1.6cm]{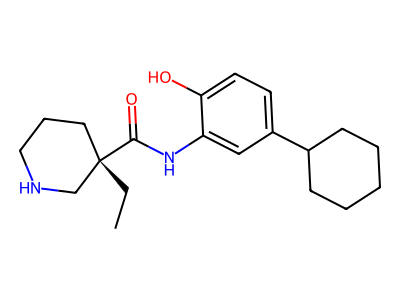} & \includegraphics[height=1.6cm]{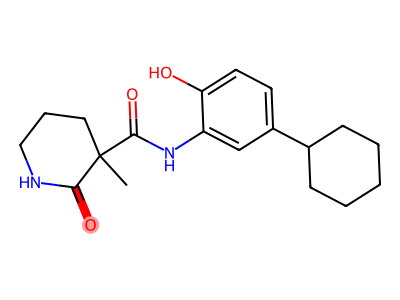} & \includegraphics[height=1.6cm]{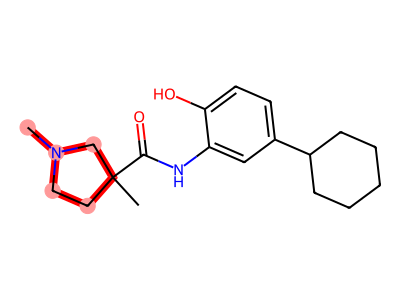} \\
\end{tabular}
\caption{\method steering examples: DRD2$\uparrow$ (GeLLM3O-LLaMA3).}
\label{fig:l3_drd2}
\end{figure*}

\begin{figure*}[ht]
\centering
\setlength{\tabcolsep}{1pt}
\begin{tabular}{ccc@{\hspace{8pt}}ccc}
\textbf{Source} & \textbf{SFT} & \textbf{+\method} & \textbf{Source} & \textbf{SFT} & \textbf{+\method} \\[2pt]
\includegraphics[height=1.6cm]{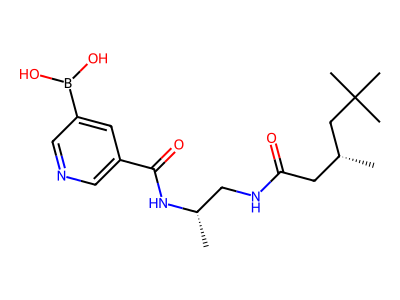} & \includegraphics[height=1.6cm]{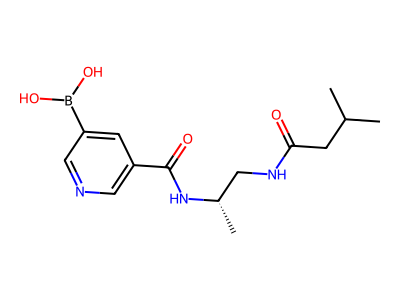} & \includegraphics[height=1.6cm]{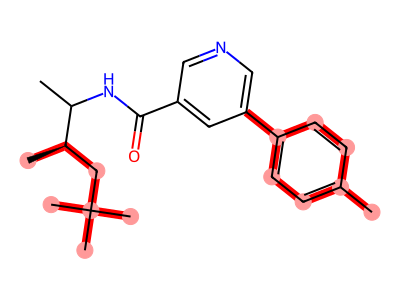} & \includegraphics[height=1.6cm]{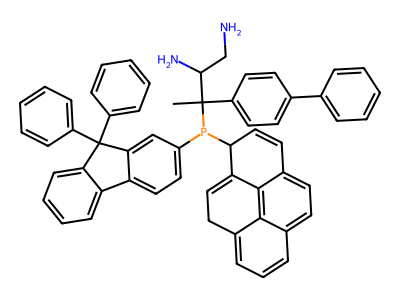} & \includegraphics[height=1.6cm]{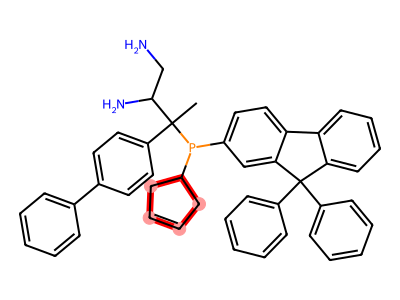} & \includegraphics[height=1.6cm]{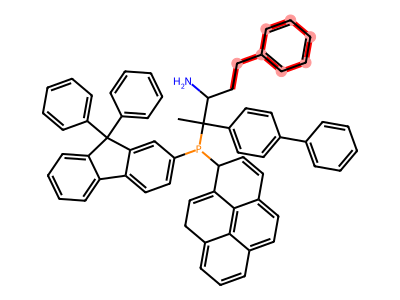} \\[-2pt]
{\scriptsize 0.46} & {\scriptsize -0.96 (\textcolor{red}{-1.42}) Sim 0.76} & {\scriptsize \textbf{5.25} (\textcolor{green!50!black}{+4.79}) Sim 0.40} & {\scriptsize 12.21} & {\scriptsize 9.74 (\textcolor{red}{-2.47}) Sim 0.59} & {\scriptsize \textbf{14.89} (\textcolor{green!50!black}{+2.67}) Sim 0.86} \\[6pt]
\includegraphics[height=1.6cm]{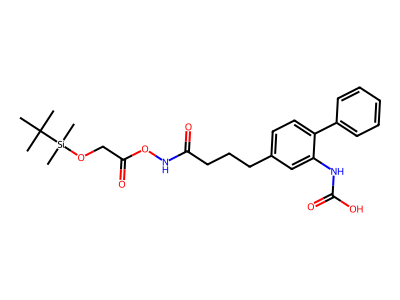} & \includegraphics[height=1.6cm]{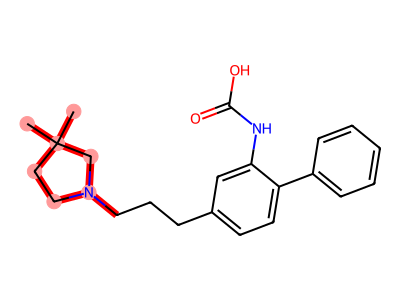} & \includegraphics[height=1.6cm]{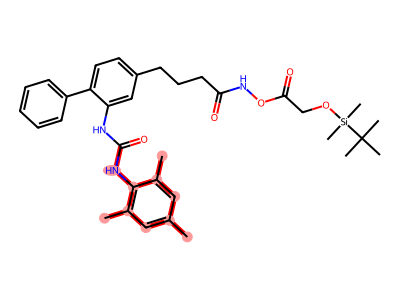} & \includegraphics[height=1.6cm]{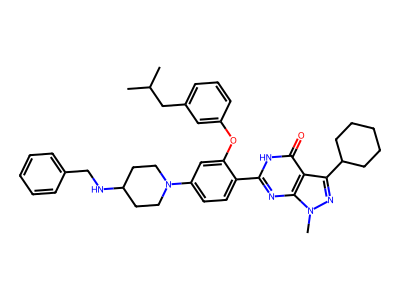} & \includegraphics[height=1.6cm]{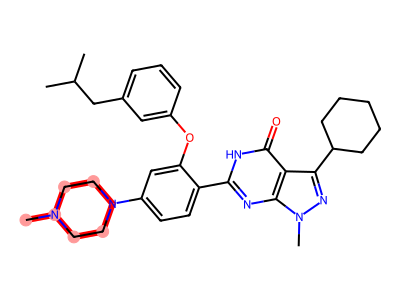} & \includegraphics[height=1.6cm]{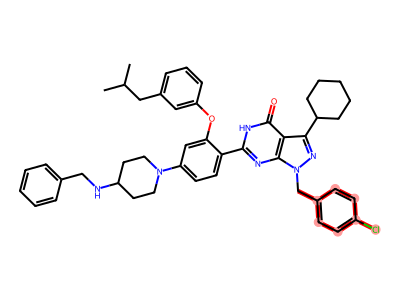} \\[-2pt]
{\scriptsize 5.36} & {\scriptsize 5.11 (\textcolor{red}{-0.25}) Sim 0.44} & {\scriptsize \textbf{7.84} (\textcolor{green!50!black}{+2.48}) Sim 0.81} & {\scriptsize 8.12} & {\scriptsize 6.11 (\textcolor{red}{-2.01}) Sim 0.82} & {\scriptsize \textbf{10.29} (\textcolor{green!50!black}{+2.16}) Sim 0.82} \\[6pt]
\includegraphics[height=1.6cm]{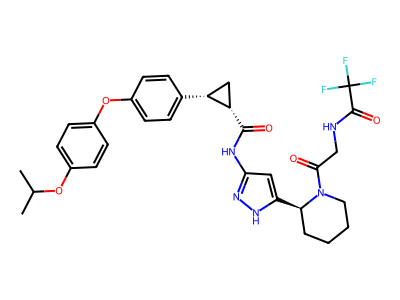} & \includegraphics[height=1.6cm]{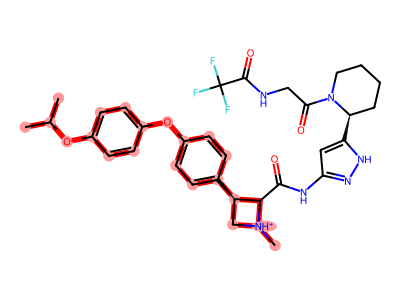} & \includegraphics[height=1.6cm]{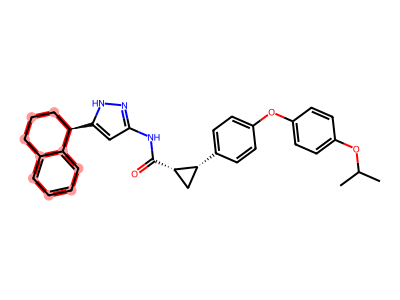} & \includegraphics[height=1.6cm]{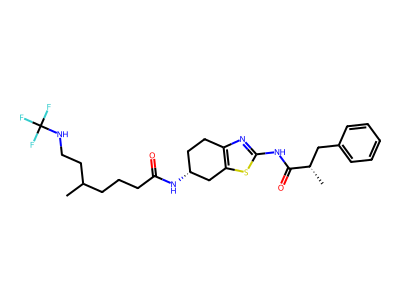} & \includegraphics[height=1.6cm]{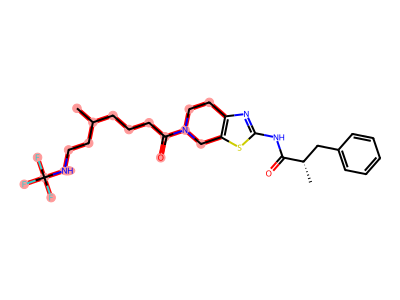} & \includegraphics[height=1.6cm]{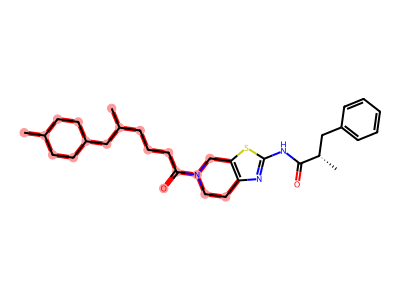} \\
\end{tabular}
\caption{\method steering examples: logP$\uparrow$ (GeLLM3O-LLaMA3).}
\label{fig:l3_logp}
\end{figure*}

\begin{figure*}[ht]
\centering
\setlength{\tabcolsep}{1pt}
\begin{tabular}{ccc@{\hspace{8pt}}ccc}
\textbf{Source} & \textbf{SFT} & \textbf{+\method} & \textbf{Source} & \textbf{SFT} & \textbf{+\method} \\[2pt]
\includegraphics[height=1.6cm]{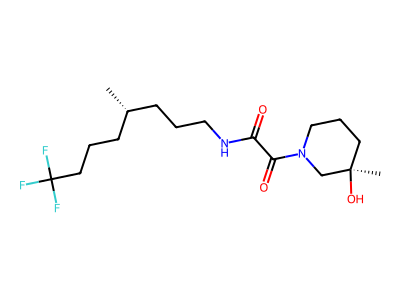} & \includegraphics[height=1.6cm]{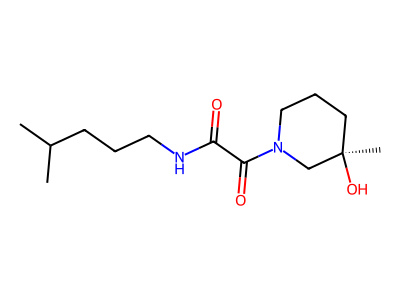} & \includegraphics[height=1.6cm]{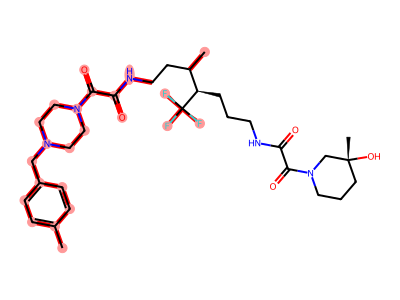} & \includegraphics[height=1.6cm]{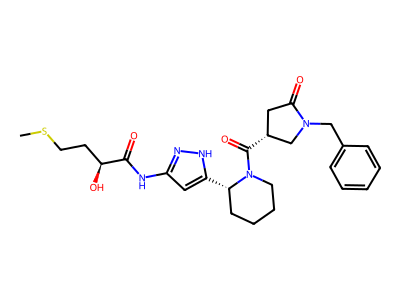} & \includegraphics[height=1.6cm]{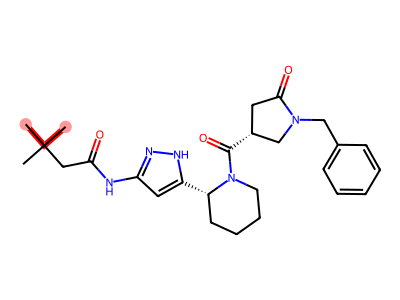} & \includegraphics[height=1.6cm]{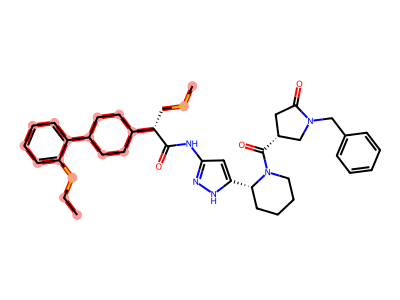} \\[-2pt]
{\scriptsize 366} & {\scriptsize 270 (\textcolor{red}{-96}) Sim 0.73} & {\scriptsize \textbf{626} (\textcolor{green!50!black}{+259}) Sim 0.54} & {\scriptsize 500} & {\scriptsize 466 (\textcolor{red}{-34}) Sim 0.71} & {\scriptsize \textbf{688} (\textcolor{green!50!black}{+188}) Sim 0.69} \\[6pt]
\includegraphics[height=1.6cm]{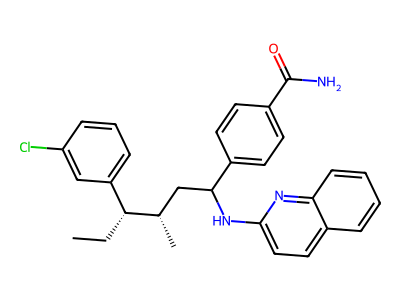} & \includegraphics[height=1.6cm]{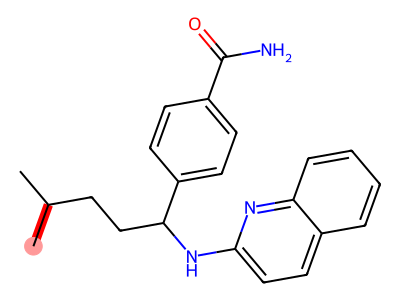} & \includegraphics[height=1.6cm]{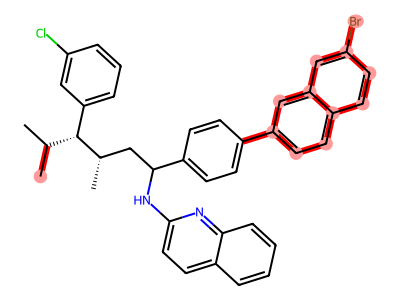} & \includegraphics[height=1.6cm]{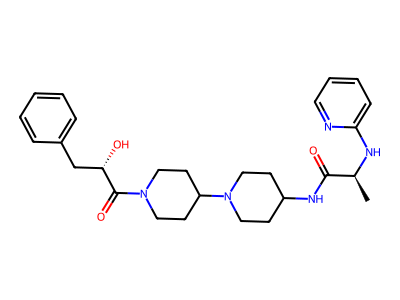} & \includegraphics[height=1.6cm]{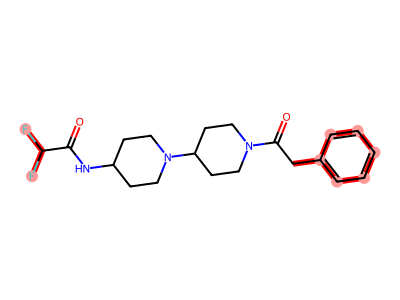} & \includegraphics[height=1.6cm]{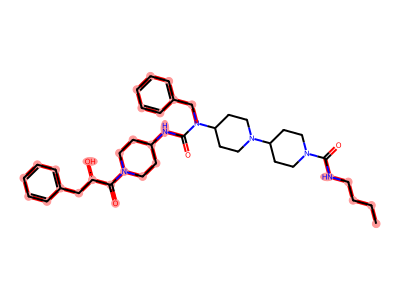} \\[-2pt]
{\scriptsize 472} & {\scriptsize 347 (\textcolor{red}{-125}) Sim 0.62} & {\scriptsize \textbf{648} (\textcolor{green!50!black}{+176}) Sim 0.53} & {\scriptsize 480} & {\scriptsize 379 (\textcolor{red}{-100}) Sim 0.48} & {\scriptsize \textbf{647} (\textcolor{green!50!black}{+167}) Sim 0.52} \\[6pt]
\includegraphics[height=1.6cm]{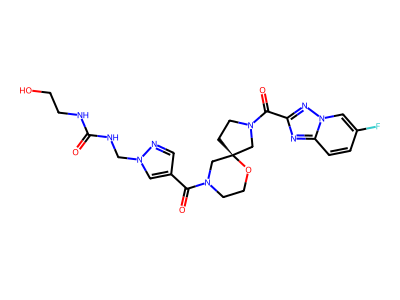} & \includegraphics[height=1.6cm]{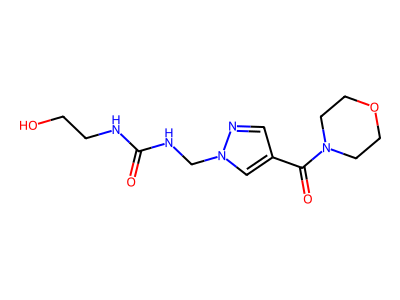} & \includegraphics[height=1.6cm]{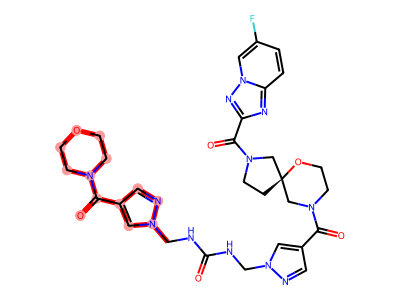} & \includegraphics[height=1.6cm]{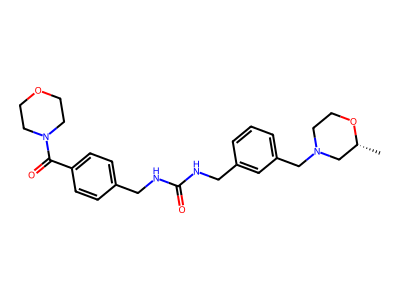} & \includegraphics[height=1.6cm]{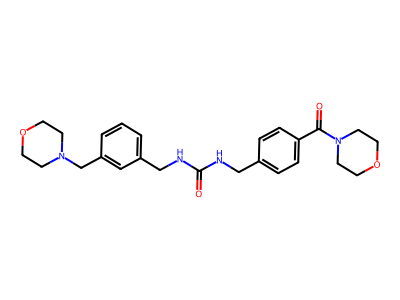} & \includegraphics[height=1.6cm]{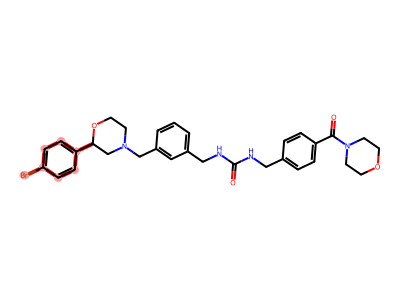} \\[-2pt]
{\scriptsize 516} & {\scriptsize 297 (\textcolor{red}{-218}) Sim 0.53} & {\scriptsize \textbf{665} (\textcolor{green!50!black}{+149}) Sim 0.86} & {\scriptsize 467} & {\scriptsize 453 (\textcolor{red}{-14}) Sim 0.80} & {\scriptsize \textbf{608} (\textcolor{green!50!black}{+141}) Sim 0.74} \\
\end{tabular}
\caption{\method steering examples: MW$\uparrow$ (GeLLM3O-LLaMA3).}
\label{fig:l3_mw}
\end{figure*}

\begin{figure*}[ht]
\centering
\setlength{\tabcolsep}{1pt}
\begin{tabular}{ccc@{\hspace{8pt}}ccc}
\textbf{Source} & \textbf{SFT} & \textbf{+\method} & \textbf{Source} & \textbf{SFT} & \textbf{+\method} \\[2pt]
\includegraphics[height=1.6cm]{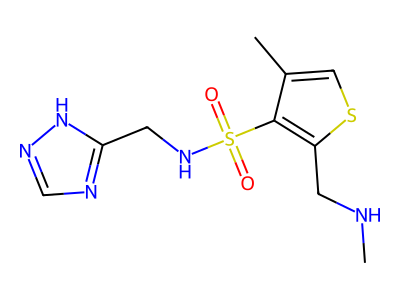} & \includegraphics[height=1.6cm]{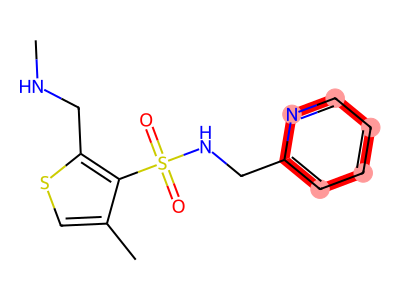} & \includegraphics[height=1.6cm]{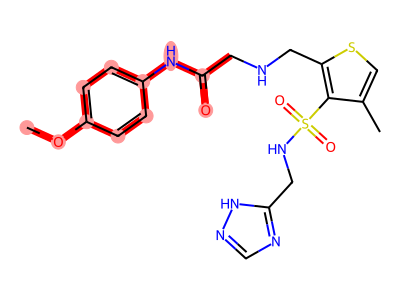} & \includegraphics[height=1.6cm]{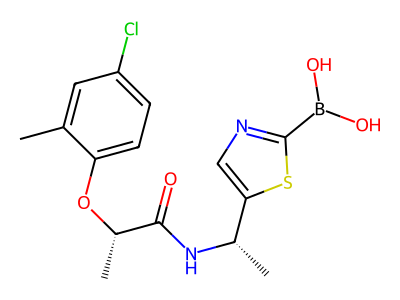} & \includegraphics[height=1.6cm]{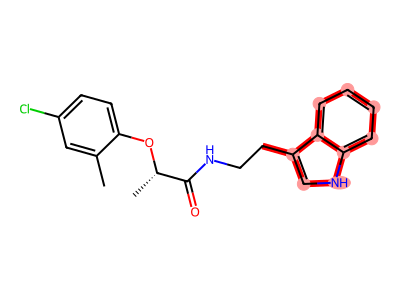} & \includegraphics[height=1.6cm]{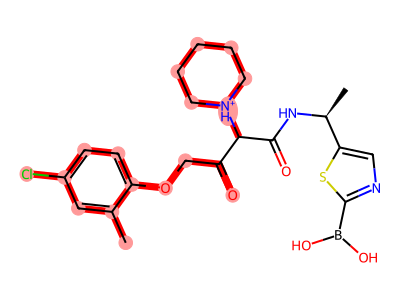} \\[-2pt]
{\scriptsize 6} & {\scriptsize 6 (\textcolor{red}{+0}) Sim 0.58} & {\scriptsize \textbf{10} (\textcolor{green!50!black}{+4}) Sim 0.61} & {\scriptsize 6} & {\scriptsize 6 (\textcolor{red}{+0}) Sim 0.42} & {\scriptsize \textbf{9} (\textcolor{green!50!black}{+3}) Sim 0.64} \\[6pt]
\includegraphics[height=1.6cm]{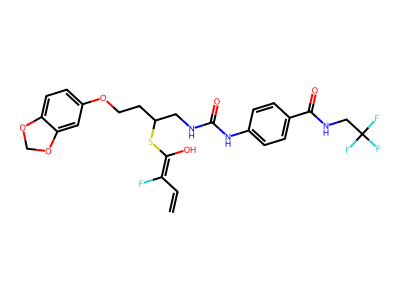} & \includegraphics[height=1.6cm]{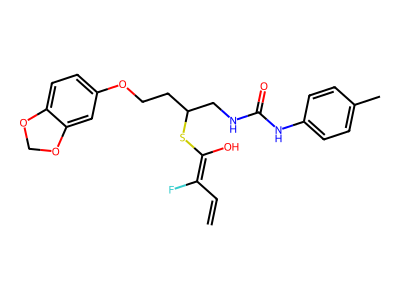} & \includegraphics[height=1.6cm]{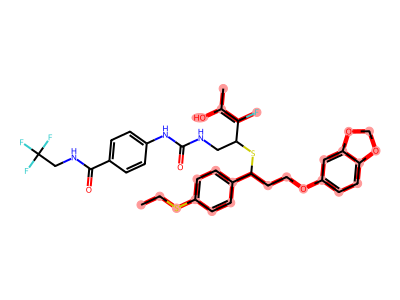} & \includegraphics[height=1.6cm]{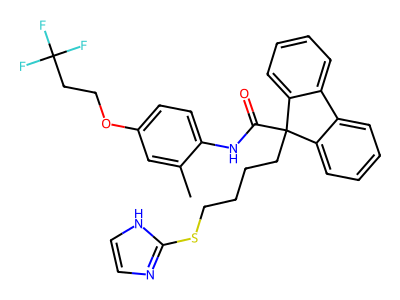} & \includegraphics[height=1.6cm]{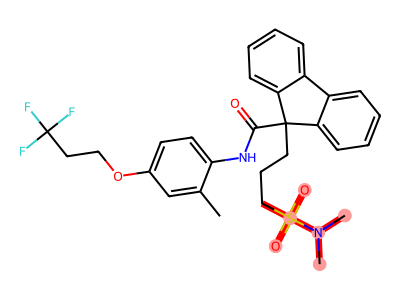} & \includegraphics[height=1.6cm]{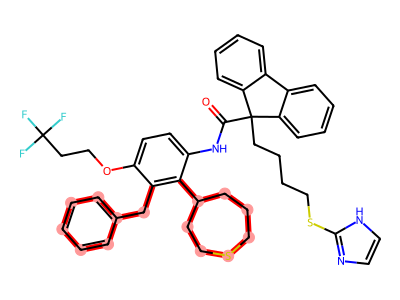} \\[-2pt]
{\scriptsize 12} & {\scriptsize 10 (\textcolor{red}{-2}) Sim 0.78} & {\scriptsize \textbf{15} (\textcolor{green!50!black}{+3}) Sim 0.58} & {\scriptsize 11} & {\scriptsize 10 (\textcolor{red}{-1}) Sim 0.65} & {\scriptsize \textbf{14} (\textcolor{green!50!black}{+3}) Sim 0.58} \\[6pt]
\includegraphics[height=1.6cm]{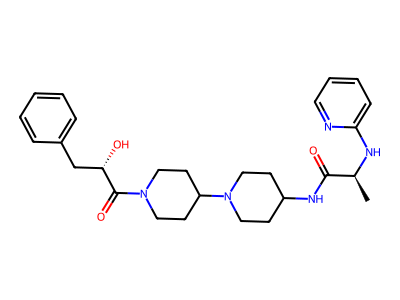} & \includegraphics[height=1.6cm]{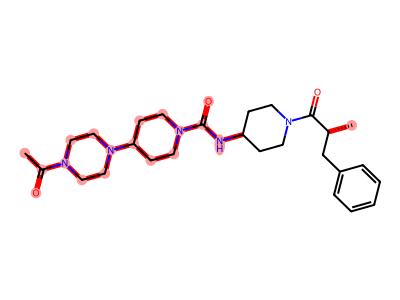} & \includegraphics[height=1.6cm]{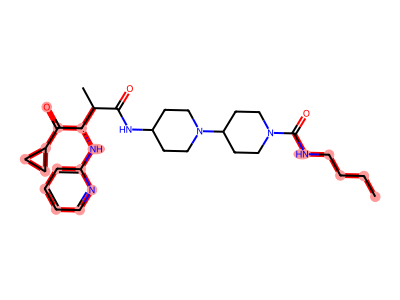} & \includegraphics[height=1.6cm]{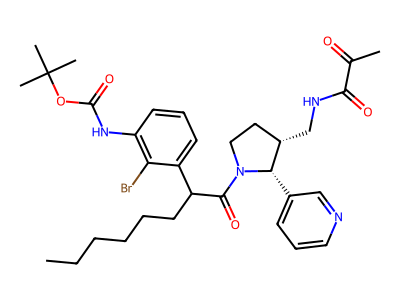} & \includegraphics[height=1.6cm]{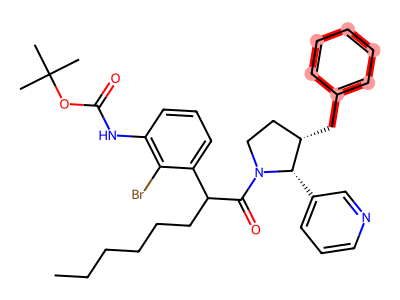} & \includegraphics[height=1.6cm]{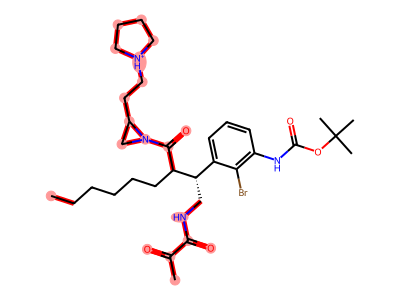} \\
\end{tabular}
\caption{\method steering examples: RotBond$\uparrow$ (GeLLM3O-LLaMA3).}
\label{fig:l3_rotbond}
\end{figure*}

\begin{figure*}[ht]
\centering
\setlength{\tabcolsep}{1pt}
\begin{tabular}{ccc@{\hspace{8pt}}ccc}
\textbf{Source} & \textbf{SFT} & \textbf{+\method} & \textbf{Source} & \textbf{SFT} & \textbf{+\method} \\[2pt]
\includegraphics[height=1.6cm]{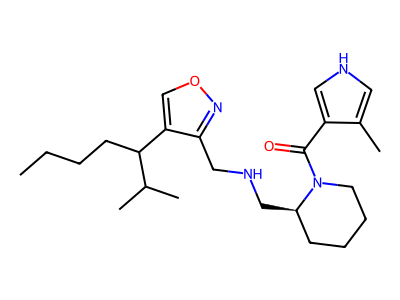} & \includegraphics[height=1.6cm]{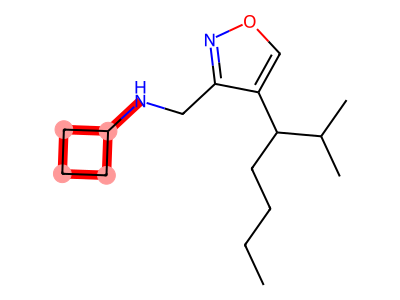} & \includegraphics[height=1.6cm]{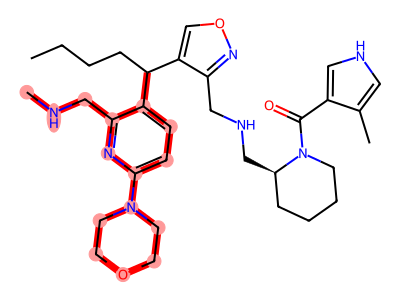} & \includegraphics[height=1.6cm]{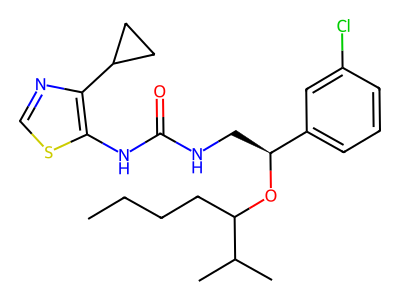} & \includegraphics[height=1.6cm]{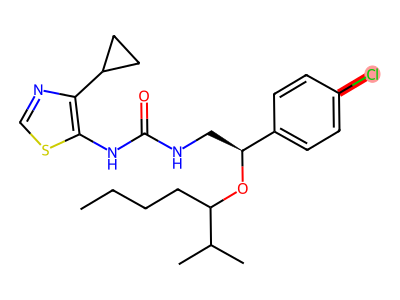} & \includegraphics[height=1.6cm]{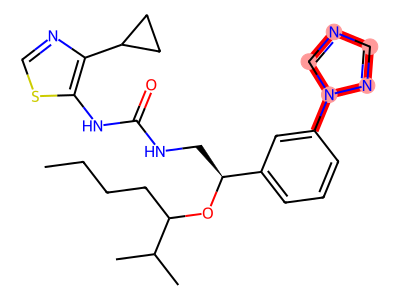} \\[-2pt]
{\scriptsize 4} & {\scriptsize 3 (\textcolor{red}{-1}) Sim 0.49} & {\scriptsize \textbf{8} (\textcolor{green!50!black}{+4}) Sim 0.65} & {\scriptsize 4} & {\scriptsize 4 (\textcolor{red}{+0}) Sim 0.82} & {\scriptsize \textbf{7} (\textcolor{green!50!black}{+3}) Sim 0.73} \\[6pt]
\includegraphics[height=1.6cm]{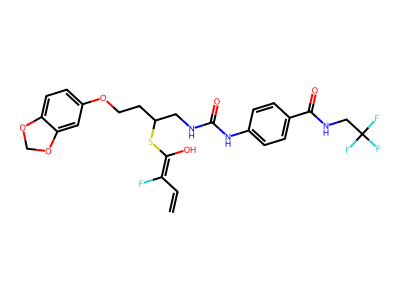} & \includegraphics[height=1.6cm]{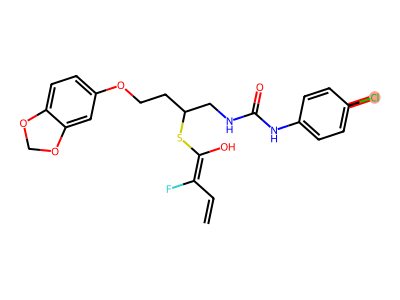} & \includegraphics[height=1.6cm]{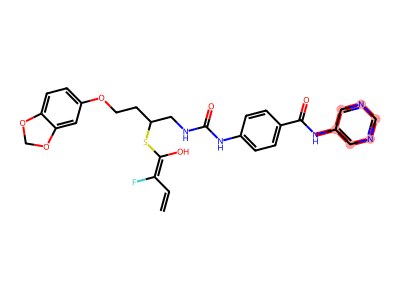} & \includegraphics[height=1.6cm]{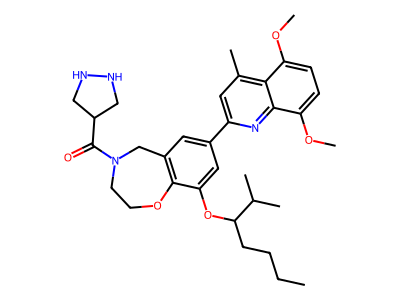} & \includegraphics[height=1.6cm]{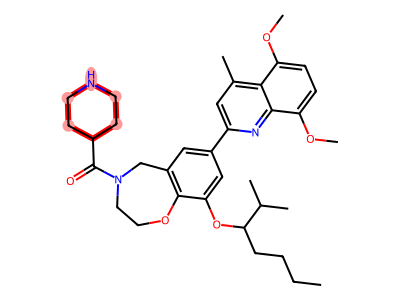} & \includegraphics[height=1.6cm]{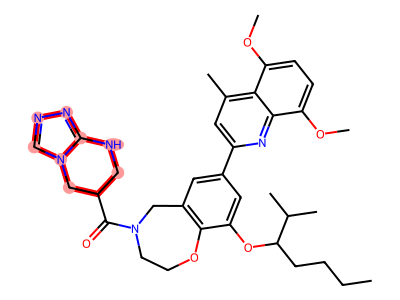} \\[-2pt]
{\scriptsize 7} & {\scriptsize 6 (\textcolor{red}{-1}) Sim 0.78} & {\scriptsize \textbf{9} (\textcolor{green!50!black}{+2}) Sim 0.78} & {\scriptsize 8} & {\scriptsize 7 (\textcolor{red}{-1}) Sim 0.87} & {\scriptsize \textbf{10} (\textcolor{green!50!black}{+2}) Sim 0.79} \\[6pt]
\includegraphics[height=1.6cm]{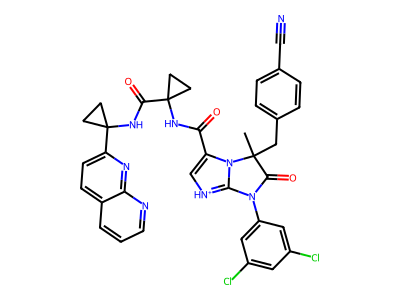} & \includegraphics[height=1.6cm]{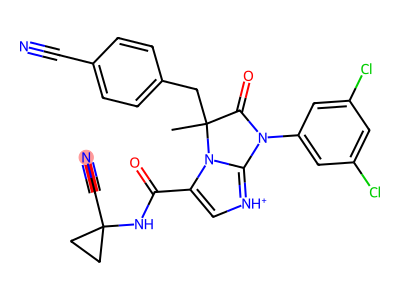} & \includegraphics[height=1.6cm]{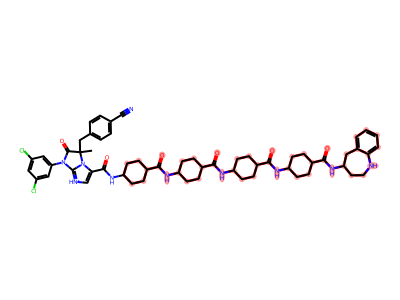} & \includegraphics[height=1.6cm]{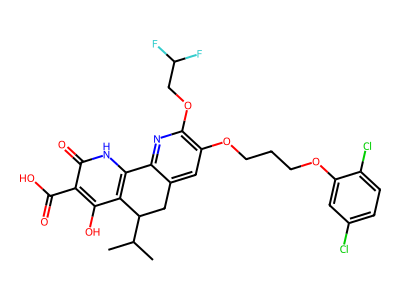} & \includegraphics[height=1.6cm]{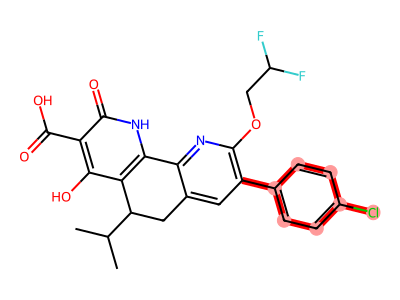} & \includegraphics[height=1.6cm]{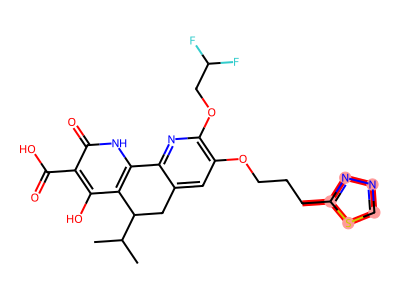} \\
\end{tabular}
\caption{\method steering examples: HBA$\uparrow$ (GeLLM3O-LLaMA3).}
\label{fig:l3_acceptor}
\end{figure*}

\begin{figure*}[ht]
\centering
\setlength{\tabcolsep}{1pt}
\begin{tabular}{ccc@{\hspace{8pt}}ccc}
\textbf{Source} & \textbf{SFT} & \textbf{+\method} & \textbf{Source} & \textbf{SFT} & \textbf{+\method} \\[2pt]
\includegraphics[height=1.6cm]{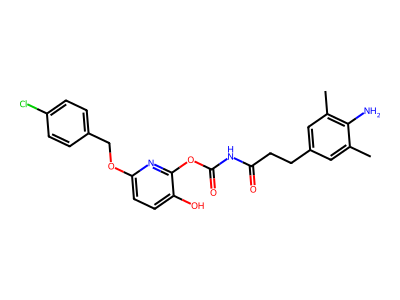} & \includegraphics[height=1.6cm]{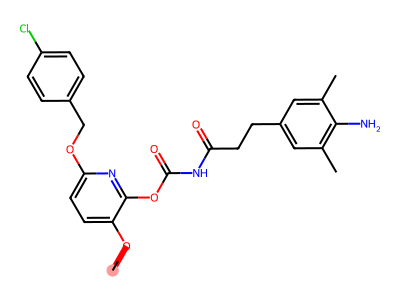} & \includegraphics[height=1.6cm]{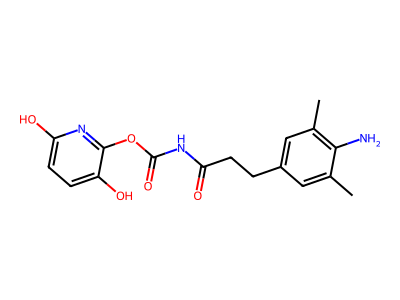} & \includegraphics[height=1.6cm]{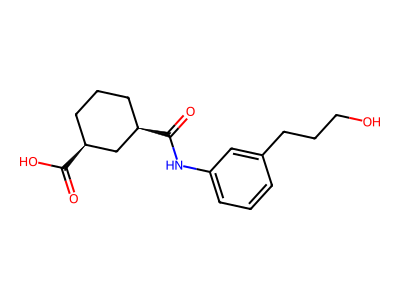} & \includegraphics[height=1.6cm]{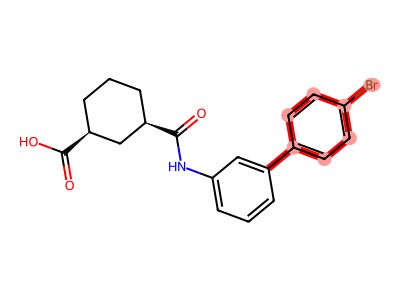} & \includegraphics[height=1.6cm]{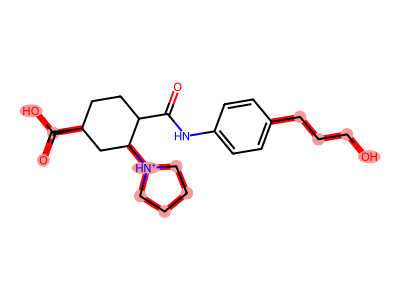} \\[-2pt]
{\scriptsize 3} & {\scriptsize 2 (\textcolor{red}{-1}) Sim 0.81} & {\scriptsize \textbf{4} (\textcolor{green!50!black}{+1}) Sim 0.68} & {\scriptsize 3} & {\scriptsize 2 (\textcolor{red}{-1}) Sim 0.58} & {\scriptsize \textbf{4} (\textcolor{green!50!black}{+1}) Sim 0.50} \\[6pt]
\includegraphics[height=1.6cm]{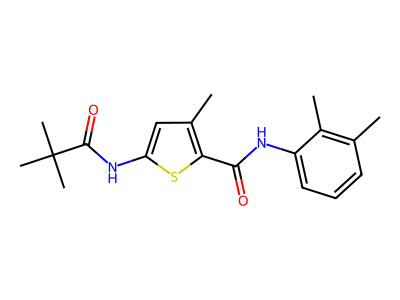} & \includegraphics[height=1.6cm]{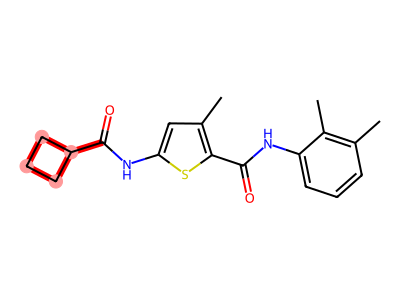} & \includegraphics[height=1.6cm]{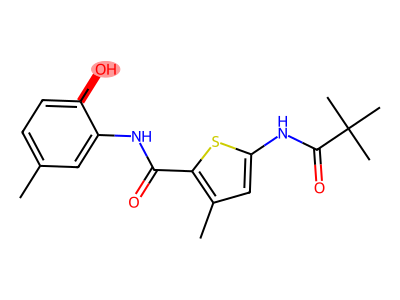} & \includegraphics[height=1.6cm]{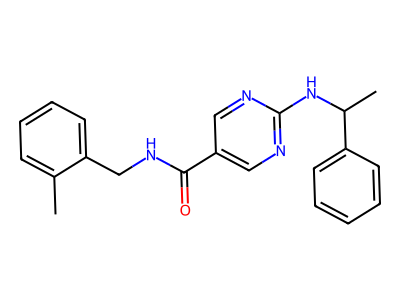} & \includegraphics[height=1.6cm]{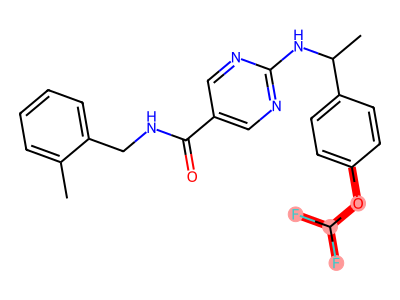} & \includegraphics[height=1.6cm]{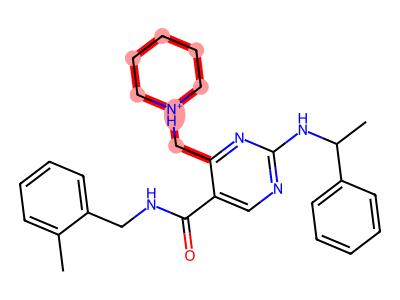} \\[-2pt]
{\scriptsize 2} & {\scriptsize 2 (\textcolor{red}{+0}) Sim 0.65} & {\scriptsize \textbf{3} (\textcolor{green!50!black}{+1}) Sim 0.65} & {\scriptsize 2} & {\scriptsize 2 (\textcolor{red}{+0}) Sim 0.76} & {\scriptsize \textbf{3} (\textcolor{green!50!black}{+1}) Sim 0.60} \\[6pt]
\includegraphics[height=1.6cm]{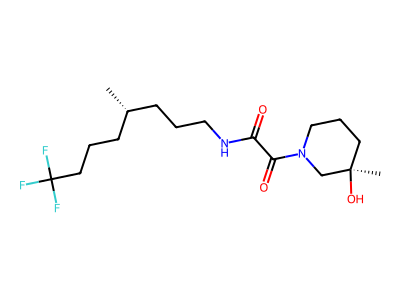} & \includegraphics[height=1.6cm]{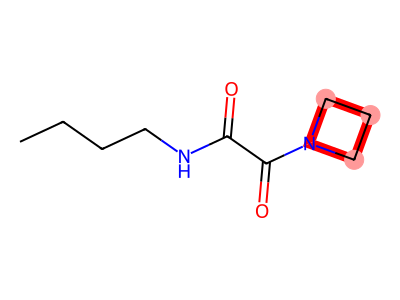} & \includegraphics[height=1.6cm]{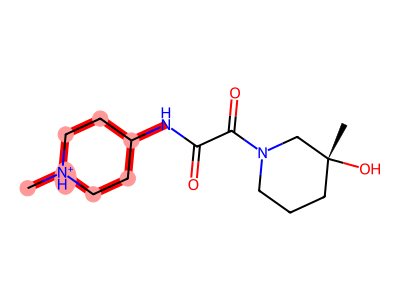} & \includegraphics[height=1.6cm]{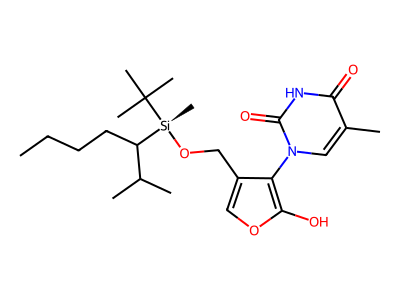} & \includegraphics[height=1.6cm]{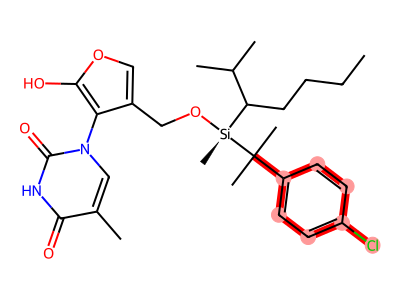} & \includegraphics[height=1.6cm]{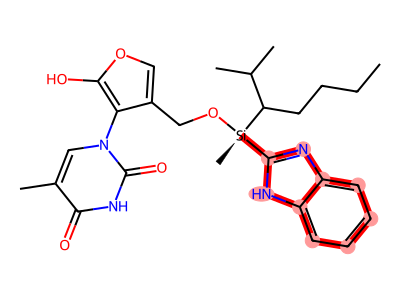} \\[-2pt]
{\scriptsize 2} & {\scriptsize 1 (\textcolor{red}{-1}) Sim 0.40} & {\scriptsize \textbf{3} (\textcolor{green!50!black}{+1}) Sim 0.42} & {\scriptsize 2} & {\scriptsize 2 (\textcolor{red}{+0}) Sim 0.79} & {\scriptsize \textbf{3} (\textcolor{green!50!black}{+1}) Sim 0.66} \\
\end{tabular}
\caption{\method steering examples: HBD$\uparrow$ (GeLLM3O-LLaMA3).}
\label{fig:l3_donor}
\end{figure*}

\subsubsection*{MolGen-Large}

\begin{figure*}[ht]
\centering
\setlength{\tabcolsep}{1pt}
\begin{tabular}{ccc@{\hspace{8pt}}ccc}
\textbf{Source} & \textbf{SFT} & \textbf{+\method} & \textbf{Source} & \textbf{SFT} & \textbf{+\method} \\[2pt]
\includegraphics[height=1.6cm]{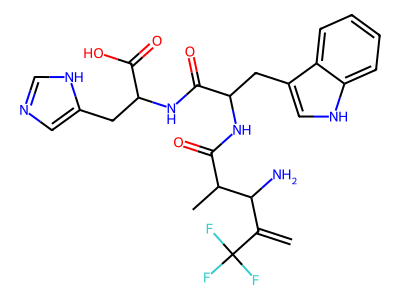} & \includegraphics[height=1.6cm]{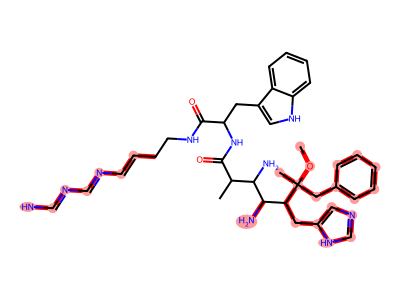} & \includegraphics[height=1.6cm]{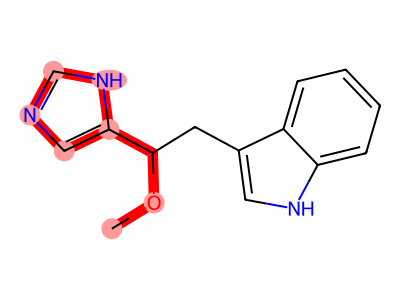} & \includegraphics[height=1.6cm]{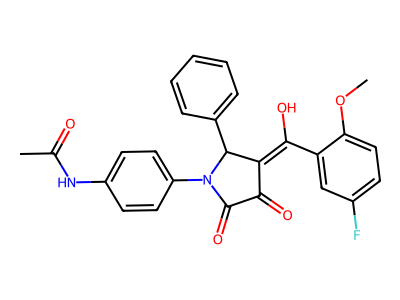} & \includegraphics[height=1.6cm]{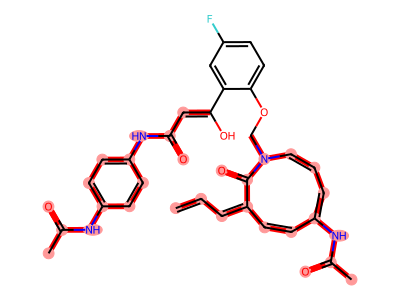} & \includegraphics[height=1.6cm]{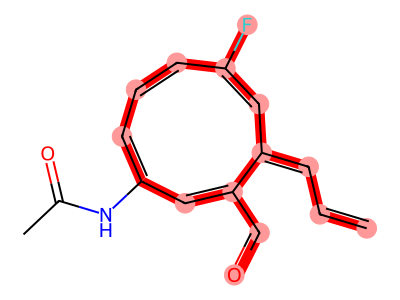} \\[-2pt]
{\scriptsize 0.21} & {\scriptsize 0.05 (\textcolor{red}{-0.16}) Sim 0.46} & {\scriptsize \textbf{0.74} (\textcolor{green!50!black}{+0.53}) Sim 0.37} & {\scriptsize 0.33} & {\scriptsize 0.25 (\textcolor{red}{-0.08}) Sim 0.33} & {\scriptsize \textbf{0.79} (\textcolor{green!50!black}{+0.46}) Sim 0.15} \\[6pt]
\includegraphics[height=1.6cm]{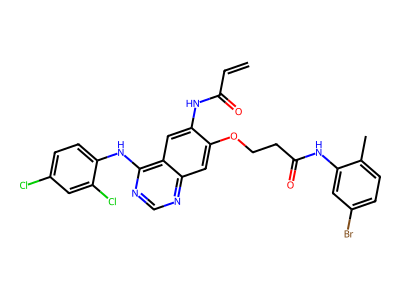} & \includegraphics[height=1.6cm]{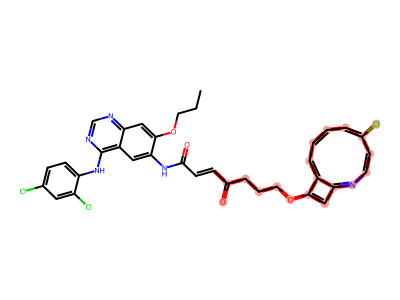} & \includegraphics[height=1.6cm]{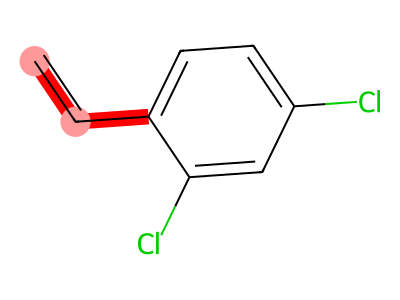} & \includegraphics[height=1.6cm]{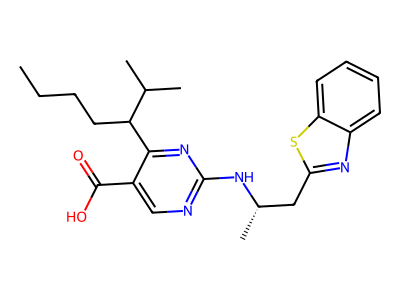} & \includegraphics[height=1.6cm]{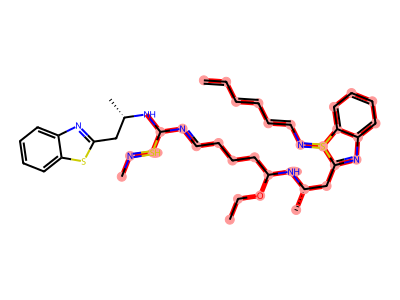} & \includegraphics[height=1.6cm]{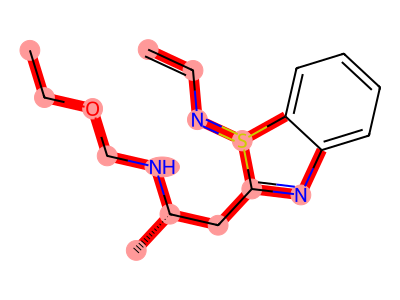} \\[-2pt]
{\scriptsize 0.17} & {\scriptsize 0.11 (\textcolor{red}{-0.07}) Sim 0.51} & {\scriptsize \textbf{0.61} (\textcolor{green!50!black}{+0.44}) Sim 0.21} & {\scriptsize 0.43} & {\scriptsize 0.04 (\textcolor{red}{-0.39}) Sim 0.28} & {\scriptsize \textbf{0.62} (\textcolor{green!50!black}{+0.19}) Sim 0.16} \\[6pt]
\includegraphics[height=1.6cm]{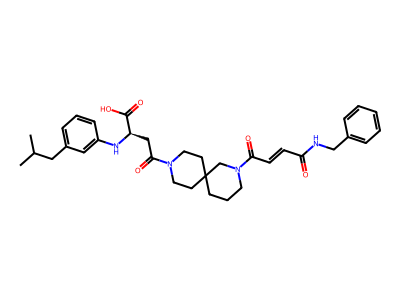} & \includegraphics[height=1.6cm]{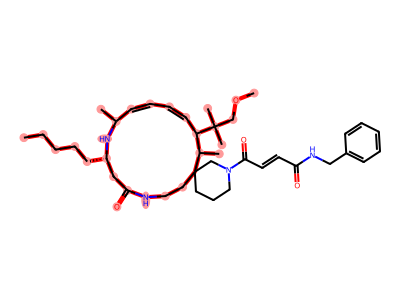} & \includegraphics[height=1.6cm]{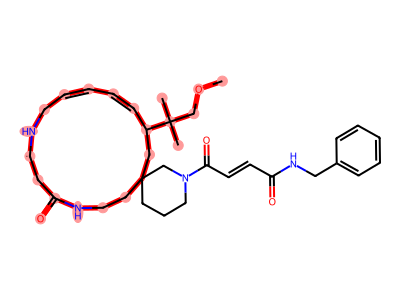} & \includegraphics[height=1.6cm]{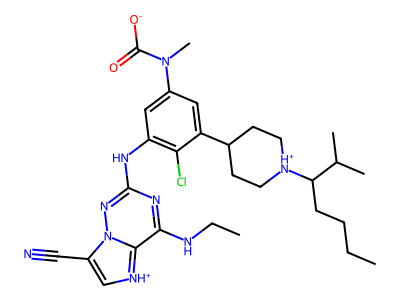} & \includegraphics[height=1.6cm]{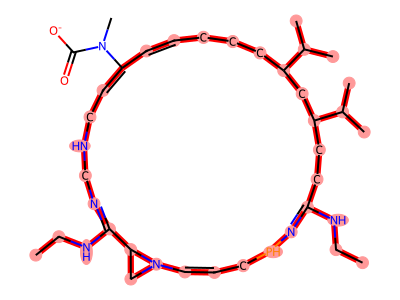} & \includegraphics[height=1.6cm]{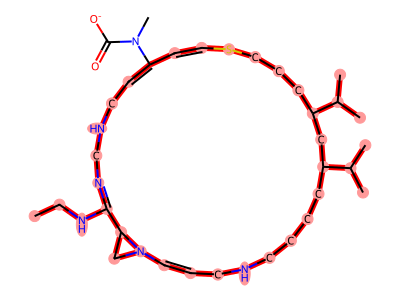} \\
\end{tabular}
\caption{\method steering examples: QED$\uparrow$ (MolGen-Large).}
\label{fig:mg_qed}
\end{figure*}

\begin{figure*}[ht]
\centering
\setlength{\tabcolsep}{1pt}
\begin{tabular}{ccc@{\hspace{8pt}}ccc}
\textbf{Source} & \textbf{SFT} & \textbf{+\method} & \textbf{Source} & \textbf{SFT} & \textbf{+\method} \\[2pt]
\includegraphics[height=1.6cm]{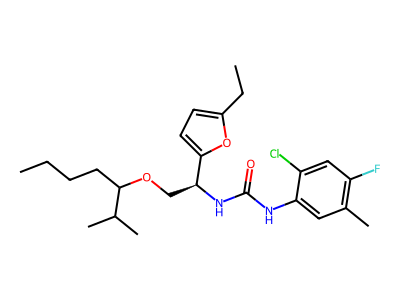} & \includegraphics[height=1.6cm]{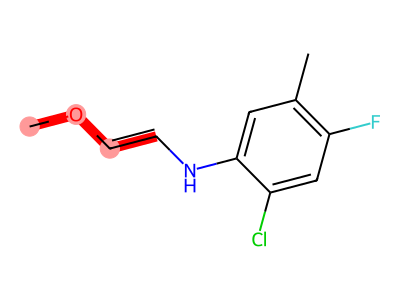} & \includegraphics[height=1.6cm]{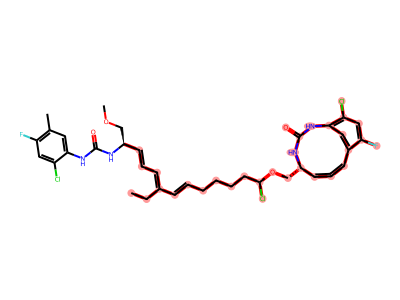} & \includegraphics[height=1.6cm]{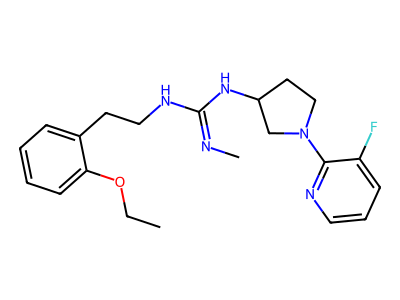} & \includegraphics[height=1.6cm]{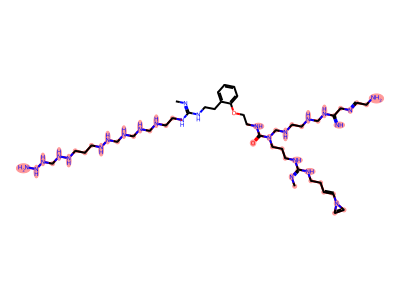} & \includegraphics[height=1.6cm]{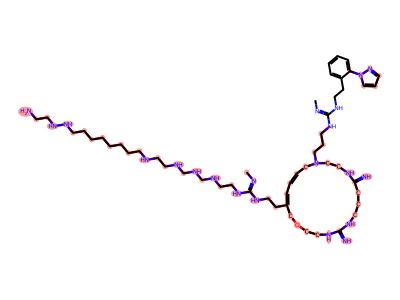} \\[-2pt]
{\scriptsize 0.08} & {\scriptsize 0.08 (\textcolor{red}{+0.00}) Sim 0.28} & {\scriptsize \textbf{0.23} (\textcolor{green!50!black}{+0.15}) Sim 0.32} & {\scriptsize 0.25} & {\scriptsize 0.13 (\textcolor{red}{-0.12}) Sim 0.25} & {\scriptsize \textbf{0.39} (\textcolor{green!50!black}{+0.14}) Sim 0.21} \\[6pt]
\includegraphics[height=1.6cm]{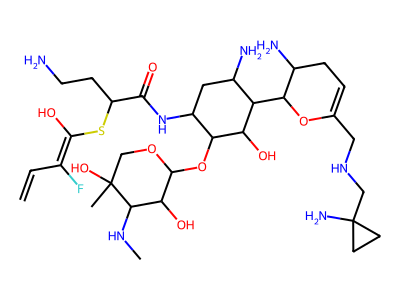} & \includegraphics[height=1.6cm]{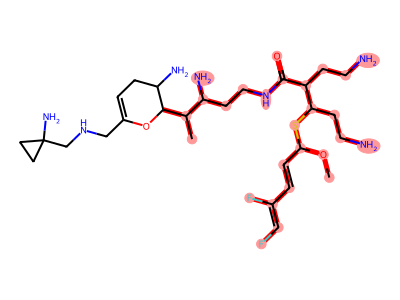} & \includegraphics[height=1.6cm]{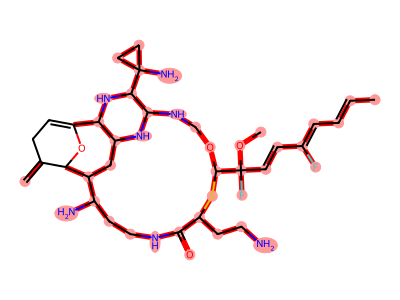} & \includegraphics[height=1.6cm]{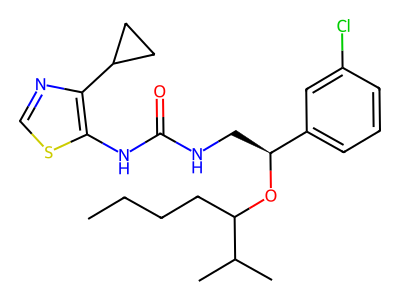} & \includegraphics[height=1.6cm]{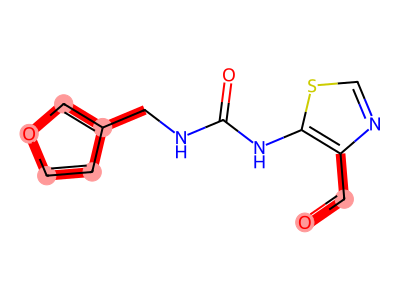} & \includegraphics[height=1.6cm]{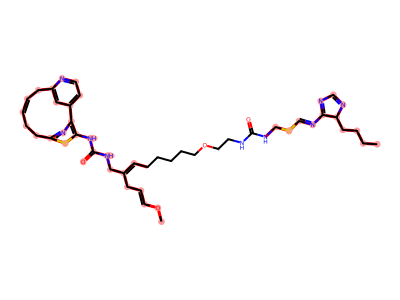} \\[-2pt]
{\scriptsize 0.15} & {\scriptsize 0.06 (\textcolor{red}{-0.09}) Sim 0.35} & {\scriptsize \textbf{0.27} (\textcolor{green!50!black}{+0.12}) Sim 0.17} & {\scriptsize 0.11} & {\scriptsize 0.01 (\textcolor{red}{-0.10}) Sim 0.24} & {\scriptsize \textbf{0.20} (\textcolor{green!50!black}{+0.09}) Sim 0.18} \\[6pt]
\includegraphics[height=1.6cm]{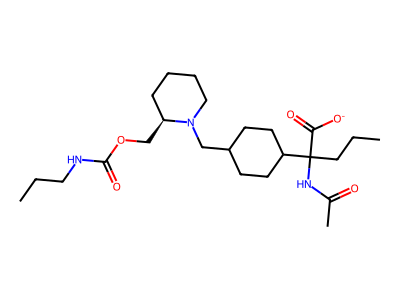} & \includegraphics[height=1.6cm]{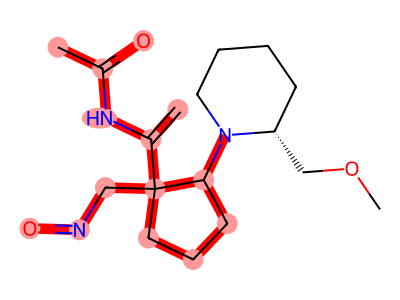} & \includegraphics[height=1.6cm]{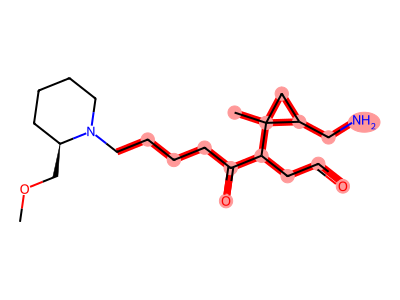} & \includegraphics[height=1.6cm]{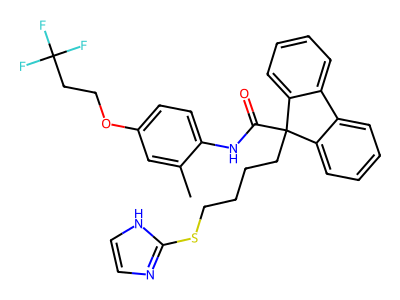} & \includegraphics[height=1.6cm]{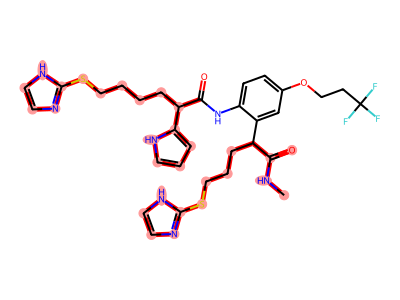} & \includegraphics[height=1.6cm]{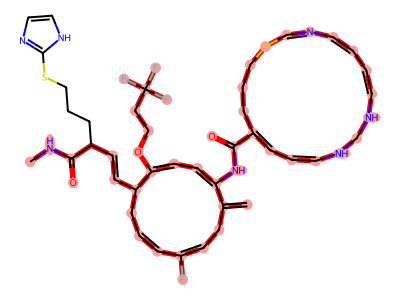} \\
\end{tabular}
\caption{\method steering examples: DRD2$\uparrow$ (MolGen-Large).}
\label{fig:mg_drd2}
\end{figure*}

\begin{figure*}[ht]
\centering
\setlength{\tabcolsep}{1pt}
\begin{tabular}{ccc@{\hspace{8pt}}ccc}
\textbf{Source} & \textbf{SFT} & \textbf{+\method} & \textbf{Source} & \textbf{SFT} & \textbf{+\method} \\[2pt]
\includegraphics[height=1.6cm]{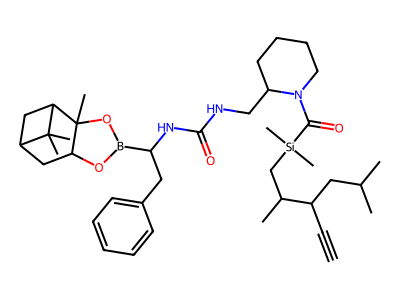} & \includegraphics[height=1.6cm]{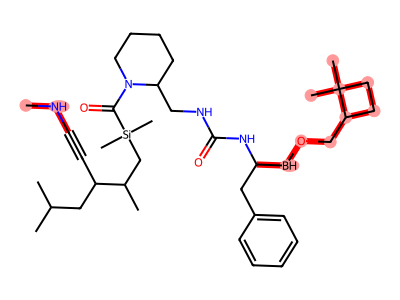} & \includegraphics[height=1.6cm]{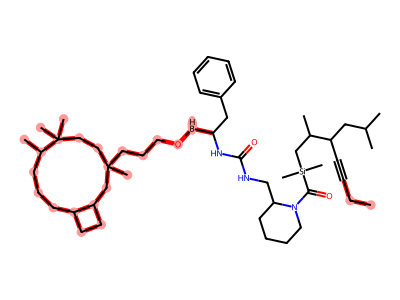} & \includegraphics[height=1.6cm]{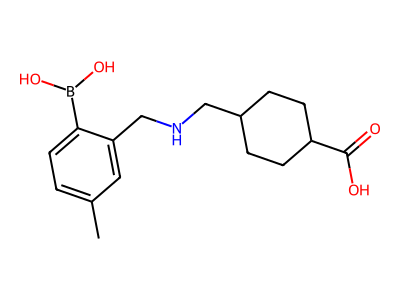} & \includegraphics[height=1.6cm]{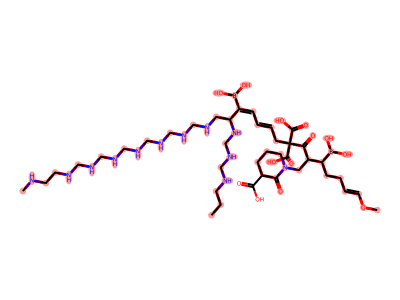} & \includegraphics[height=1.6cm]{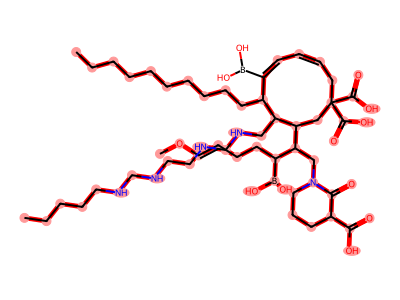} \\[-2pt]
{\scriptsize 7.36} & {\scriptsize 6.40 (\textcolor{red}{-0.96}) Sim 0.51} & {\scriptsize \textbf{12.03} (\textcolor{green!50!black}{+4.67}) Sim 0.45} & {\scriptsize 0.66} & {\scriptsize -5.13 (\textcolor{red}{-5.78}) Sim 0.16} & {\scriptsize \textbf{4.00} (\textcolor{green!50!black}{+3.34}) Sim 0.19} \\[6pt]
\includegraphics[height=1.6cm]{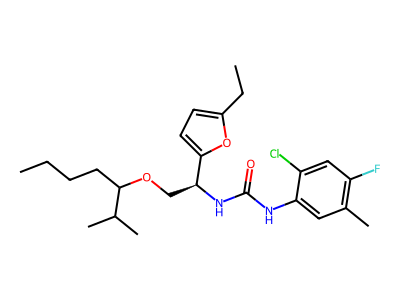} & \includegraphics[height=1.6cm]{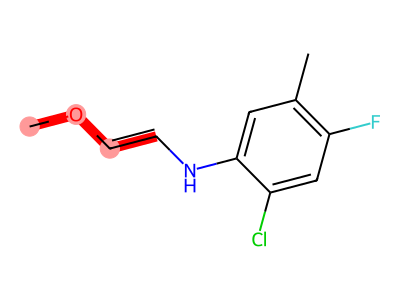} & \includegraphics[height=1.6cm]{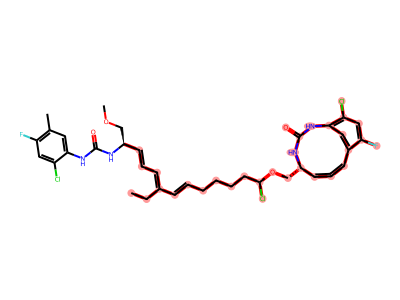} & \includegraphics[height=1.6cm]{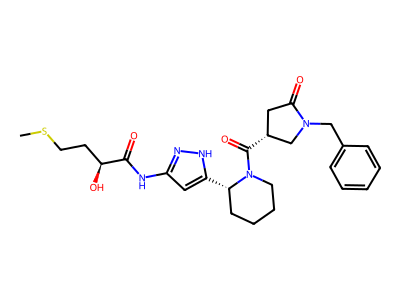} & \includegraphics[height=1.6cm]{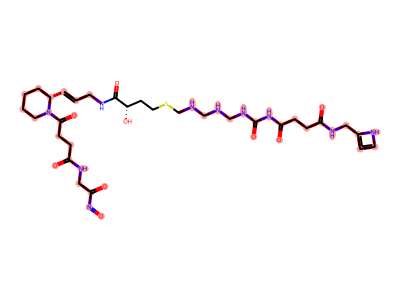} & \includegraphics[height=1.6cm]{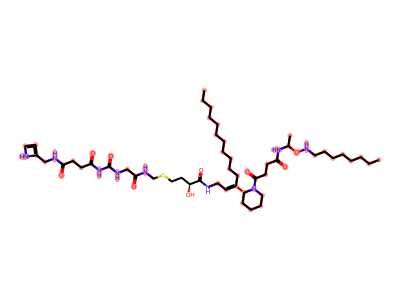} \\[-2pt]
{\scriptsize 7.04} & {\scriptsize 3.32 (\textcolor{red}{-3.72}) Sim 0.28} & {\scriptsize \textbf{9.68} (\textcolor{green!50!black}{+2.64}) Sim 0.32} & {\scriptsize 2.56} & {\scriptsize -2.28 (\textcolor{red}{-4.85}) Sim 0.21} & {\scriptsize \textbf{4.96} (\textcolor{green!50!black}{+2.40}) Sim 0.20} \\[6pt]
\includegraphics[height=1.6cm]{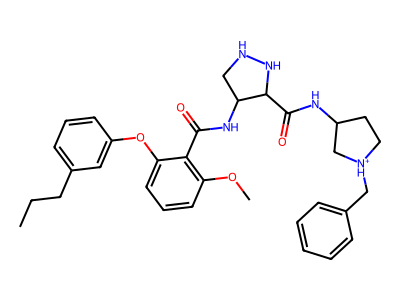} & \includegraphics[height=1.6cm]{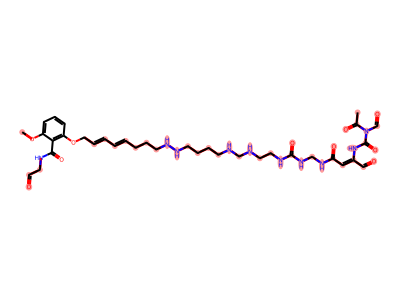} & \includegraphics[height=1.6cm]{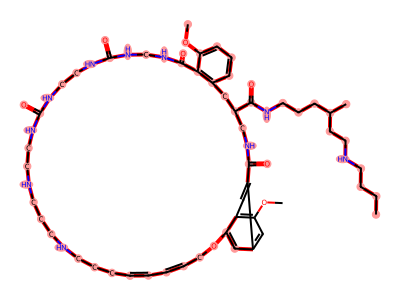} & \includegraphics[height=1.6cm]{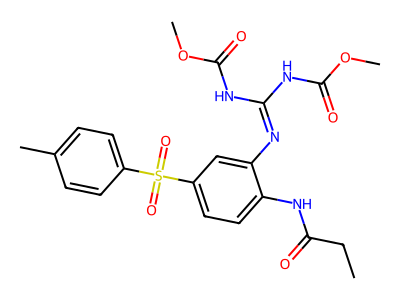} & \includegraphics[height=1.6cm]{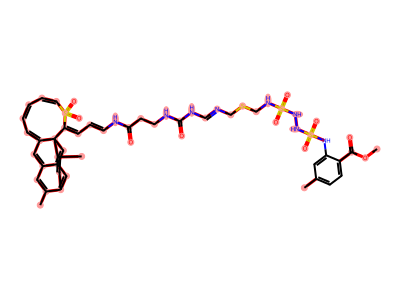} & \includegraphics[height=1.6cm]{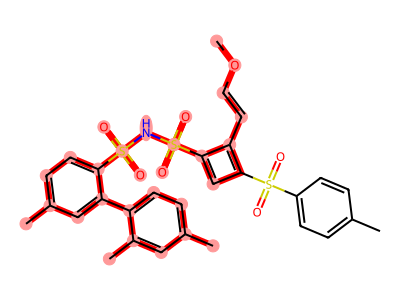} \\
\end{tabular}
\caption{\method steering examples: logP$\uparrow$ (MolGen-Large).}
\label{fig:mg_logp}
\end{figure*}

\begin{figure*}[ht]
\centering
\setlength{\tabcolsep}{1pt}
\begin{tabular}{ccc@{\hspace{8pt}}ccc}
\textbf{Source} & \textbf{SFT} & \textbf{+\method} & \textbf{Source} & \textbf{SFT} & \textbf{+\method} \\[2pt]
\includegraphics[height=1.6cm]{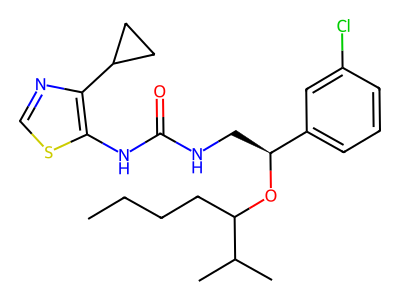} & \includegraphics[height=1.6cm]{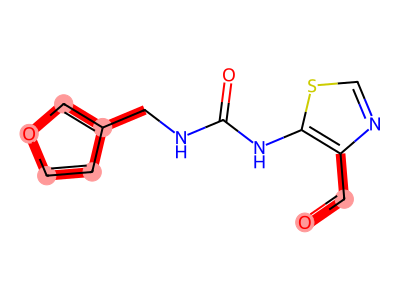} & \includegraphics[height=1.6cm]{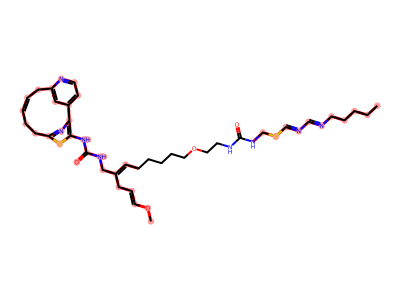} & \includegraphics[height=1.6cm]{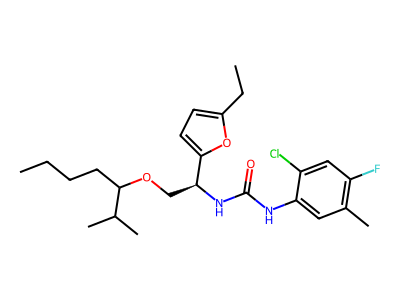} & \includegraphics[height=1.6cm]{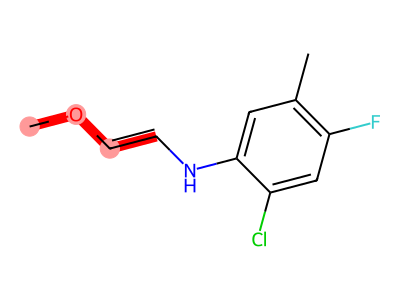} & \includegraphics[height=1.6cm]{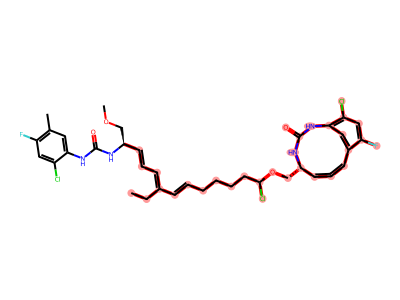} \\[-2pt]
{\scriptsize 450} & {\scriptsize 251 (\textcolor{red}{-199}) Sim 0.24} & {\scriptsize \textbf{738} (\textcolor{green!50!black}{+288}) Sim 0.18} & {\scriptsize 453} & {\scriptsize 216 (\textcolor{red}{-237}) Sim 0.28} & {\scriptsize \textbf{738} (\textcolor{green!50!black}{+285}) Sim 0.32} \\[6pt]
\includegraphics[height=1.6cm]{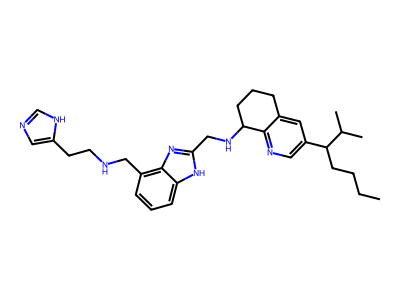} & \includegraphics[height=1.6cm]{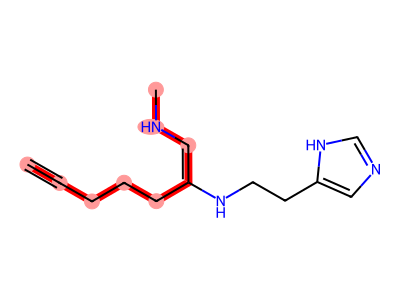} & \includegraphics[height=1.6cm]{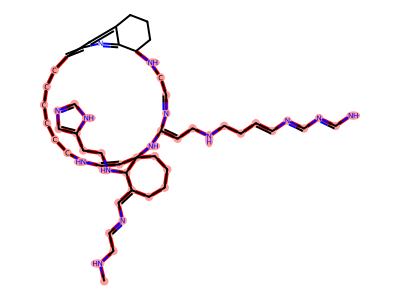} & \includegraphics[height=1.6cm]{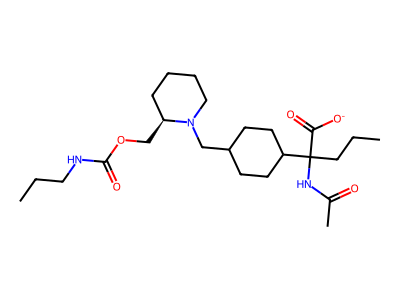} & \includegraphics[height=1.6cm]{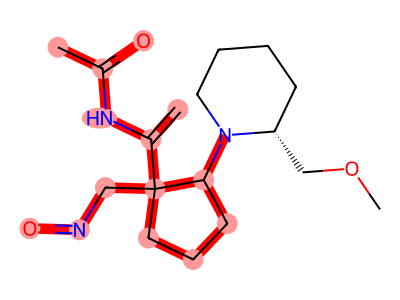} & \includegraphics[height=1.6cm]{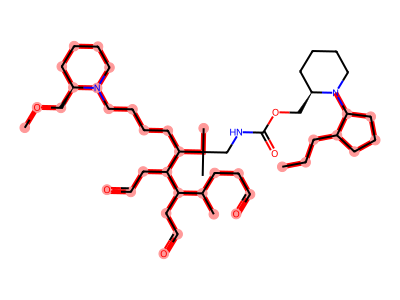} \\[-2pt]
{\scriptsize 514} & {\scriptsize 232 (\textcolor{red}{-281}) Sim 0.23} & {\scriptsize \textbf{792} (\textcolor{green!50!black}{+278}) Sim 0.24} & {\scriptsize 453} & {\scriptsize 323 (\textcolor{red}{-129}) Sim 0.23} & {\scriptsize \textbf{718} (\textcolor{green!50!black}{+265}) Sim 0.28} \\[6pt]
\includegraphics[height=1.6cm]{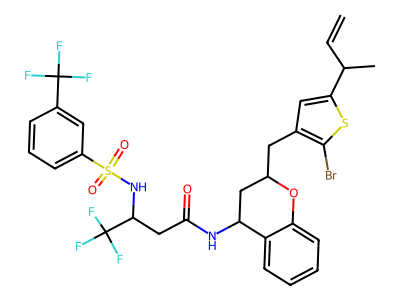} & \includegraphics[height=1.6cm]{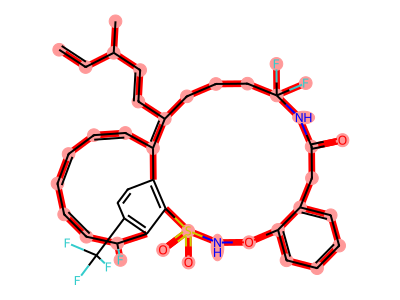} & \includegraphics[height=1.6cm]{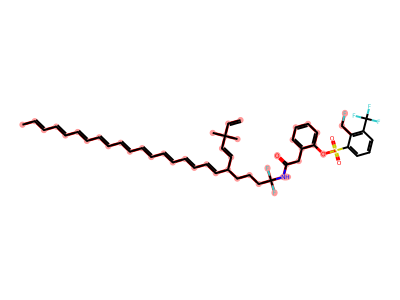} & \includegraphics[height=1.6cm]{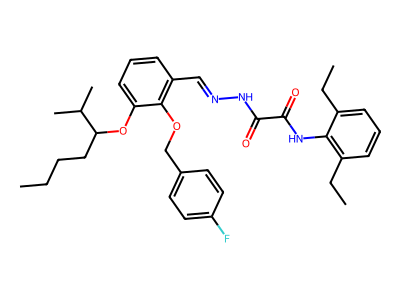} & \includegraphics[height=1.6cm]{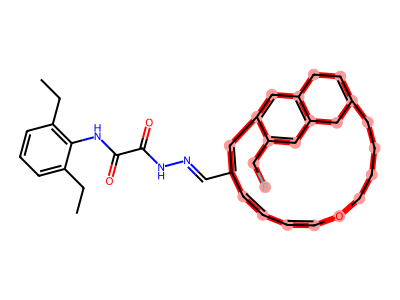} & \includegraphics[height=1.6cm]{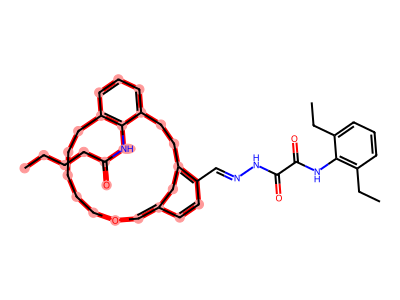} \\[-2pt]
{\scriptsize 726} & {\scriptsize 679 (\textcolor{red}{-47}) Sim 0.18} & {\scriptsize \textbf{840} (\textcolor{green!50!black}{+114}) Sim 0.19} & {\scriptsize 576} & {\scriptsize 556 (\textcolor{red}{-20}) Sim 0.30} & {\scriptsize \textbf{625} (\textcolor{green!50!black}{+49}) Sim 0.35} \\
\end{tabular}
\caption{\method steering examples: MW$\uparrow$ (MolGen-Large).}
\label{fig:mg_mw}
\end{figure*}

\begin{figure*}[ht]
\centering
\setlength{\tabcolsep}{1pt}
\begin{tabular}{ccc@{\hspace{8pt}}ccc}
\textbf{Source} & \textbf{SFT} & \textbf{+\method} & \textbf{Source} & \textbf{SFT} & \textbf{+\method} \\[2pt]
\includegraphics[height=1.6cm]{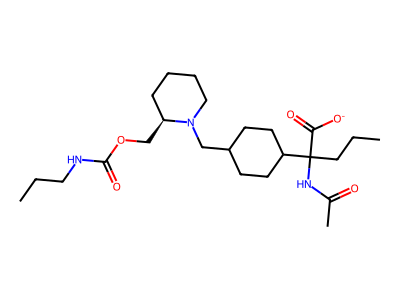} & \includegraphics[height=1.6cm]{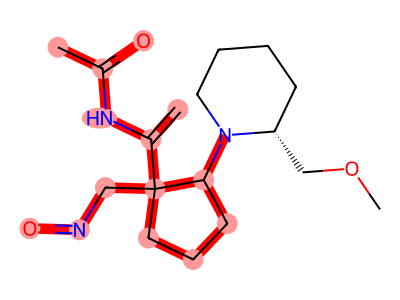} & \includegraphics[height=1.6cm]{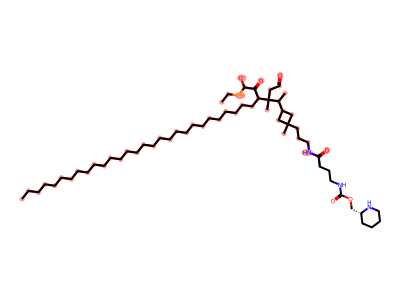} & \includegraphics[height=1.6cm]{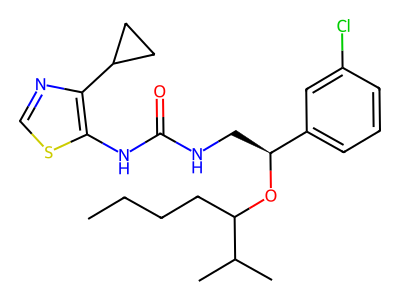} & \includegraphics[height=1.6cm]{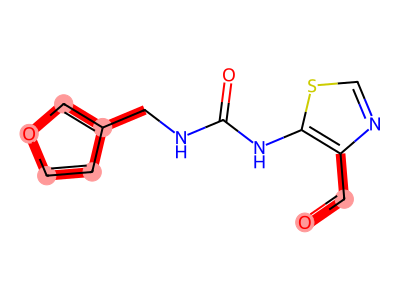} & \includegraphics[height=1.6cm]{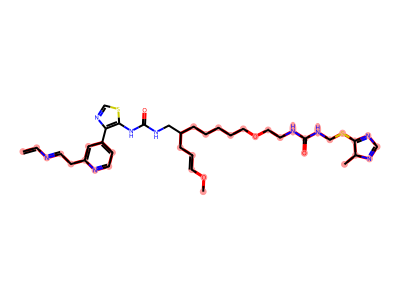} \\[-2pt]
{\scriptsize 11} & {\scriptsize 7 (\textcolor{red}{-4}) Sim 0.23} & {\scriptsize \textbf{47} (\textcolor{green!50!black}{+36}) Sim 0.21} & {\scriptsize 11} & {\scriptsize 4 (\textcolor{red}{-7}) Sim 0.24} & {\scriptsize \textbf{21} (\textcolor{green!50!black}{+10}) Sim 0.24} \\[6pt]
\includegraphics[height=1.6cm]{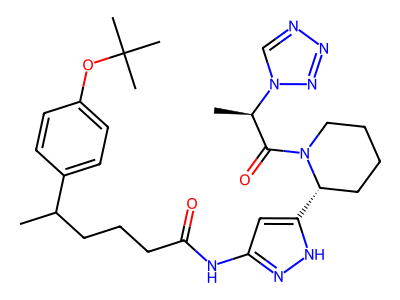} & \includegraphics[height=1.6cm]{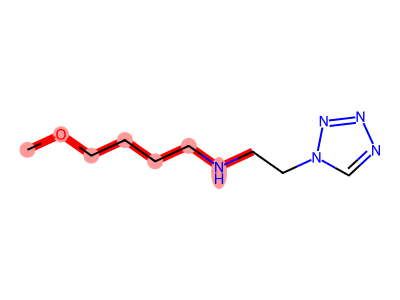} & \includegraphics[height=1.6cm]{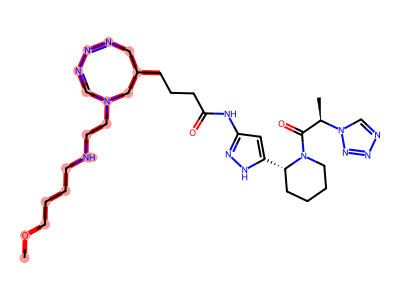} & \includegraphics[height=1.6cm]{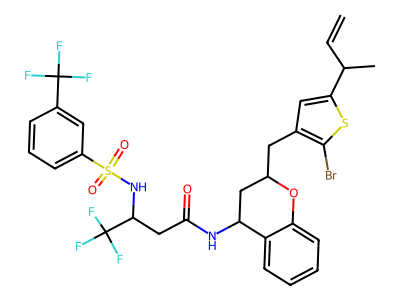} & \includegraphics[height=1.6cm]{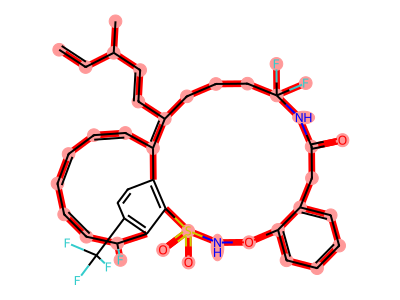} & \includegraphics[height=1.6cm]{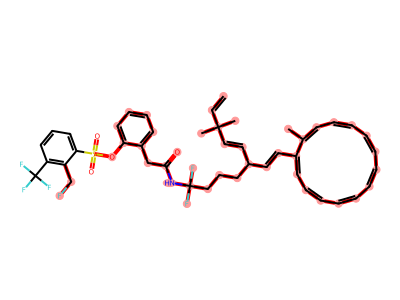} \\[-2pt]
{\scriptsize 10} & {\scriptsize 8 (\textcolor{red}{-2}) Sim 0.15} & {\scriptsize \textbf{16} (\textcolor{green!50!black}{+6}) Sim 0.55} & {\scriptsize 10} & {\scriptsize 3 (\textcolor{red}{-7}) Sim 0.18} & {\scriptsize \textbf{16} (\textcolor{green!50!black}{+6}) Sim 0.19} \\[6pt]
\includegraphics[height=1.6cm]{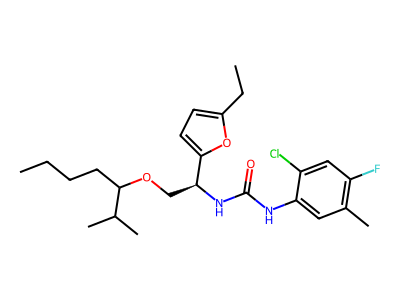} & \includegraphics[height=1.6cm]{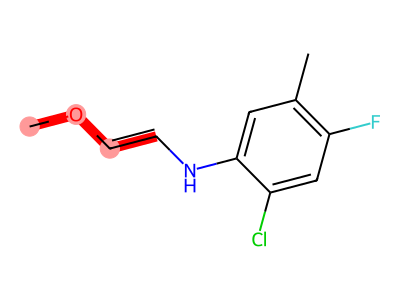} & \includegraphics[height=1.6cm]{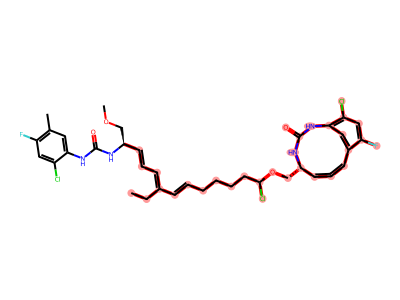} & \includegraphics[height=1.6cm]{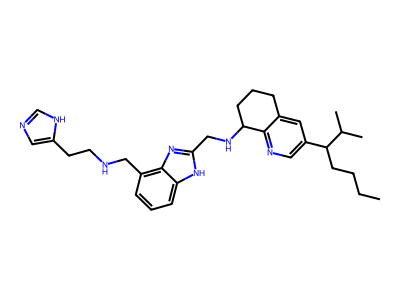} & \includegraphics[height=1.6cm]{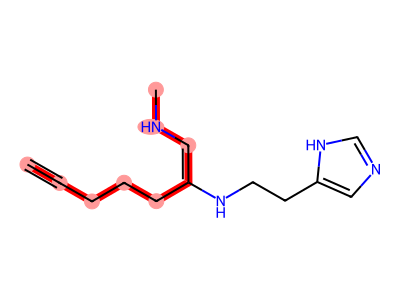} & \includegraphics[height=1.6cm]{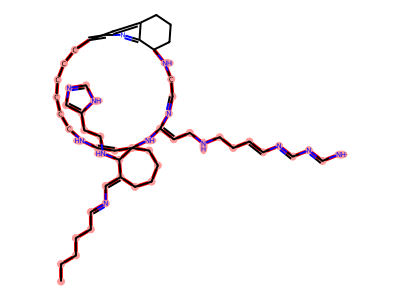} \\
\end{tabular}
\caption{\method steering examples: RotBond$\uparrow$ (MolGen-Large).}
\label{fig:mg_rotbond}
\end{figure*}

\begin{figure*}[ht]
\centering
\setlength{\tabcolsep}{1pt}
%
\caption{\method steering examples: SA$\downarrow$ (MolGen-Large).}
\label{fig:mg_sa}
\end{figure*}

\begin{figure*}[ht]
\centering
\setlength{\tabcolsep}{1pt}
%
\caption{\method steering examples: HBA$\uparrow$ (MolGen-Large).}
\label{fig:mg_acceptor}
\end{figure*}

\begin{figure*}[ht]
\centering
\setlength{\tabcolsep}{1pt}
%
\caption{\method steering examples: HBD$\uparrow$ (MolGen-Large).}
\label{fig:mg_donor}
\end{figure*}

\end{document}